\documentclass[acmlarge]{acmart}
\usepackage{multirow}
\usepackage{lipsum}

\usepackage{amsmath}
\usepackage{xspace}
\usepackage{hyperref}
\usepackage{wrapfig}
\usepackage{geometry}
\usepackage{color}
\usepackage{wrapfig,graphicx}
\usepackage{booktabs} 
\usepackage{array}
\usepackage{tabularx}
\usepackage{comment}
\usepackage[ruled]{algorithm2e} 

\newcolumntype{L}{>{\centering\arraybackslash}m{6cm}}
\SetAlFnt{\small}
\SetAlCapFnt{\small}
\SetAlCapNameFnt{\small}
\SetAlCapHSkip{0pt}
\IncMargin{-\parindent}

\acmJournal{CSUR}
\acmVolume{0}
\acmNumber{0}
\acmArticle{0}
\acmYear{0000}
\acmMonth{0}
\acmArticleSeq{00}

\setcopyright{usgovmixed}

\acmDOI{0000001.0000001}

\received{XXXXXXX 0000}
\received{XXXXXX 0000}
\received[accepted]{22 January 2019}

\begin{document} 
\title{Handcrafted and Deep Trackers: Recent Visual Object Tracking Approaches and Trends}  

\author{Mustansar Fiaz}
\affiliation{%
\href{https://orcid.org/0000-0003-2289-2284}{\includegraphics[scale=0.04]{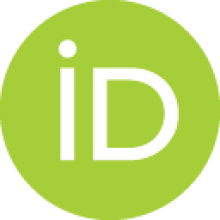}}
  \institution{Kyungpook National University}
  \department{School of Computer Science and Engineering}
  \city{Daegu}
  \country{Republic of Korea}
  }

\author{Arif Mahmood}
\affiliation{%
\href{https://orcid.org/0000-0001-5986-9876}{\includegraphics[scale=0.04]{figures/orcid.png}}
  \institution{Information Technology University}
  \department{Department of Computer Science}
  \city{Lahore}
  \country{Pakistan}
  }
  \author{Sajid Javed}
\affiliation{%
  \institution{University of Warwick}
  \department{Department of Computer Science}
  \country{United Kingdom}
  }
\author{Soon Ki Jung}
\affiliation{%
\href{https://orcid.org/0000-0003-0239-6785}{\includegraphics[scale=0.04]{figures/orcid.png}}
  \institution{Kyungpook National University}
  \department{School of Computer Science and Engineering}
  \city{Daegu}
  \country{Republic of Korea}
  }

\begin{abstract}


In recent years visual object tracking has become a very active research area. An increasing number of tracking algorithms  are being proposed each year. It is because tracking has wide applications in various real world problems such as human-computer interaction, autonomous vehicles, robotics, surveillance and security just to name a few. In the current study, we  review  latest trends and advances in the tracking area and evaluate the robustness of different trackers based on the feature extraction methods. The first part of this work comprises a comprehensive survey of the recently proposed trackers. We broadly categorize trackers into Correlation Filter based Trackers (CFTs) and  Non-CFTs. Each category is further classified into various types  based on the architecture and the tracking mechanism. In the second part, we experimentally evaluated 24 recent trackers for robustness, and compared handcrafted  and deep feature based trackers. We observe that trackers using deep features performed better, though in some cases a fusion of both increased performance significantly. In order to overcome the drawbacks of the existing benchmarks, a new benchmark  Object Tracking and Temple Color (OTTC) has also been proposed and used in the evaluation of different algorithms.  
We analyze the performance of  trackers over eleven different challenges in OTTC, and three other benchmarks.  Our study concludes that Discriminative Correlation Filter (DCF) based trackers  perform  better than the others. Our  study  also reveals that inclusion of different types of regularizations over DCF  often results in  boosted  tracking performance. Finally, we sum up our study by pointing out some insights and indicating future trends in visual object tracking field.

\end{abstract}

%
%



\maketitle

\renewcommand{\shortauthors}{Fiaz et al.}
\section{Introduction}


Visual Object Tracking (VOT) is a promising but difficult sub-field of computer vision. It attained much reputation because of its widespread use in different applications for instance autonomous vehicles \cite{laurense2017path}, traffic flow monitoring \cite{ tian2011video}, surveillance and security \cite{ali2016visual}, human machine interaction \cite{severson2017human}, medical diagnostic systems \cite{walker2017systems}, and activity recognition \cite{aggarwal2014human}. 
VOT is an attractive research area of computer vision due to opportunities and  different tracking challenges. In the previous few decades, remarkable endeavors are made by research community, but still VOT has much potential to explore further. The difficulty of VOT lies in the myriad of challenges, such as occlusion, background clutter, illumination changes, scale variation, low resolution, fast motion, out of view, motion blur,  deformation,  in and out planer rotation \cite{wu2013online, wu2015object}. 

VOT is the process of identifying a region of interest in a sequence and  consists of four sequential elements, including target initialization, appearance model, motion prediction, and target positioning. Target initialization is the process of annotating object position, or region of interest, with any of the following representations: object bounding box, ellipse, centroid, object skeleton,  object contour, or object silhouette. Usually, an object bounding box is provided in the initial frame of a sequence and the tracking algorithm estimates target position in the remaining frames. Appearance modelling is composed of identifying visual object features for better representation of a region of interest and effective construction of mathematical models to detect objects using learning techniques. In motion prediction, the target location is estimated in subsequent frames. The target positioning operation involves maximum posterior prediction, or greedy search. Tracking problems can be simplified by constraints imposed on the appearance and motion models. During the tracking, new target appearance is integrated by updating the appearance and motion models.

Currently, we have focused on monocular, model-free, single-target,  casual, and short-term trackers for experimental study. The \textit{model-free} characteristics hold for supervised training example in the initial frame provided by bounding box. The \textit{causality} means that tracker will predict the target location on current frame without prior knowledge of future frames. While, \textit{short-term} means that if a tracker is lost (fails) during the tracking, re-detection is not possible. And trackers output is specified by a bounding box.

Literature shows that much research has been performed on object tracking and various surveys have been published.
An excellent and extensive review of tracking algorithms is presented in \cite{yilmaz2006object} along with feature representations and challenges. However, the field has greatly advanced in recent years.  Cannons et al. \cite{cannons2008review} covered the fundamentals of object tracking problems, and discussed the building blocks for object tracking algorithms, the evolution of feature representations and different tracking evaluation techniques. Smeulders et al. \cite{smeulders2014visual} compared the performance of tracking algorithms and introduced a new benchmark. Li et al. \cite{li2013survey} and Yang et al. \cite{yang2011recent} discussed object appearance representations, and performed surveys for online generative and discriminative learning. Most of the surveys are somewhat outdated and subject to traditional tracking methods. Recently, 
the performance of tracking algorithm was boosted by the inclusion of deep learning techniques. Li et. al \cite{li2018deep} classified the deep trackers into Network Structure (NS), Network Function (NF), and Network Training (NT). Moreover, VOT challenge \cite{kristan2013visual, Kristan2014visual, Kristan2015visual, Kristan2016visual, Kristan2017visual} is providing the efficient comparison of the various trackers based and their brief introduction. However, our study differs in  two aspects: (1) recent tracking approaches and (2) Comparative study of trackers based on their feature extraction method.

The objective of the current study is to provide an overview of the recent progress, research trends, and to categorize existing tracking algorithms. Our motivation is to provide  interested readers an organized reference about the diverse tracking algorithms being developed, to help them find research gaps, and provide insights for developing new tracking algorithms. 

Features play an important  role in the performance of a tracker. There are two broad categories of the features used by the tracking algorithms including  HandCrafted (HC) and deep features. HC features such as Histogram of Oriented Gradients (HOG), Scale-Invariant Feature Transform (SIFT), Local Binary Pattern (LBP) and color names  were  commonly used to represent target appearance. Recently researchers have shifted their methodology and focus on deep features. Deep learning has shown remarkable success in various computer vision tasks such as object recognition and tracking, image segmentation, pose estimation, and image captioning. Deep features have many advantages over HC features because of having more potential to encode multi-level information and exhibit more invariance to target appearance variations. There are various deep feature extraction methods including Recurrent Neural Networks (RNN) \cite{graves2013speech}, Convolutional Neural Networks (CNN) \cite{simonyan2014very}, Residual Networks \cite{he2016deep}, and Auto-encoders \cite{zhuang2016visual}. In contrast to HC approaches, deep models are data-hungry and requiring a lot of training data. In applications with scarce training data, deep features are extracted using off-the-shelf pre-trained models such as VGGNet \cite{simonyan2014very}.
Despite the fact that deep features have achieved much success in single object tracking \cite{danelljan2015convolutional, DanelljanCVPR2017, bertinetto2016fully} but still HC features \cite{danelljan2015learning, Lukezic_CVPR_2017} produce comparative results and are being employed in tracking algorithms. 
We have  investigated  different trackers  performance on the basis of HC, deep features and combination of these features to get the broader aspect of the role of features in tracking performance.

As mentioned earlier, visual object tracking faces several  challenges and therefore numerous  algorithms are introduced. For example Zhang et al. \cite{zhang2014partial}, Pan and Hu \cite{pan2007robust} and Yilmaz et al. \cite{yilmaz2004contour} proposed tracking algorithms to handle occlusion in videos.  Similarly, to handle illumination variations, algorithms have been proposed by Zhang et al. \cite{zhong2012robust}, Adam et al. \cite{adam2006robust}, and  Babenko et al. \cite{babenko2009visual}. Moreover, Mei et al. \cite{mei2009robust}, Kalal et al. \cite{kalal2010pn}, and Kwon et al. \cite{kwon2010visual} handled the problem of cluttered background. Likewise, various tracking techniques have been developed to deal with other tracking challenges. Hence, there is a dire need to organize the literature associated with these challenges, to analyze the robustness of the trackers, and to categorize these algorithms according to the challenges available in the existing benchmarks. In the current work, we categorized the trackers according to  feature representation schemes such as handcrafted and deep feature based trackers, and analyzed the performance over eleven different  challenges.


The rest of this paper is organized as follows: Section II demonstrates related work; the  classification of recent tracking algorithms  and their brief introduction is explained in section III;  
experimental investigation is described in section IV; and the conclusion and future directions are described in section V.

\section{Related Work}

The research community has shown  keen interest in VOT, and developed  various state-of-the-art tracking algorithms. Therefore, an overview of research methodologies and techniques will be helpful in organizing domain knowledge. Trackers can be categorized as single-object vs. multiple-object trackers, generative vs. discriminative, context-aware vs. non-aware, and   online vs. offline learning algorithms. Single object trackers \cite{leang2018line, lee2016globally} are the algorithms tracking only one object in the sequence, while multi-object trackers \cite{berclaz2011multiple, leal2017tracking} simultaneously track multiple targets and follow their trajectories.  In generative models, the tracking task is carried out via searching the best-matched window, while discriminative models discriminate target patch from the background  \cite{qin2014object, yu2008online, yang2011recent}. 
In the current paper, recent tracking algorithms are classified as Correlation-Filter based Trackers (CFTs) and Non-CFTs (NCFTs). It is obvious from the names that CFTs \cite{gundogdu2016evaluation, sui2016real, chen2017visual}  utilize correlation filters, and non-correlation  trackers use other techniques \cite{gu2010efficient, khan2011robust, gong2011multi}.

Yilmaz et al. \cite{yilmaz2006object} presented a taxonomy of tracking algorithms and discussed tracking methodologies, feature representations, data association, and various challenges. Yang et al. \cite{yang2011recent} presented an overview of the local and global feature descriptors used to present object appearance, and reviewed  online learning techniques such as generative versus discriminative,  Monte Carlo sampling techniques, and integration of contextual information for tracking. Cannons \cite{cannons2008review}  discussed  object tracking components initialization, representations, adaption, association and estimation. He discussed the advantages and disadvantages of different feature representations and their combinations.  Smeulders et al. \cite{smeulders2014visual} performed  analysis and evaluation of different trackers with respect to a variety of tracking challenges.  They found sparse and local features more suited to handle illumination variations, background clutter, and occlusion. They used various evaluation techniques,  such as survival curves, Grubs testing, and Kaplan Meier statistics, and provided evidence that  F-score is the best  measure of tracking performance.
Li et al. \cite{li2013survey} gave a detailed summary of target appearance models. Their study included local and global feature representations, discriminative, and generative, and hybrid learning techniques. In recent times, deep learning has shown significant progress in visual trackers and Li et el. \cite{li2018deep} categorized deep learning trackers into three aspects: (1) network structure; network function; and network training.

Some relatively limited or focused reviews include the following works.
Qi et al. \cite{liu2014survey} focused on classification of online single target trackers. Zhang et al. \cite{zhang2013sparse} discussed tracking based on sparse coding, and classified sparse trackers. Ali et al. \cite{ali2016visual} discussed some  classical  tracking algorithms. Yang et al. \cite{yang2009context}  considered  context of tracking scene considering auxiliary objects \cite{yang2006intelligent} as the target context. Chen et al. \cite{chen2015experimental} examined only  CFTs. 
Arulampalam et al. \cite{arulampalam2002tutorial} presented  Bayesian tracking methods using particle filters. Most of these studies are outdated or consider only few algorithms and thus are limited in scope. 
In contrast, we presented a more comprehensive survey of recent contributions. We classified tracking algorithms as CFTs and NCFTs. We evaluated the performance accuracy of  HC and deep trackers. Moreover, tracking robustness has been examined over different challenges.


\section{Classification of Tracking Algorithms}\label{tracking_alogrithms_classification}

In this section, recent tracking algorithms are studied and most of them are proposed during the last four years. Each algorithm presents a different method to exploit target structure for predicting target location in a sequence. By analyzing the tracking procedure, we arrange these algorithms in a hierarchy and classify them  into two main categories: Correlation Filter Trackers (CFT) and Non-CFT (NCFT) with a number of subcategories in each class.

\begin{figure*}
\centering
\includegraphics[width=0.8\textwidth]{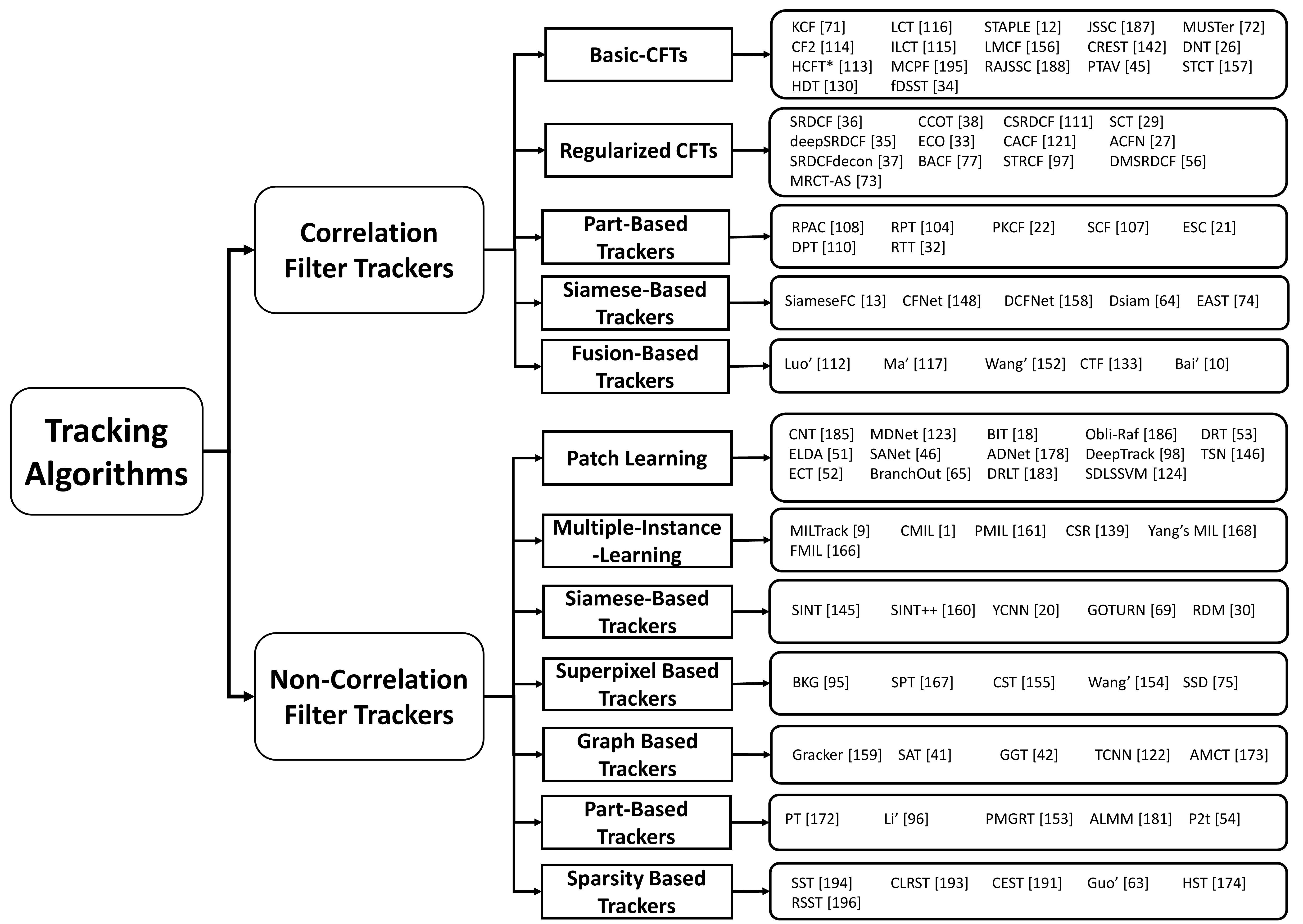}
\vspace{-10pt}
\caption{Taxonomy of tracking algorithms.}
\label{Hier1}
\end{figure*}

\subsection{Correlation Filter Trackers}


Discriminative Correlation Filters (DCF) are actively utilized in various computer vision applications including object recognition \cite{felzenszwalb2010object}, image registration \cite{essannouni2006fast}, face verification \cite{savvides2002face}, and action recognition \cite{rodriguez2008action}. In object tracking, Correlation Filters (CF)  have  been used to improve robustness and efficiency.  Initially, the requirement of training made CF inappropriate for online tracking.  In the later years, the development  of Minimum Output of Sum of Squared Error (MOSSE) filter \cite{bolme2010visual}, that allows for efficient adaptive training, changed the situation. The objective of the MOSSE filter is to minimize the sum of the squared error between the desired output and actual output in Fourier domain. MOSSE is an improved version of Average Synthetic Exact Filter (ASEF) \cite{bolme2009average} which is trained offline to detect objects. ASEF computes mean for a set of  exact filters, each computed from a different training image of the same object, to get best filter at the output.  Later on,  many state-of-the-art CFT were proposed based on MOSSE. Traditionally, the aim of designing inference of CF is to yield  response map that has low value for background and high value for region of interest in the scene. One such algorithm is Circulant Structure with Kernal (CSK) tracker \cite{henriques2012exploiting}, which exploits circulant structure of the target appearance and is trained using kernel regularized least squares method.


CF-based tracking schemes perform computation in the frequency domain to manage computational cost. General architecture of these algorithms follow "tracking-by-detection" approach and is presented in Fig. \ref{CF1}. Correlation filters are initialized from the initial frame of the sequence with a target patch cropped at the target position. During tracking, the target location is estimated in the new upcoming frame using the target estimated position in the last frame. To effectively represent  appearance of the target, appropriate feature extraction method is employed to construct feature map  from the input patch. Boundaries are smoothed by applying a cosine filter. Correlation operation is performed instead of exhausted convolution operation. The response map is computed using Element-wise multiplication between adaptive learning filter and extracted features,  and by using a Discrete Fourier Transform (DFT). DFT operates  in the frequency domain using Fast Fourier Transform (FFT). Confidence map is obtained in spatial domain by applying Inverse FFT (IFFT) over the response map. The maximum confidence score estimates the new target position. At the outcome, the target appearance at the newly predicted location is updated by extracting features and updating correlation filters.


Let $h$ be a correlation filter  and $x$ be the current frame, which may consist of the extracted features or the raw image pixels.  CNN convolutional filters  perform similar to correlation filters in Fourier domain. According to convolution theorem, correlation in frequency domain that computes a response map by performing element-wise multiplication between zero padded versions of  \textit{f($h$)} and \textit{f($x$)} is equivalent to circulant convolution in spatial domain. Often $h$ is of much smaller size compared to $x$, therefore before transformation to Fourier domain, zero padding is used such that transformed sizes of both are the same.
\vspace{-4.5pt}
 \begin{equation} \label{eq:cf1}
x	\otimes h= \mathfrak{F}^{-1}(\widehat{x} \odot \widehat{h}^\ast),
\end{equation}
where $\mathfrak{F}^{-1}$ indicates the IFFT, $\,$ $\widehat{ }$ $\,$ denotes Fourier representation, $\otimes$ represents convolution, $\odot$ means element-wise multiplication, and $\ast$ is the complex conjugate. Equation yields a confidence map between $x$ and $h$. To update the correlation filter, the estimated target around the maximum confidence position is selected. Assume $y$ is the desired output. Correlation filter $h$ must satisfy for new target appearance $z$ as:
 \begin{equation} \label{eq:cf2}
y= \mathfrak{F}^{-1}(\widehat z \odot \widehat h^\ast), \hspace{10mm} \widehat h^\ast ={\widehat y}/{\widehat z},   
\end{equation}
where $\widehat y$ denotes the desired output $y$ in frequency domain and division operation is performed during element-wise multiplication. FFT reduces the computational cost, as circulant convolution has a complexity of $O(n^4)$ for image size $n$x$n$ while FFT require only $O(n^2 \log n)$.

CF-based tracking frameworks face different difficulties, such as the training of the target appearance (orientation, and shape), as it may change over time. Another challenge is the selection of  an efficient feature representation, as powerful features may improve the performance of CFTs. Another important challenge for CFTs is scale adaption, as the size of correlation filters are fixed during tracking. A target may change its scale over time. Furthermore, if the target is lost then it cannot be recovered again. CFTs are further divided into the categories  \textit{B}-CFTs, regularized CFTs, part-based, Siamese-based, and Fusion-based CFTs as explained below. 

\subsubsection{Basic Correlation Filter based Trackers}

\textbf{Basic-CFTs} are  tackers that use Kernelized Correlation Filters (KCF) \cite{henriques2015high} as their baseline tracker. Trackers may use different features such as the HOG, colour names (CN) \cite{danelljan2014adaptive} and deep features using  Recurrent Neural Networks (RNN) \cite{williams1989learning}, Convolutional Neural Networks (CNN) \cite{simonyan2014very},  and Residual Networks \cite{he2016deep}. Numerous trackers have been developed using KCF as base tracker including \cite{ma2015hierarchical,  Ma2018RobustVT, qi2016hedged,  ma2015long,  ma2018adaptive, Zhang_2017_CVPR, danelljan2017discriminative, bertinetto2016staple,  zhang-ras2015joint, zhang-JSSC2015robust, song-iccv17-CREST, fan2017parallel, wang2016stct, hong2015multi, chi2017dual, wang2017large}.

A KCF \cite{henriques2015high} algorithm performs tracking using Gaussian kernel function for distinguishing between target object and its surroundings. HOG descriptors of cell size four is employed by KCF. During tracking, an image patch is cropped in new frame, HOG features are computed for that patch, and a response map is computed by multiplying adaptive filters on input features in Fourier domain. A new target position is predicted at the position of maximum confidence score in the confidence map obtained by applying inverse Fourier transform on response map. A new patch containing object  is then cropped and HOG features are recomputed to update the CF. 

\begin{figure}
  \begin{minipage}[b]{.49\textwidth}
    \centering
    \includegraphics[width=\linewidth]{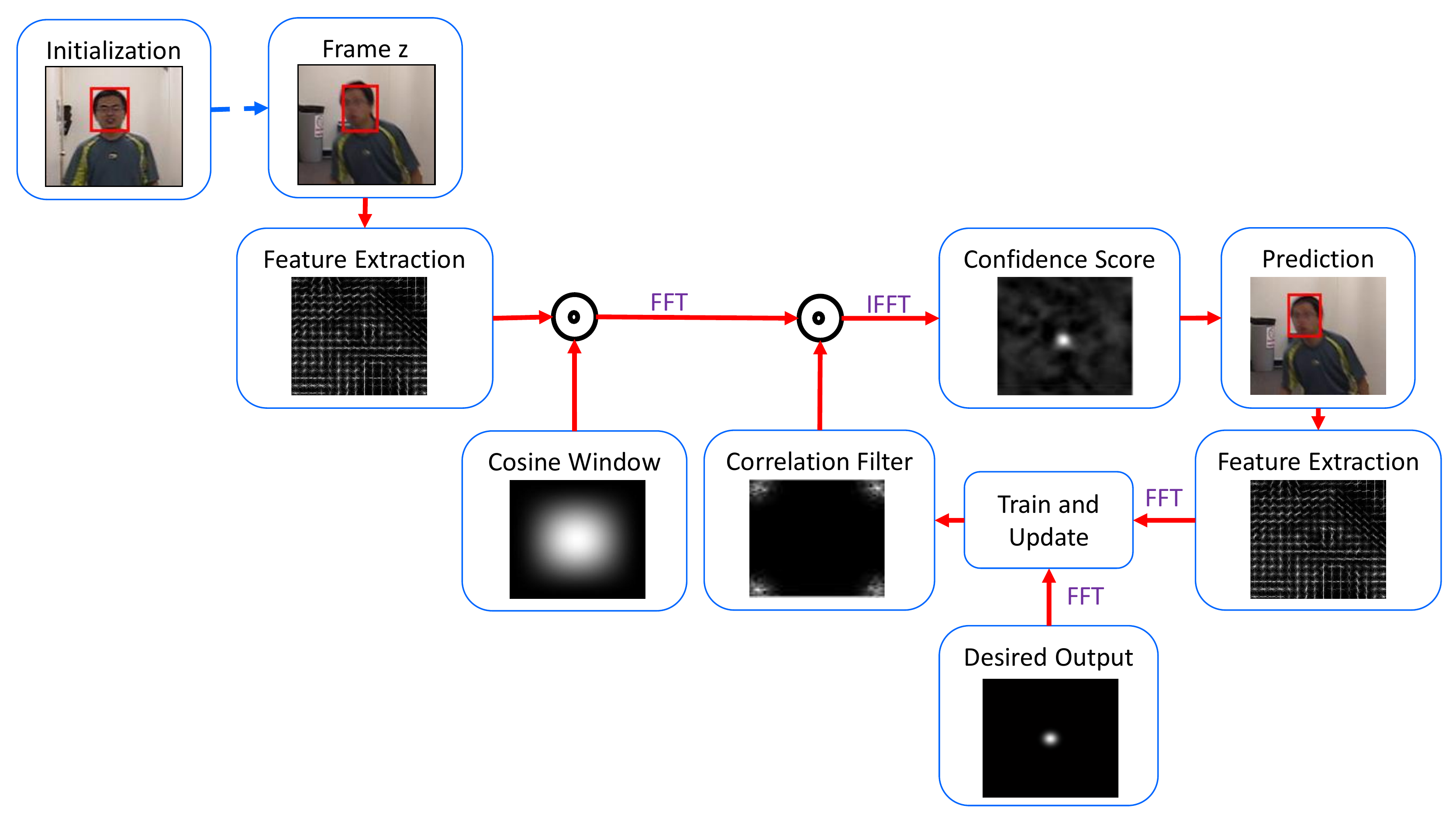}
    \vspace{-8mm}
    \caption{The framework of correlation filter for visual object tracking  \cite{chen2015experimental}.}
     \label{CF1}
  \end{minipage}\hfill
  \begin{minipage}[b]{.49\textwidth}
    \centering
    \includegraphics[width=\linewidth]{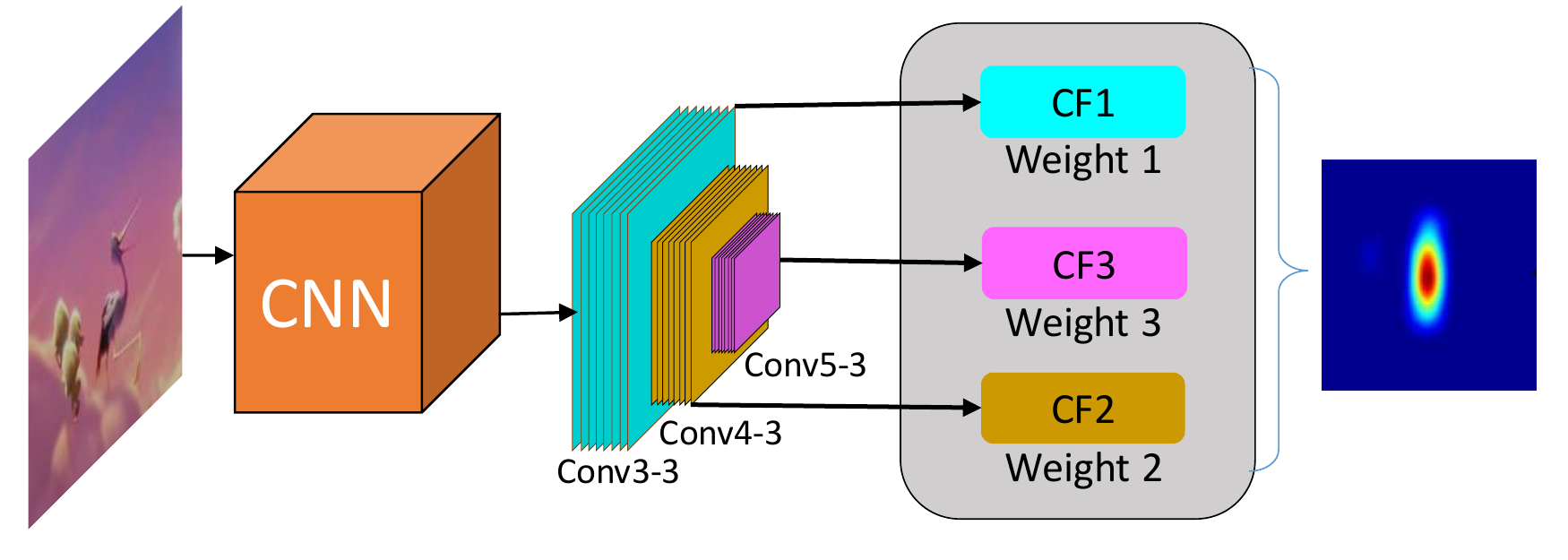}
    \vspace{-8mm}
    \caption{CF2 framework \cite{ma2015hierarchical}.}
     \label{cf2_structure}
  \end{minipage}
 \vspace{-15pt}
\end{figure}


Ma et al.\cite{ma2015hierarchical} exploited  rich hierarchical Convolutional Features using CF represented as CF2 (Fig. \ref{cf2_structure}). For every subsequent frame, the search area is cropped at the center based on previously predicted target position. Three hierarchical convolutional features are extracted from VGGNet  layers conv3-4, conv4-4, and conv5-4 to exploit target appearance. 
Deep features are resized to the same size using bilinear interpolation.  
An independent adaptive CF is utilized for each CNN feature, and response maps are computed. A coarse-to-fine methodology is applied over the set of correlation response maps to estimate the new target position. Adaptive hierarchical CFs are updated on newly-predicted target location. Ma et al. \cite{Ma2018RobustVT} also proposed Hierarchical Correlation Feature based Tracker (HCFT*), which is an extension of CF2 that integrates re-detection and scale estimation of target.

The Hedged Deep Tracking  (HDT) \cite{qi2016hedged} algorithm takes advantage of multiple levels of CNN features. In HDT, authors hedged many weak trackers  together to attain a single strong tracker. During tracking, the target position at the previous frame is utilized to crop a new image 
to compute six deep features using VGGNet. 
Deep features are exploited to individual CF to compute response maps also known as weak experts. The target position is estimated by each weak tracker, and the loss for each expert is also computed.  A standard hedge algorithm is used to estimate the final position. All weak trackers are hedged together into a strong single tracker by applying an adaptive online decision algorithm. Weights for each weak tracker are updated during online tracking. In an adaptive Hedge algorithm, a regret measure is computed for all weak trackers as a weighted average loss. A stability measure is computed for each expert based on the regret measure. The hedge algorithm strives  to minimize the cumulative regret of weak trackers depending upon its historical information. 

The Long-term Correlation Tracking (LCT) \cite{ma2015long} involves exclusive prediction of target translation and scale using correlation filters and random fern classifier \cite{ozuysal2007fast} is used for online re-detection of the target during tracking. In LCT algorithms, the search window is cropped on the previously estimated target location and a feature map is computed. Translation estimation is performed using adaptive translation correlation filters. A target pyramid is generated at the newly predicted target translation position, and scale is estimated using a separate regression correlation model. In case of failure, the LCT tracking algorithm performs re-detection. If the  estimated target score is less then a threshold,  re-detection is then performed using online random fern classifier. Average response is computed using posteriors from all the ferns. LCT selects the positive samples to predict new patch as target by using the \textit{k}-nearest neighbor (KNN) classifier. Author has further Improved LCT (ILCT) \cite{ma2018adaptive} using SVM classifier instead of fern classifier for re-detection. 


The Multi-task Correlation Particle Filter (MCPF) proposed by Zhang et al. \cite{Zhang_2017_CVPR} employ particle filter framework. 
The MCPF shepherd particles in the search region by exploiting all states of the target. The MCPF computes response maps of particles, and target position is estimated as weighted sum of the response maps. 
Danelljan et al. \cite{danelljan2017discriminative} proposed Discriminative Scale Space Tracking (DSST) to separately estimate translation and scale by  learning independent CFs.  Scale estimation is done by learning the target sample at various scale variations. In proposed framework, first translation is predicted by applying a standard translation filter to every incoming frame. After translation estimation, the target size is approximated by employing trained scale filter at the target location obtained by the translation filter. 
This way, the proposed strategy learns the target appearance induced by scale rather than by using exhaustive target size search methodologies. 
The author further improved the computational performance without sacrificing  the robustness and accuracy. Fast DSST (fDSST)  employs sub-grid interpolation to compute correlation scores.

The Sum of Template And Pixel-wise LEarners (STAPLE)  \cite{bertinetto2016staple} employed two separate regression models to solve the tacking problem by utilizing the inherent structure for each target representation.  The tracking design takes advantage of two complementary factors from two different patch illustrations to train a model. HOG features  and global color histograms are used to represent the target. In the colour template, foreground and background regions are computed at previously estimated location. The frequency of each colour bin for object and background are updated, and a regression  model for colour template is computed. In the search area, a per-pixel score is calculated based on previously estimated location, and the integral image is used to compute response, while for the HOG template, HOG features are extracted at the  position predicted in the previous frame, and CF are updated. At every incoming frame, a search region is extracted centered at previous estimated location, and their HOG features are convolved with CF to obtain a dense template response. Target position is estimated from template and histogram response scores as a linear combination. Final estimated location is influenced by the model which has more scores.  

The Convolutional RESidual learning for visual Tracking (CREST) algorithm \cite{song-iccv17-CREST} utilizes residual learning \cite{he2016deep} to adapt target appearance and also performs scale estimation by searching patches at different scales. During  tracking, the search patch is cropped at previous location, and convolutional features are computed. Residual and base mapping are used to compute the response map. The maximum response value gives the newly estimated target position. Scale is estimated by exploring different scale patches at newly estimated target center position. 
The Parallel Tracking And Verifying (PTAV) \cite{fan2017parallel}  is composed of two major modules, i.e. tracker and verifier. Tracker module is responsible for computing the real time inference and estimate tracking results, while the verifier is responsible for checking whether the results are correct or not. 
The Multi-Store tracker (MUSTer) \cite{hong2015multi} avoid drifting and stabilizes the tracking by aggregating image information using short and long term stores, and is based on the Atkinson-Shiffrin memory model. Short term storage involves an integrated correlation filter to incorporate spatiotemporal consistency, while long term storage involves integrated RANSAC estimation and key point match tracking to control the output. 

\subsubsection{Regularized Correlation Filter Trackers}

Discriminative CF (DCF) tracking algorithms are limited in their detection range because they require filter size and patch size to be equal. The DCF may learn the  background for irregularly-shaped target objects. The DCF is formulated from periodic assumption, learns from a set of training samples, and  thus may learn negative training patches. DCF response maps have accurate scores close to the centre, while other  scores are influenced due to periodic assumption, thus degrading DCF performance. Another limitation of DCFs is that they are restricted to only a fixed search region. DCF trackers perform poorly on a target deformation problem due to over fitting of model caused by learning from target training samples but missing the negative samples. Thus, the tracker fails to re-detect in case of occlusion.  A larger search region may solve the occlusion problem but the model will learn background information which  degrades the  discrimination power of the tracker. Therefore, there is a need to incorporate a measure of regularization for these DCF limitations and those trackers are classified as \textbf{Regularized Correlation Filter Trackers} (R-CFTs). Several R-CFTs have been proposed such as \cite{danelljan2015learning,  danelljan2015convolutional, danelljan2016adaptive, Li2018STRCF, gladh2016deep, DanelljanECCV2016, DanelljanCVPR2017, Lukezic_CVPR_2017, mueller2017context, kiani2017learning, choi2016visual, choi2017attentional, hu2017manifold}.
\begin{figure}
     \vspace{-20pt}
     \captionsetup{width=.7\textwidth}
  \begin{center}
    \includegraphics[width=0.7\textwidth]{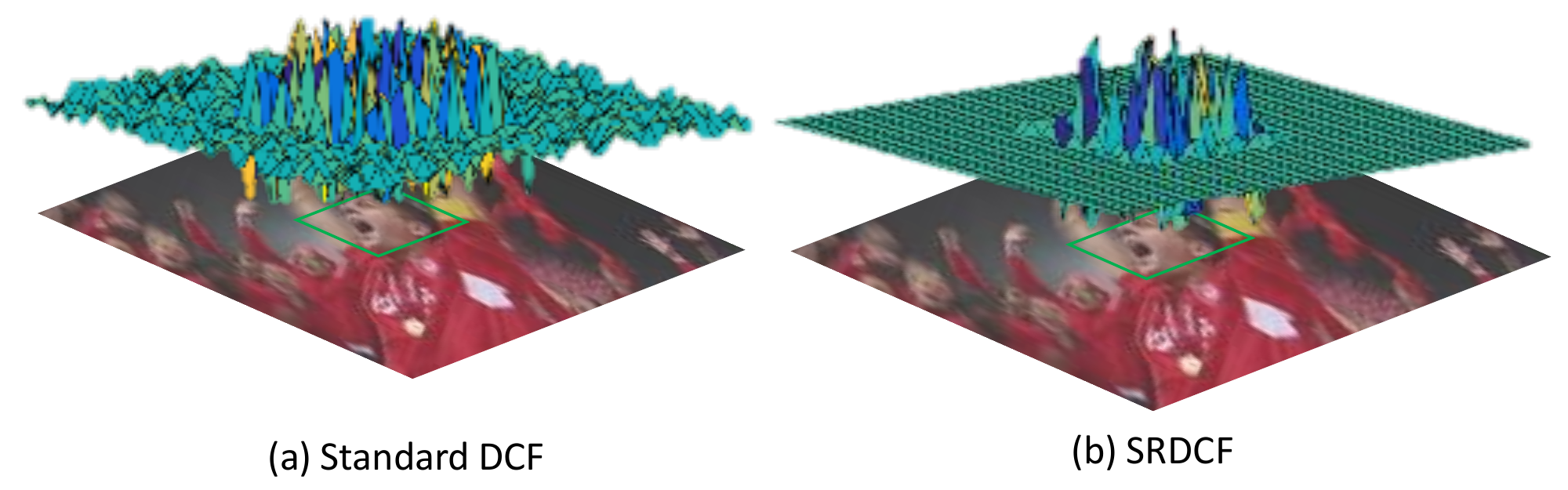}
    \end{center}
     \vspace{-15pt}
         \caption{Difference between standard DCF and SRDCF \cite{danelljan2015learning}.}
\label{SRDCF_regularization}
\vspace{-17pt}
\end{figure}

Danelljan et al.\cite{danelljan2015learning} presented Spatially Regularized DCF (SRDCF) by introducing spatial regularization in DCF learning. During  tracking, the regularization component weakens the background information as shown in Fig. \ref{SRDCF_regularization}. Spatial regularization constraints the filter coefficients based on spatial information. The background is suppressed by assigning higher values to the  coefficients that are located outside  the target territory and vice verse. 
The SRDCF framework has been updated by using deep CNN features in deepSRDCF \cite{danelljan2015convolutional}. The SRDCF framework has also been modified to handle contaminated training samples in SRDCFdecon \cite{danelljan2016adaptive}. It down weights corrupted training samples and estimate good quality samples. SRDCFdecon extracts training samples from previous frames and then assign higher weights to correct training patches. SRDCFdecon performs joint adaptation of both appearance model and weights of the training samples.  

Li et al. \cite{Li2018STRCF} introduced the temporal regularization in SRDCF and introduced Spatial-Temporal Regularized CF (STRCF). Temporal regularization has been induced using passive aggressive learning to SRDCF with single image.
Recently, deep motion features have been used for activity recognition \cite{kiani2015correlation}. Motion features are obtained from  information obtained directly  from optical flow applied to images. A CNN is then applied to optical flow to get deep motion features. Gladh et al.\cite{gladh2016deep} presented Deep Motion SRDCF (DMSRDCF) which fused deep motion features along with hand-crafted appearance features using SRDCF as baseline tracker. Motion features are computed as reported by \cite{cheron2015p}. Optical flow is calculated on each frame  on previous frame. 
The x, y components and magnitude of optical flow constitute three channels in the flow map, which is normalized between 0 and 255 and fed to the CNN to compute deep motion features.

Danelljan et al.\cite{DanelljanECCV2016} proposed  learning multi-resolution feature maps, which they name as Continuous Convolutional Operators for Tracking (CCOT). The convolutional filters are learned in a continuous sequence of resolutions which generates a sequence of response maps. These multiple response maps are then fused to obtain final unified response map to estimate target position. 
The Efficient Convolution Operators (ECO) \cite{DanelljanCVPR2017} tracking scheme is an improved version of CCOT. The CCOT learns a large number of filters to capture target representation from high dimensional features, and updates the filter for every frame, which involves training on a large number of sample sets.  In contrast, ECO constructs a smaller set of filters to efficiently capture target representation using matrix factorization.  The CCOT learns over consecutive samples in a sequence which forgets target appearance for a long period thus causes overfitting to the most recent appearances and   leading to high computational cost. In contrast, ECO uses a Gaussian Mixture Model (GMM) to represent diverse target appearances.  Whenever a new appearance is found during tracking, a new GMM component is initialized. 
If the maximum limit of components is achieved, then a GMM component with minimum weight is discarded if its weight is less than a threshold value. Otherwise, the two closest components are merged into one component. 

The Channel Spatial Reliability for DCF (CSRDCF) \cite{Lukezic_CVPR_2017} tracking algorithm integrates channel and spatial reliability with DCF. Training patches also contain non-required background information in addition to the required target information. Therefore, DCFs may also learn background information, which may lead to the drift problem.
In CSRDCF, spatial reliability is ensured by estimating a spatial binary map at current target position to learn only target information. 
Foreground and background models retained as colour histogram are used to compute appearance likelihood using Bayes' rule. 
A constrained CF is obtained by convolving the CF with spatial reliability map that indicates which pixels should be ignored. Channel reliability is measured as a product of channel reliability measure and detection reliability measures. The channel reliability measure is the maximum response of channel filter. Channel detection reliability in response map is computed from the ratio between the second and first major modes, clamped at 0.5. Target position is estimated at maximum response of search patch features and the constrained CF, and is weighted by channel reliability. 

Mueller et al. \cite{mueller2017context} proposed Context Aware Correlation Filter tracking (CACF) framework where global context information is integrated within Scale Adaptive Multiple Feature (SAMF) \cite{li2014scale} as baseline tracker. The model is improved to compute high responses for targets, while close to zero responses for context information. 
The SAMF uses KCF as baseline and solves the scaling issue by constructing a pool containing the target at different scales. Bilinear interpolation is employed to resize the samples in the pool to a fixed size template. 
Kiani et al. \cite{kiani2017learning} exploited the background patches and proposed Background Aware Correlation Filter (BACF) tracker.
The Structuralist Cognitive model for Tacking (SCT) \cite{choi2016visual} divides the target into several cognitive units.
During tracking, the search region is decomposed into fixed-size grid map, and an individual Attentional Weight Map (AWM) is computed for each grid cell. The AWM is computed from the weighted sum of Attentional Weight Estimators (AWE). The AWE assigns more weights to target grid which less weights are given to background grid using a Partially Growing Decision Tree (PGDT) \cite{choi2015user}. Each unit works as individual KCF with Attentinal CF (AtCF), having different kernel types with distinct features  and corresponding AWM.  
The priority and reliability of each unit are computed based on relative performance among AtCFs and its own performance, respectively. Integration of response maps of individual units gives target position. 

Choi et al. proposed a Attentional CF Network (ACFN) \cite{choi2017attentional} exploits  target dynamic based on an attentional mechanism. An ACFN is composed of a CF Network (CFN) and Attentional Network (AN). The CFN has several tracking modules that compute tracking validation scores as precision. The KCF is used for each tracking module with AtCF and AWM. The AN selects tracking modules to learn target dynamics and properties. The AN  is composed of 2 Sub Networks (SN) such as Prediction SN (PSN) and Selection SN (SSN). Validation scores for all modules are predicted in PSN. The SSN chooses active tracking modules based on current estimated scores. Target is estimated as that having the best response among the selected subset of tracking modules.  

\subsubsection{Siamese-Based Correlation Filter Trackers}


\begin{figure}
  \begin{minipage}[b]{.48\textwidth}
    \centering
    \includegraphics[width=\linewidth]{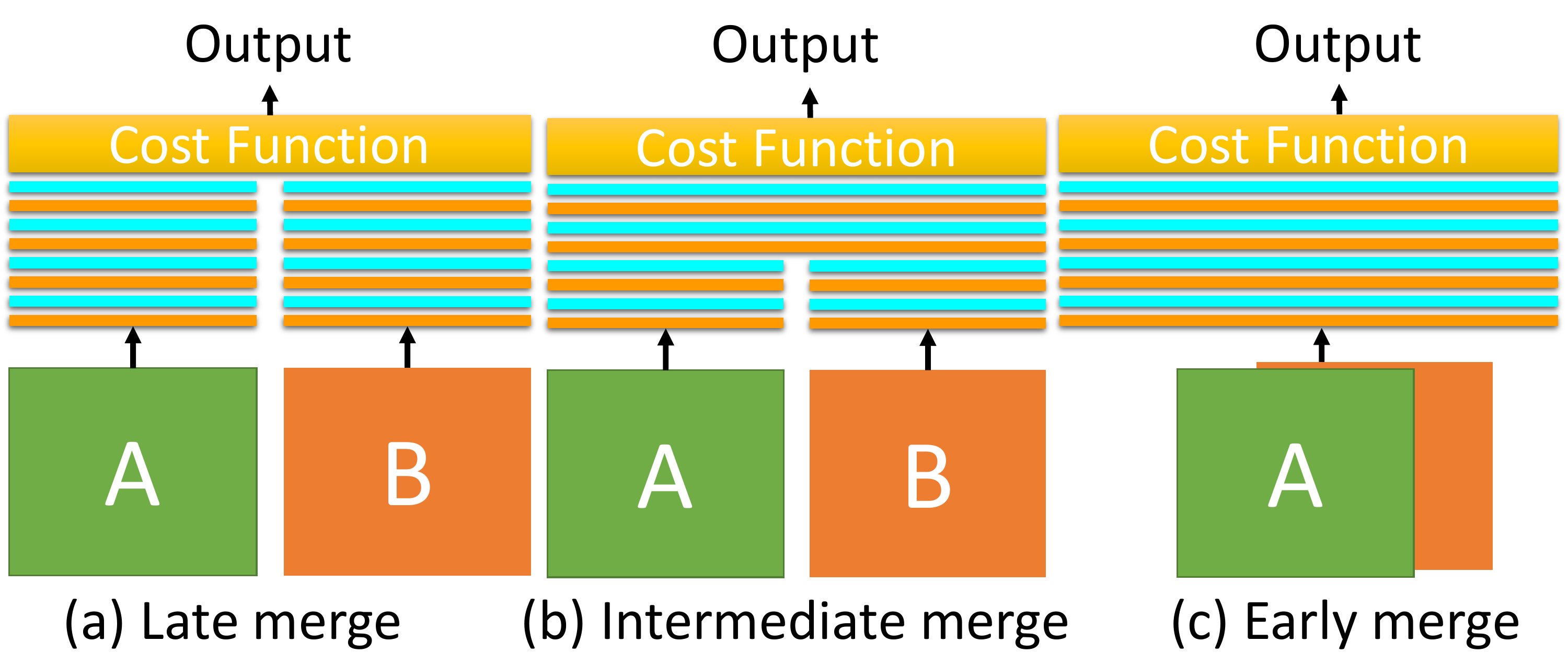}
    \vspace{-8mm}
    \caption{Siamese CNN typologies \cite{leal2016learning}.}
     \label{Siamese_topology}
  \end{minipage}\hfill
  \begin{minipage}[b]{.48\textwidth}
    \centering
    \includegraphics[width=\linewidth]{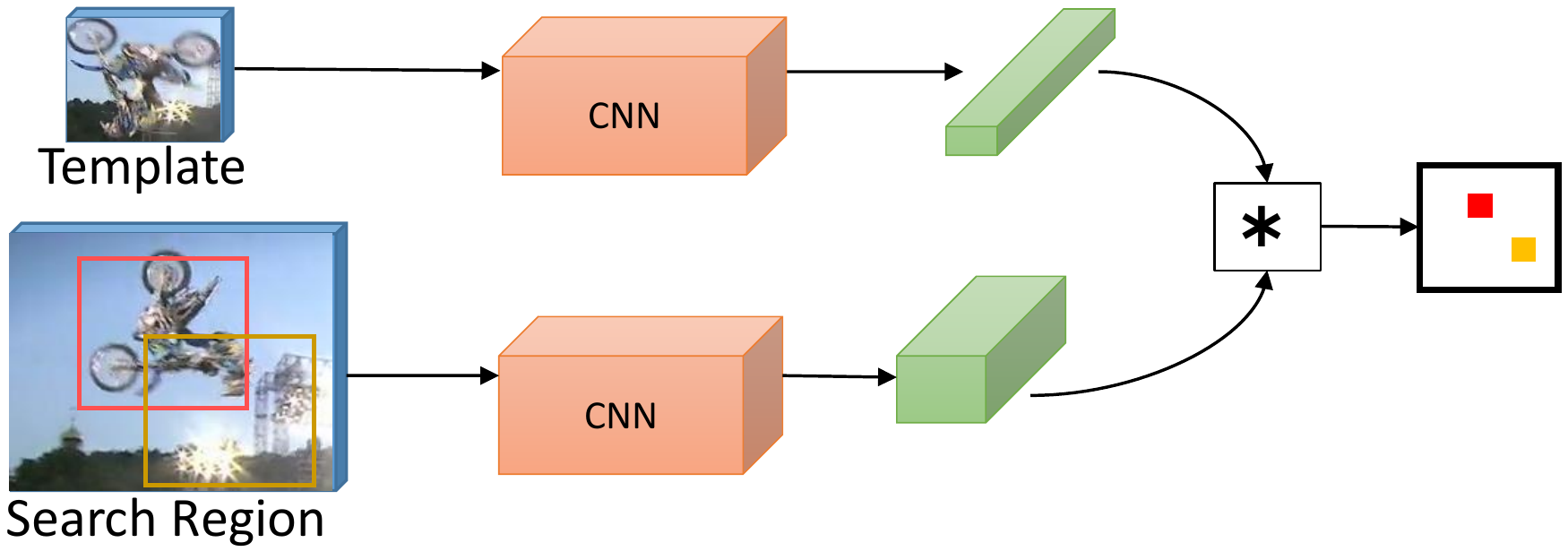}
    \vspace{-8mm}
    \caption{The framework of SiameseFC \cite{bertinetto2016fully}.}
     \label{SiameseFC}
  \end{minipage}
  \vspace{-5mm}
\end{figure}

A Siamese network joins two inputs and produces a single output. The objective is to determine whether identical objects exist, or not, in the two image patches that are input to the network. The network measures similarity between the two inputs, and has the capability to learn similarity and features jointly. The  concept of Siamese network was initially used for signature verification and fingerprint recognition.
Later, Siamese networks were used in various applications, including face recognition and verification \cite{schroff2015facenet}, stereo matching \cite{zbontar2015computing}, optical flow \cite{dosovitskiy2015flownet}, large scale video classification \cite{karpathy2014large} and patch matching \cite{zagoruyko2015learning}.   We observe that Siamese architecture using CNN find similarity between two images using shared (convolutional and/or fully connected) layers. Siamese CNN can be categorized into three main groups based on late merge, intermediate merge, and early merge \cite{leal2016learning} (Fig .\ref{Siamese_topology}).
\vspace{-3pt}
\begin{itemize}
\item Late merge: Two image patches are evaluated separately in parallel  using the same network and combined at the final layer \cite{chopra2005learning}.
\item Intermediate merge: The network processes input images separately and then merges into a single stream well before the final layer \cite{dosovitskiy2015flownet}. 
\item  Early merge: Two image are stacked and a unified input is fed to single CNN. 
\end{itemize}
\vspace{-3pt}
Integration of CFTs with Siamese network for visual tracking are classified as \textbf{Siamese-Based CFTs} and has been used to handle  tracking challenges \cite{guo2017learning, bertinetto2016fully, valmadre2017end, wang17dcfnet, huang2017learning}.


Siamese Fully Convolutional  networks (SiameseFC) \cite{bertinetto2016fully} shown in fig. \ref{SiameseFC} solves the tracking problem using similarity learning that compares exemplar (target) image with a same-size candidate image, and yields high scores if the two images are the same. The SiameseFC algorithm is fully convolutional, and its ouptput is a scalar-valued score map that takes as input an example target and search patch larger than target predicted in the previous frame. 
The SiameseFC network utilizes a convolutional embedding function and a correlation layer to integrate the deep feature maps of the target and search patch. Target position is estimated at  maximum value in response map. This  gives frame to frame target displacement.  
Valmadre et al. \cite{valmadre2017end} introduced Correlation Filter Network (CFNet) for end-to-end learning of underlying feature representations through gradient back propagation. 
SiameseFC is used as base tracker, and CFNet is employed in forward mode for online tracking. During the online tracking of CFNet, target features are compared with the larger search area on new frame based on previously estimated target location. A similarity map between the target template and the search patch is produced by calculating the cross-correlation. 

The Discriminative Correlation Filters Network (DCFNet) \cite{wang17dcfnet}  utilizes lightweight CNN network with correlation filters to perform tracking using offline training. The DCFNet performs back propagation to adapt the CF layer utilizing a probability heat-map of target position. 

Recently, Guo et al. \cite{guo2017learning} presented DSaim that has the potential to reliably learn online temporal appearance variations. The DSaim exploits CNN features for target appearance and search patch. Contrary to the SiameseFC, the DSaim learns target appearance and background suppression from previous frame by introducing regularized linear regression . 
Target appearance variations are learned from first frame to current frame, while background suppression is performed by multiplying the search patch with the learned Gaussian weight map. The DSaim performs element-wise deep feature fusion through circular convolution layers to multiply inputs with weight map. Huang presented  EArly Stopping Tracker (EAST) \cite{huang2017learning} to learn polices using deep Reinforcement Learning (RL) and improving speedup while maintaining accuracy. The tracking problem is solved using Markov decision process. A RL agent makes decision based on  multiple  scales with an early stopping criterion. 

\subsubsection{Part-Based Correlation Filter Trackers}

These kind of trackers learn target appearance in parts, while in other CFTs  target template is learned as a whole. Variations may appear in a sequence, not just because of illumination and viewpoint, but also due to intra-class variability, background clutter, occlusion, and deformation. For example, an object may appear in front of the object being tracked, or a target may undergo non-rigid appearance variations.  Part-based strategies are widely utilized in several applications, including object detection \cite{forsyth2014object}, pedestrian detection \cite{prioletti2013part} and face recognition \cite{karczmarek2017application}.
Several part-based trackers \cite{liu2015real, li2015reliable, cui2016recurrently, chen2017visual, liu2016structural, chen2016robust, lukevzivc2017deformable} have been developed to solve the challenges where targets are occluded or deformed in the sequences. 

\begin{figure*}
     \vspace{-5pt}
  \begin{center}
    \includegraphics[width=0.7\textwidth]{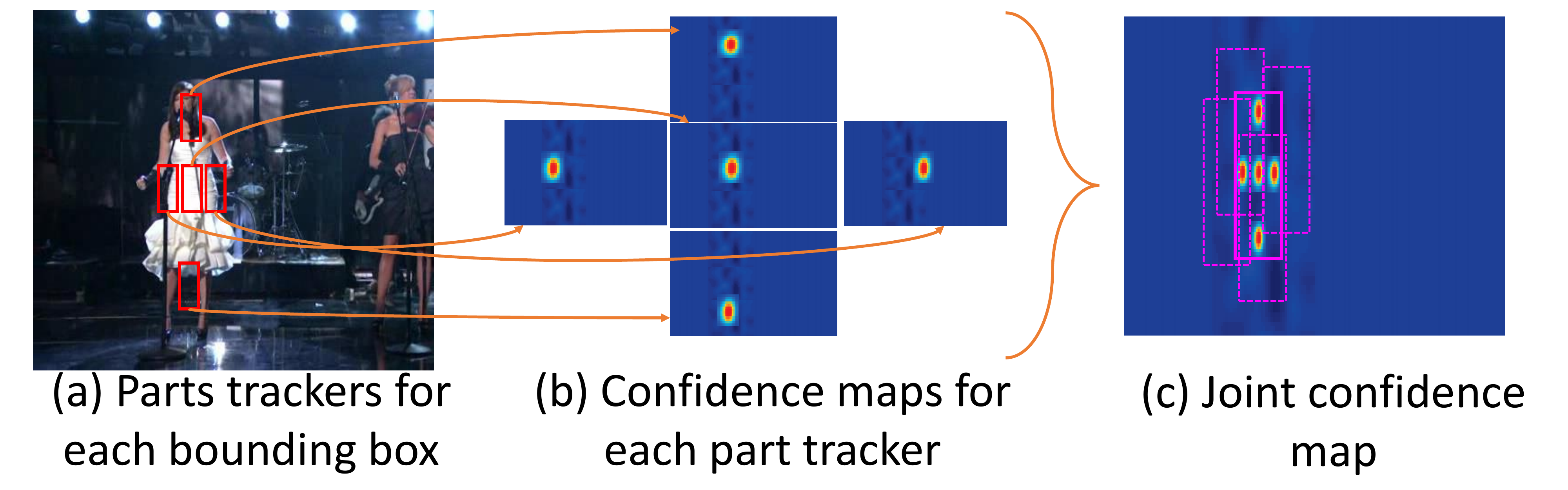}
    \end{center}
     \vspace{-16pt}
         \caption{Each part tracker computes response map  independently. Response maps are combined to get a joint confidence map as weighted sum of individual maps in the Bayesian framework. Purple rectangles represent the sample candidates while solid purple denote the maximum likelihood on the confidence map \cite{liu2015real}.}
\label{part_CFTs}
\vspace{-15pt}
\end{figure*}

Real time Part based tracking with Adaptive CFs (RPAC) \cite{liu2015real} adds a spatial constraint to each part of object as shown in Fig. \ref{part_CFTs}. In RPAC, KCF tracker is employed to track individually five parts of a target. Confidence score map for each part is computed by assigning adaptive weights during tracking for every new input frame. A joint map is constructed  by assigning adaptive weights to  each five confidence score maps and a new target position is estimated using particle filter method. 
During  tracking, adaptive weights or confidence scores for each part are calculated by computing sharpness of response map and Smooth Constraint of Confidence Map (SCCM). Response sharpness is calculated using Peak-to-Side-lobe Ratio (PSR),  while SCCM is defined by the spatial shift of a part between two consecutive frames. Adaptive part trackers are updated for parts with weights  higher then a threshold value. A Bayesian inference theorem is employed to compute the target position by calculating the Maximum A Posteriori (MAP) for all  parts. 

The Reliable Patch Tracker (RPT) \cite{li2015reliable} is based on particle filter framework which apply KCF as base tracker for each particle, and exploits local context by tracking the target with reliable patches. During tracking, the weight for each reliable patch is calculated based on whether it is a trackable patch, and whether it is a patch with target properties. The PSR score is used to identify patches, while motion information is exploited for probability that a patch is on target. Foreground and background particles are tracked along with relative trajectories of particles. A patch is discarded if it is no longer reliable, and re-sampled to estimate a new reliable patch. A new target position is estimated by Hough Voting-like strategy by obtaining all the weighted, trackable, and reliable positive patches.
Recurrently Target attending Tracking (RTT) \cite{cui2016recurrently} learns the model by discovering and exploiting the reliable spatial-temporal parts using DCF. Quard-directional RNNs are employed to identify reliable parts from different angles as long-range contextual cues. Confidence maps from RNNs are used to weight adaptive CFs during tracking to suppress the negative effects of background clutter. 
Patch based KCF (PKCF) \cite{chen2016robust} is a particle filter framework to train target patches using KCF as base tracker. 
Adaptive weights as confidence measure for all parts based on the PSR score are computed.  For every incoming frame, responses for each template patch are computed. The PSR for each patch is computed, and maximum weighted particles are selected. 

The Enhanced Structural Correlation (ESC) tracker \cite{chen2017visual} exploits holistic and object parts information. The target is estimated based on weighted responses from non-occluded parts. Colour histogram model, based on Bayes' classifier is used to suppress background by giving higher probability to objects. The background context is enhanced from four different directions, and is considered for the histogram model of the object's surroundings. The enhanced image is broke down into patches (one holistic and four local) and CF is employed to all patches. The CF is employed to all image patches and final responses are obtained from the weighted response of the filters. Weight as a confidence score for each part is measured from the object likelihood map and the maximum responses of the patch. Adaptive CFs are updated for those patches whose confidence score exceeds a threshold value. Histogram model for object are updated if the confidence score of object is greater then a threshold value,  while background histogram model is updated on each frame.
Zuo et al.\cite{liu2016structural} proposed Structural CF (SCF) to exploit the spatial structure among the parts of an object in its dual form. The position for each part is estimated at the maximum response from the filter response map.  Finally, the target is estimated based on the weighted average of translations for all parts.

\subsubsection{Fusion-based Correlation Filter Trackers}
\begin{figure}
     \vspace{-5pt}
  \begin{center}
    \includegraphics[width=0.5\textwidth]{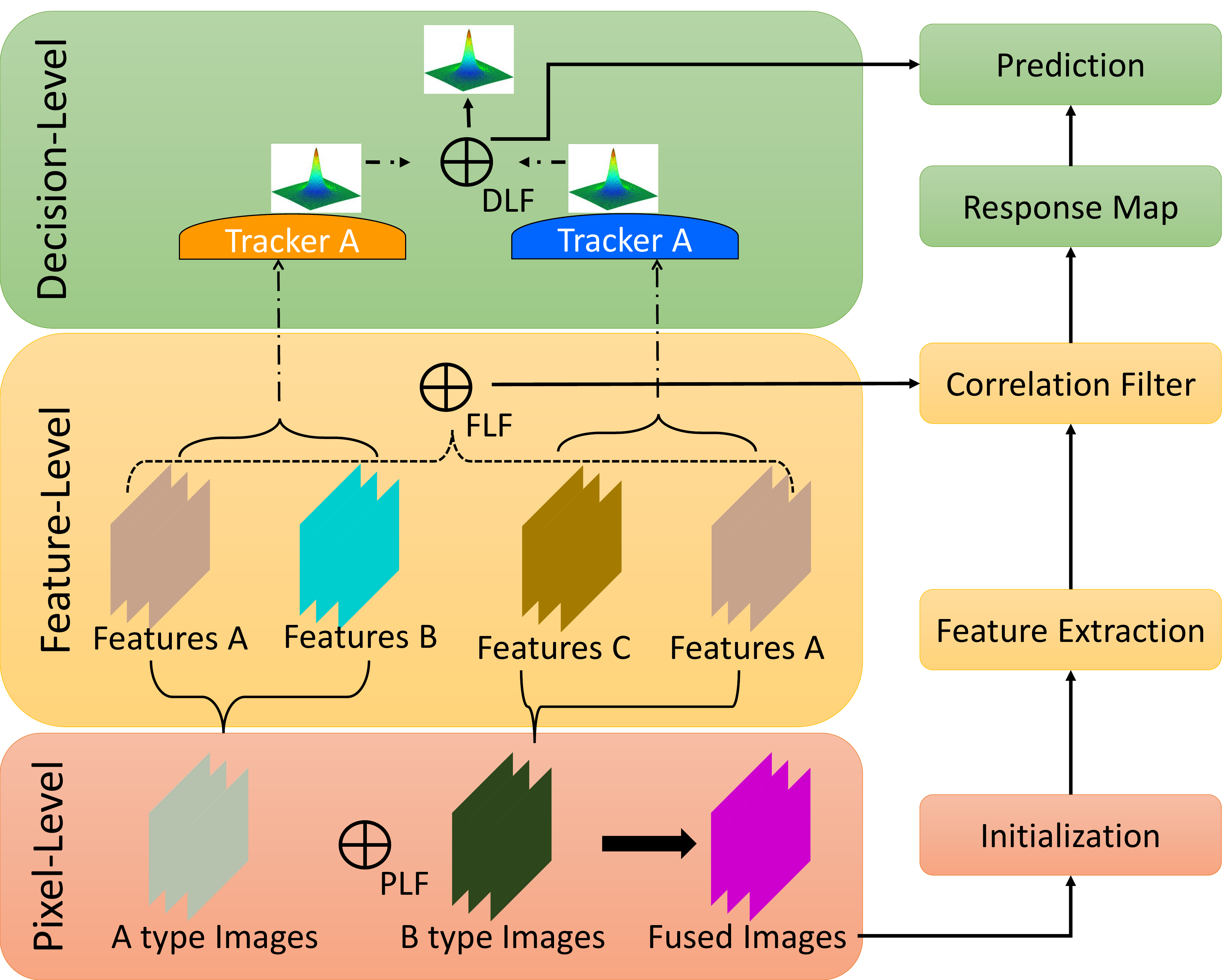}
    \end{center}
     \vspace{-14pt}
         \caption{Pixel-level, feature-level and decision level fusion-based tracking framework \cite{luo2018comparison}.}
\label{fusion_tracking}
\vspace{-15pt}
\end{figure}

In image fusion, complementary information is fused to improve performance in numerous applications including medical imaging \cite{blum2006multi}, face recognition \cite{chen2006fusion}, image segmentation \cite{wang2017unsupervised}, and image enhancement \cite{farid2019multi}.  Image fusion may be performed as Pixel-Level Fusion (PLF), Feature-Level Fusion (FLF), and Decision-Level Fusion (DLF) as shown in Fig. \ref{fusion_tracking}. Numerous trackers using different types of fusion have been developed including \cite{luo2018comparison, wang2017robust, rapuru2017correlation, ma2015multiple, bai2018kernel}.

Luo et al. \cite{luo2018comparison} used SAMF as a baseline tracker and performed those fusions. Authors used visual and infrared images for PLF, HOG, grey while colorname features are used for FLF.  Ma et al. \cite{ma2015multiple} proposed FLF tracker where raw pixels, color histogram and Haar features were employed. Authors divided the region of interest into sub-parts and evaluated the features individually, which are fused using  weighted entropy to exploit the complementary information. Target appearances are updated via  subspace for each kind of feature with new object samples. Wang et al. \cite{wang2017robust} introduced deep feature fusion technique using two networks, Local Detection Network (LDN) and Global Network Detection (DND). LND  employs VGG-16 and fuses the deep features extracted from conv4-3 and conv5-3 layers to generate response map. Feature map from conv5-3 is upsampled using a deconvolution layer and added to conv4-3 featuers.  If confidence score is less then a threshold then target is considered to be lost. GDN detects the target if LDN fails to track. It employs Region Proposal Network (RPN) after conv4-3. LDN is updated to integrate target variations while GDN parameters are fixed.

Rapuru et al. \cite{rapuru2017correlation} proposed Correlation based Tracker level Fusion (CTF) by integrating two complementary trackers, Tracking-Learning-Detection (TLD) \cite{kalal2012tracking} and KCF. TLD tracker has resumption ability and is composed of three basic units, tracking to predict the new target position;  target localization in current frame;  rectifying the detector error by learning different target variations. KCF tracker performs tracking by detection and uses non-linear regression kernels. KCF exploits the training and testing data in circulant structure that results in low computational cost. During tracking, if the output of TLD tracker is valid then output for KCF tracker is computed.  Conservative correspondence $C^{c}$ for both trackers is calculated as confidence score such that the first 50\% of positive patches has resemblance with the sample patches. Current bounding box (\textit{cbb}) is the output of that tracker which has maximum $C^{c}$ score. TLD calculates clusters of positive patches as TLD detector response (\textit{$dbb$}). Final bounding box (\textit{fbb}) is calculated  as: if overlap score between \textit{cbb} and \textit{$dbb$} is greater than a threshold, and relative similarity is also greater then a threshold, otherwise the target is considered lost. Relative similarity depicts the target confidence. The \textit{fbb} is computed as weighted mean of \textit{dbb} and \textit{cbb}. 

\subsection{Non-Correlation Filter Trackers (NCFT)}

We categorize all trackers which do not employ correlation filters as \textbf{Non-Correlation Filter based Trackers} (NCFTs). 
We categorize NCFTs into multiple categories including  patch learning, sparsity, superpixel, graph, multiple-instance-learning, part-based , and Siamese-based trackers. These trackers are discriminative except sparsity-based trackers which are generative trackers.

\subsubsection{Patch Learning Trackers}
\textbf{Patch learning trackers} exploit both target and background patches. A tracker is trained on positive and negative samples. The trained tracker is tested on number of samples, and the maximum response gives the target position. Several trackers have been proposed including \cite{nam2016mdnet, fan2016sanet, han2017branchout, zhang2016robust, gao2017robust, gao2016enhancement, cai2016bit, yun2017adnet, zhang2017deep, Zhang2017CVPR, teng2017robust, ning2016object, gao2017deep, li2016deeptrack}.

\begin{figure}
  \begin{minipage}[b]{.48\textwidth}
    \centering
    \includegraphics[width=\linewidth]{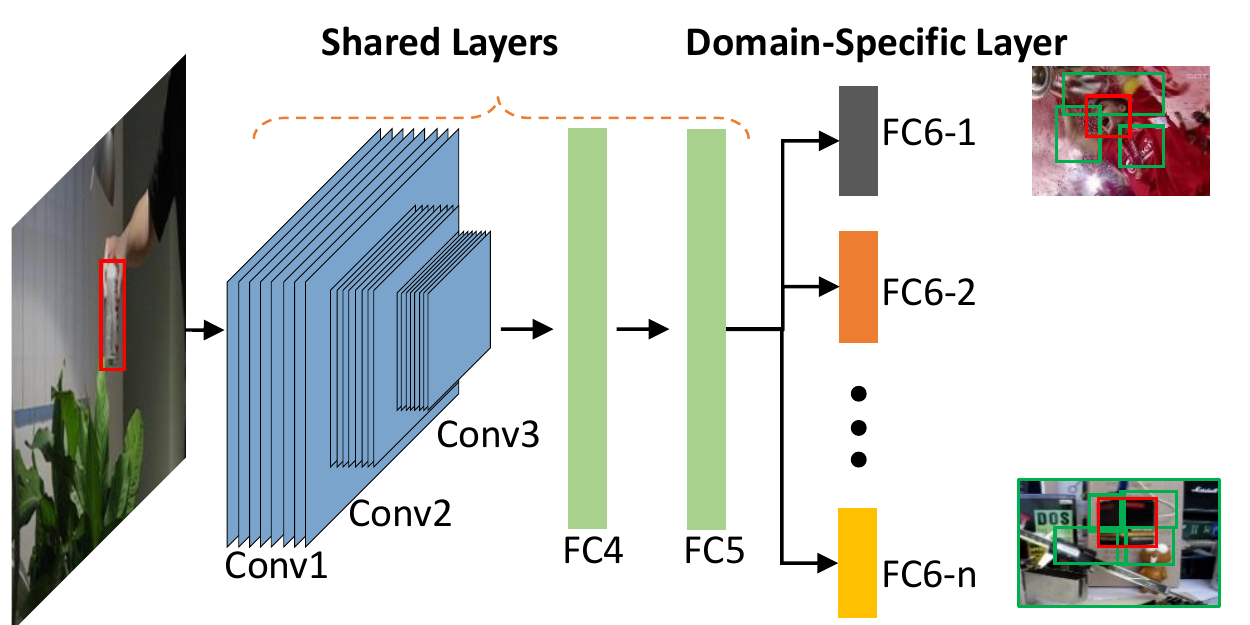}
    \vspace{-8mm}
          \caption{MDNet architecture \cite{nam2016mdnet}.}
     \label{mdnet_structure}
  \end{minipage}\hfill
  \begin{minipage}[b]{.48\textwidth}
    \centering
    \includegraphics[width=\linewidth]{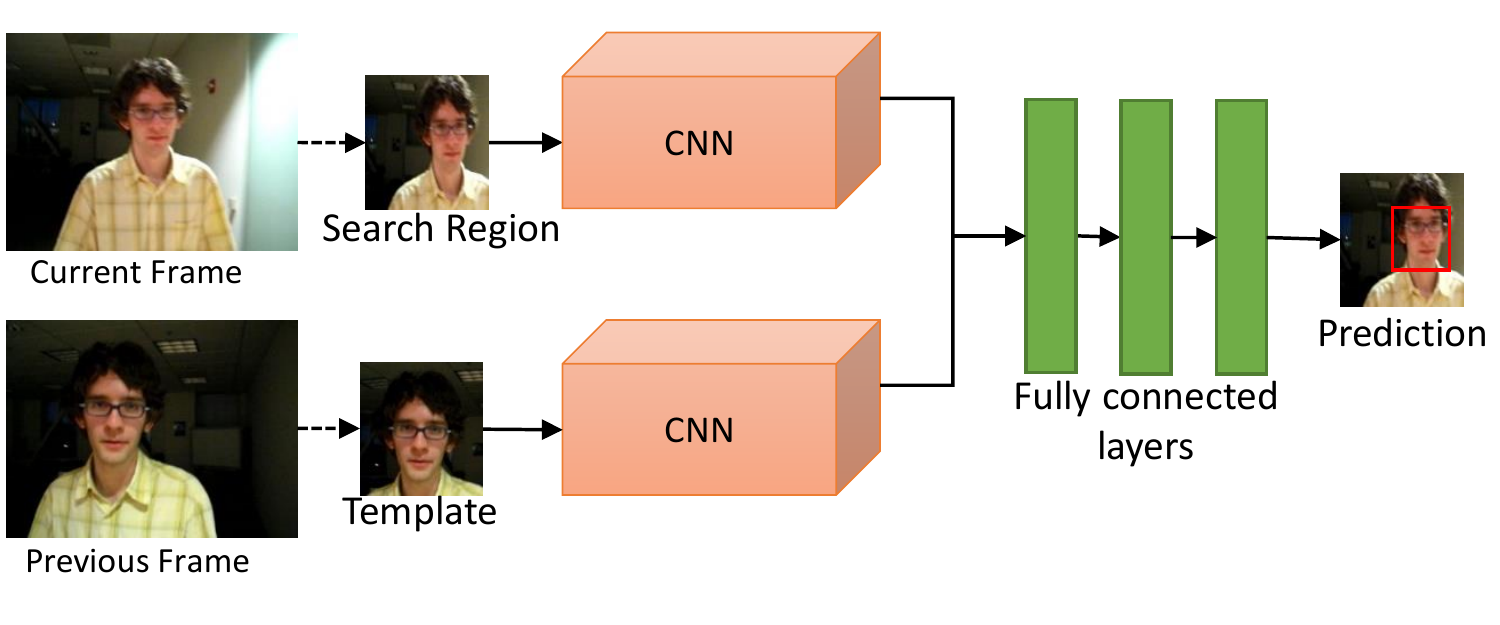}
    \vspace{-8mm}
    \caption{GOTURN tracking framework \cite{held2016learning}.}
     \label{GOTURN_framework}
  \end{minipage}
  \vspace{-6mm}
\end{figure}

A Multi-Domain Network (MDNet) \cite{nam2016mdnet}  consists of shared layers (three convolutional and two Fully-Connected (FC) layers) and one domain-specific FC layer as shown in Fig. \ref{mdnet_structure}. Shared layers exploit generic target representation from all the sequences, while domain specific layer is responsible for identification of target using binary classification for a specific sequence. During  online tracking, the domain specific layer is independently learned at the first frame. Samples are generated based on previous target location, and a maximum positive score yields the new target position. Weights of the three convolutional layers are fixed while  weights of three fully connected layers are updated for short- and long-term updates. Long-term update is performed after a fixed interval from positive samples. The short-term update is employed whenever tracking fails and the weight adaption for FC layers is performed using positive and negative samples from the current short term interval. Target position is adjusted by employing a bounding box regression  model \cite{girshick2014rich} in the subsequent frames.



A Structure Aware Network (SANet) \cite{fan2016sanet} exploits the target's structural information based on particle filter framework. CNN and RNN deep features are computed for particles. RNN encodes the structural information of the target using directed acyclic graphs. SANet is a modified version of MDNet, with the addition of RNN layers to amplify the rich object representation. Convolutional  and recurrent  features are fused using a skip concatenation strategy to encode the rich information. 
Han et al. \cite{han2017branchout} presented Branch-Out algorithm, which uses MDNet as a base tracker. The Branch-Out architecture comprises of three CNN and multiple FC layers as branches. Some branches consists of one fully-connected layer, while  others have two fully-connected layers. During  tracking, a random subset of branches is selected by Bernoulli distribution to learn target appearance.

Zhang et al. \cite{zhang2016robust} proposed Convolutional Networks without Training (CNT) tracker employ particle filter framework that exploits the inner geometry and local structural information of the target. The CNT algorithm is an adaptive algorithm in which appearance variation of target is adapted during the tracking. CNT employs a hierarchical architecture with two feed forward layers of convolutional network to generate an effective target representation. In the CNT, pre-processing is performed on each input image where image is warped and normalized. The normalized image is then densely sampled as overlapping local image patches of fixed size, also known as filters, in the first frame. After pre-processing, a feature map is generated from a bank of filters selected with k-mean algorithm. Each filter is convolved with normalized image patch, which is known as simple cell feature map. In second layer, called complex cell feature map, a global representation of target is formed by stacking simple cell feature map which encodes local as well as geometric layout information. 
 

Exemplar based Linear Discriminant Analysis (ELDA) \cite{gao2017robust} employs LDA to discriminate  target and background. ELDA takes several negative samples from the background and single positive sample at current target position. ELDA has object and background component models. The object model consists of  long-term  and  short-term  models. The target template at the first frame corresponds to  long-term model, while  short-term model corresponds to the target appearance in a short periods.  The background models comprises of  an online and offline background models. The online is built from negative samples around the target, while the offline background model is trained on large number of negative samples from natural images. 
The ELDA tracker is comprised of  short and long term detectors. Target location is estimated from the sum of long-term and  weighted sum of   short-term detection scores. 
ELDA has been enhanced by integration with CNN, and named as Enhanced CNN Tracker (ECT) \cite{gao2016enhancement}. 

The Biologically Inspired Tracker (BIT) \cite{cai2016bit} performs tracking like ventral stream processing. The  BIT  tracking framework consists of an appearance  and  tracking model.  The appearance model has two units, classical simple cells (S1) and cortical complex cells (C1). A S1 is responsible to exploiting colour and texture information, 
while a C1 performs pooling and combining of color and texture features to form complex cell. 
A two-layer tracking model is composed of generative and discriminative models: a view-tuned learning (S2) unit and a task dependent learning (C2) unit. 
Generative S2 unit computes response map  by performing convolution between the input features, and the target  and response maps are fused via average pooling. The discriminative C2 unit then computes new target position by applying CNN classifier.
An Action-Decision Network (ADNet) \cite{yun2017adnet} controls sequential actions (translation, scale changes, and stopping action) for tracking using deep RL. 
The ADNet is composed of three convolutional and three FC layers.
An ADNet is defined as an agent with the objective to find target bounding box. The agent is pretrained to make decision about target's movement from a defined set of actions. During  tracking, target is tracked based on estimated action from network at the current tracker location. Actions are repeatedly estimated by agent unless reliable target position is estimated. Under the RL, the agent gets rewarded when it succeeds in tracking the target, otherwise, it gets penalized. 


The Oblique Random forest (Obli-Raf) \cite{Zhang2017CVPR}  exploits geometric structure of the target. During  tracking, sample patches are drawn as particles and forwarded to an oblique random forest classifier, based on estimated target position on previous frame. Obli-Raf  generates the hyperplane from data particles in a semi-supervised manner to recursively cluster sample particles using proximal support vector machine. 
Particles are classified as target or background, based on votes at each leaf node of the tree. Particle with maximum score will be considered as newly-predicted target position. If the maximum votes limit is less then a predefined threshold, then a new particle samples set is produced from the estimated target location. If maximum limit is achieved, the model is updated otherwise the previous model is retained. 


Dual Linear Structured Support Vector Machine (SSVM) (DLSSVM) \cite{ning2016object} which is the motivation of Struck \cite{harestruck}. Usually, classifier is trained to discriminate target from background but Struck employs kernelized structured output of the SVM for adaptive tracking. DLSSVM uses dual SSVM linear kernels as discriminative classifier for explicit high dimensional features as compared to Struck. The difference between both trackers is the selection of optimization scheme to update dual coefficients. DLSSVM employs Dual Coordinate Descent (DCD) \cite{ramanan2013dual} optimization method to compute a closed form solution. Another difference is that in Struck, pair of dual variables are selected and optimized while DLSSVM selects only one dual variable. Scale DLSSVM (SDLSSVM) improves the tracker by incorporating multi-scale estimation.

\subsubsection{Multiple-Instance-Learning Based Trackers}

Multiple-Instance-Learning (MIL) was introduced by Dietterich and is widely used in many computer visions tasks where MIL is being used for example object detection \cite{zhang2006multiple}, face detection \cite{guillaumin2010multiple} and action recognition \cite{ali2010human}. Various researcher have employed MIL to track targets \cite{babenko2011robust, wang2017patch, yang2017visual, sharma2017mil, xu2015robust, abdechiri2017visual}.
In MIL based tracking, training samples are placed in bags instead of considering individual patches, and labels are given at bags level. Positive label is assigned to a bag if it has at-least one positive sample in it and on the other hand, negative bag contains all negative samples. Positive bag may contain positive and negative instances. During training in MIL, label for instances are unknown but bag labels are known. In the MIL tracking framework instances are used to construct weak classifiers and a few instances are selected and combined to form a strong classifier.

Babenko et al. \cite{babenko2011robust} designed a novel MILTrack to label ambiguity of instances using Haar features. MILBoost as baseline tracker is utilized which employs the gradient boosting algorithm to maximize the log likelihood of bags. A strong classifier is trained to detect a target by choosing  weak classifiers. A weak classifier is computed using log  odds ratio in a Gaussian distribution.  Bag probabilities are computed using a Noisy-OR model. 
Online boosting algorithm is employed to get new target position from weighted sum of weak classifiers.

Xu et al. \cite{xu2015robust} proposed an MIL framework that uses Fisher information using MILTrack (FMIL) to select weak classifiers.
Uncertainty is measured from unlabeled samples using Fisher information criterion instead of log likelihood. An online boosting method is employed for feature selection to maximize the Fisher information of the bag. Abdechiri et al. \cite{abdechiri2017visual} proposed Chaotic theory in MIL (CMIL). Chaotic representation exploits complex local and global target information. HOG and Distribution Fields (DF) features with optimal dimension are used for target representation.
Chaotic approximation is employed to enhance the discriminative ability of the classifier.
The significance of the instance is calculated using position distance and fractal dimensions of state space simultaneously. The appearance model known as chaotic model is learned to adapt dynamic of target through chaotic map to maximize likelihood of bags using. To encode chaotic information, state space is reconstructed by converting an image into a vector form and normalizing it with a zero mean and variance equivalent to one. Taken's embedding theory generate a multi-dimensional space map from one-dimension space. The minimum delay in time and prediction of the embedding dimension is performed by false nearest neighbours to reduce dimensionality for state space reconstruction. Finally, GMM is imposed to model state space.

Wang et al. \cite{wang2017patch} presented Patch based MIL (P-MIL) that decomposes the target into several blocks. The MIL for each block is applied, and the P-MIL generates strong classifiers for target blocks. The average classification score, from classification scores for each block, is used to detect whole target. Sharma and Mahapatra \cite{sharma2017mil} proposed a MIL tracker depends on maximizing the Classifier ScoRe (CSR) for feature selection. The tracker computes Haar-features for target with kernel trick, half target space, and scaling strategy. 

Yang et al.\cite{yang2017visual} used Mahalanobis distance to compute the instance significance to bag probability in a MIL framework, and employed gradient boosting to train classifiers. Instance are computed using a coarse-to-fine search technique during tracking. The Mahalanobis distance describes the importance between instances and bags. Discriminative weak classifiers are selected based on maximum margin between negative and positive bags by exploiting average gradient and average classifier strategy.

\subsubsection{Siamese Network Based NCFT Trackers}

Siamese network based NCFT perform tracking based on matching mechanism. The learning process exploits the general target appearance variations. Siamese network-based trackers match target templates with candidate samples to yield the similarities between patches. Various Siamese-based CFTs have been developed including \cite{held2016learning, tao2016siamese, wang2018sint++, chen2017once, choi2017visual}. 

Generic Object Tracking Using Regression Network (GOTURN) proposed by Held et al. \cite{held2016learning} exploits object appearance and motion relationships. During tracking, template and search regions are cropped at previous and current frames respectively, and those crops are padded with context information as shown in Fig. \ref{GOTURN_framework}. Target template and search regions are fed to five individual convolutional layers. Deep features from two separate flows are fused into shared three sequential fully-connected layers. GOTURN is a feed-forward offline tracker that does not require fine-tuning, and directly regresses target location.  


A Siamese INstance Search (SINT) \cite{tao2016siamese} performs tracking using offline learned matching function, and finds best-matched patch between target template and candidate patches in new frames  without updating matching function. The SINT architecture have two steams: a query stream and search stream. Each steam consists of five convolutional layers, three region-of-interest pooling layers, one FC layer, and a contrastive loss function layer which is responsible to discriminate target from background to fuse features. During tracking, target template as query from the initial frame is matched with candidate samples from each frame. The output bounding box is refined using four Ridge bounding box regression trained over the bounding box from the initial frame. 
Chen and Tao \cite{chen2017once} proposed  two flow CNN tracker called as YCNN that is learned end-to-end to calculate similarity map between the search region and the target patch using shallow and deep features. YCNN architecture has two flows: an object and search flow. Deep features obtained from object and search flows having three convolutional layers are concatenated, and are forwarded to two FC layers to yield prediction map. Maximum score in the prediction map infers the new target location. 


Reinforced Decision Making (RDM) \cite{choi2017visual} model is composed of a matching and a policy network. Prediction heatmaps are generated from the matching network,  while the policy network is responsible for producing normalized scores from prediction heatmaps. During tracking, a cropped search patch and along with $N$ target templates are forwarded to two separate convolutional layers and these deep features are fused using shared FC layers in matching networks to produce prediction maps. Using prediction map, policy network computes normalized scores. Prediction map gives estimated target at the maximum score. 
The policy network contains two convolutional and two FC layers that make decisions about a reliable state using RL.


\begin{figure}
  \begin{minipage}[b]{.48\textwidth}
    \centering
    \includegraphics[width=\textwidth]{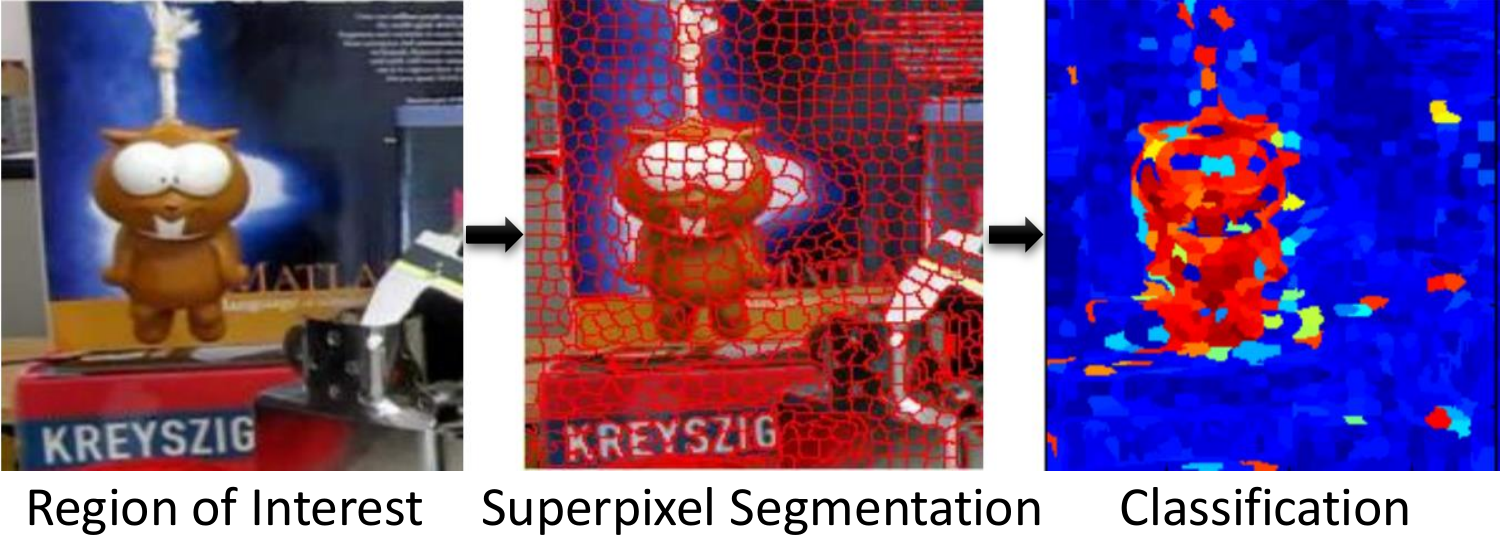}
    \vspace{-8mm}
    \caption{Superpixel classification.}
     \label{superpixel_classification}
  \end{minipage}\hfill
  \begin{minipage}[b]{.48\textwidth}
    \centering
    \includegraphics[width=\linewidth]{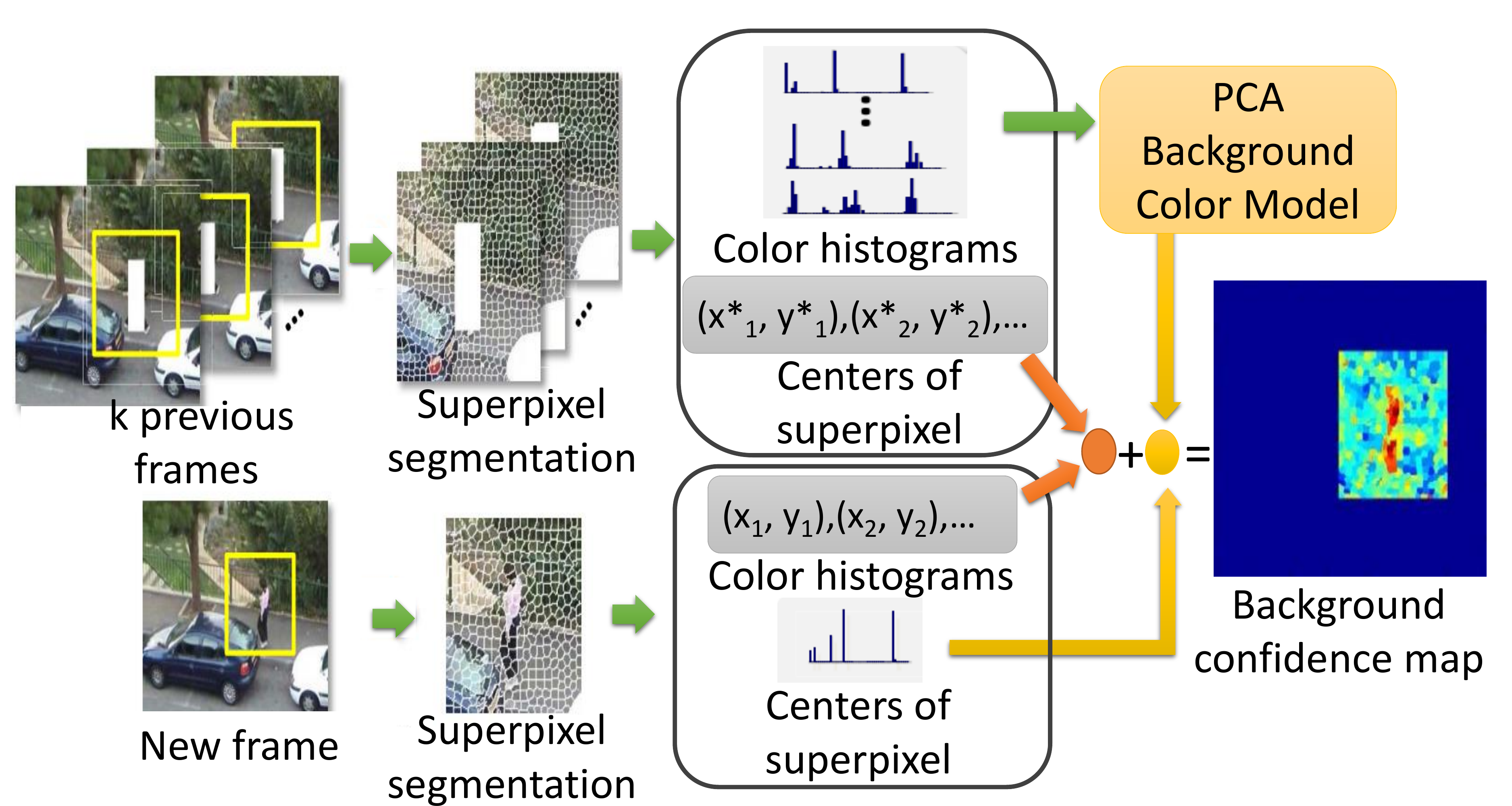}
    \vspace{-8mm}
    \caption{BKG flowchart \cite{li2014object}.}
    \label{BKG_framework}
  \end{minipage}
  \vspace{-15pt}
\end{figure}

\subsubsection{Superpixel Based Trackers}

Superpixels represent a group of pixels having identical pixel values \cite{achanta2012slic}. Region of interest is segmented into superpixels and classification is performed over superpixels for discrimination as shown in Fig. \ref{superpixel_classification}. A superpixel based representation got much attention by computer vision community for object recognition \cite{brekhna2017robustness}, human detection \cite{mori2004recovering}, activity recognition \cite{wu2007scalable}, and image segmentation \cite{ achanta2012slic}.
Numerous tracking algorithms have been developed using superpixels \cite{huang2017structural, wang2017constrained, wang2017two, yang2014robust, li2014object}.

Li et al. \cite{li2014object} used BacKGround (BKG) cues in a particle framework for tracking (Fig. \ref{BKG_framework}).  The background is segmented excluding target area for superpixels from previous $k$ frames. These  superpixels are representing the background. Superpixels for target are also computed in the current frame and compared with the background superpixels  using Euclidean distance and color histograms. Proposed scheme computes confidence map based on difference between the target and background. Current frame superpixels dissimilar to the background superpixels are considered as the target suerpixels.  


Yang et al. \cite{yang2014robust} also proposed a SuperPixel based Tracker (SPT).  Mean shift clustering is performed on superpixels to model target  and the background appearance. Similarity of superpixels in the current frame is computed from the target and the background models to find the target position.
The Constrained Superpixel Tracking (CST) \cite{wang2017constrained} algorithm employs graph labeling using superpixels as nodes and enforces spatial smoothness, temporal smoothness, and appearance fitness constraints. 
Spatial smoothness is enforced by exploiting the latent manifold structure using unlabeled and labeled superpixels. Optical flow is used for the temporal smoothness to impose short-term target appearance, while appearance fitness servers as long-term appearance model to enforce objectness.
Wang et al. \cite{wang2017two} presented a Bayesian tracking method at coarse-level and fine-level superpixel appearance model. 
The coarse-level appearance model computes few superpixels such that there is only one superpixel in the target bounding box, and a confidence measure defines whether that superpixel corresponds to background/target.  The fine-level appearance model calculates more superpixels in the target region based on target location in the previous frame and Jaccard distance is used to find fine-superpixels belonging to the target in the current frame. 
The Structural Superpixel Descriptor (SSD) \cite{huang2017structural} exploits the structural information via superpixels and preserves the intrinsic target properties. It decomposes a target into a hierarchy of different sized superpixels and assigns greater weights to superpixels closer to the target center. A particle filter framework is used and background information is alleviated through adaptive patch weighting. 


\begin{figure}
  \begin{minipage}[b]{.48\textwidth}
    \centering
    \includegraphics[width=0.6\linewidth]{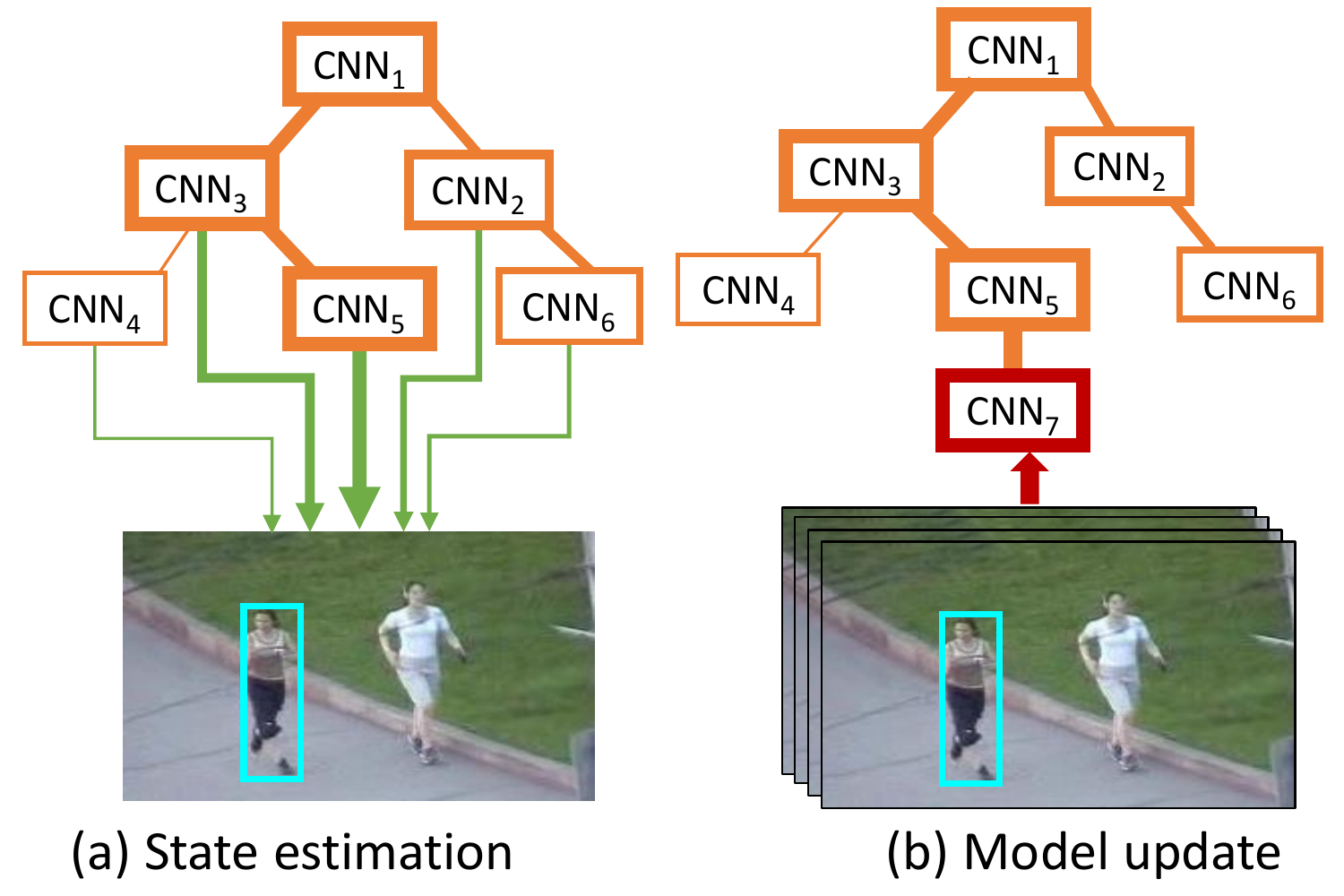}
    \vspace{-5mm}
          \caption{Visualization of (a) target state estimation and  (b)  model update in Tree Structure CNN \cite{nam2016modeling}. CNN weights  are shown using green arrows width for state estimation. Affinity among CNNs is shown by the width of orange edge for model update while reliability of CNN is indicated by the width of CNN box.}
     \label{tCNN_framework}
  \end{minipage}\hfill
  \begin{minipage}[b]{.48\textwidth}
    \centering
    \includegraphics[width=\linewidth]{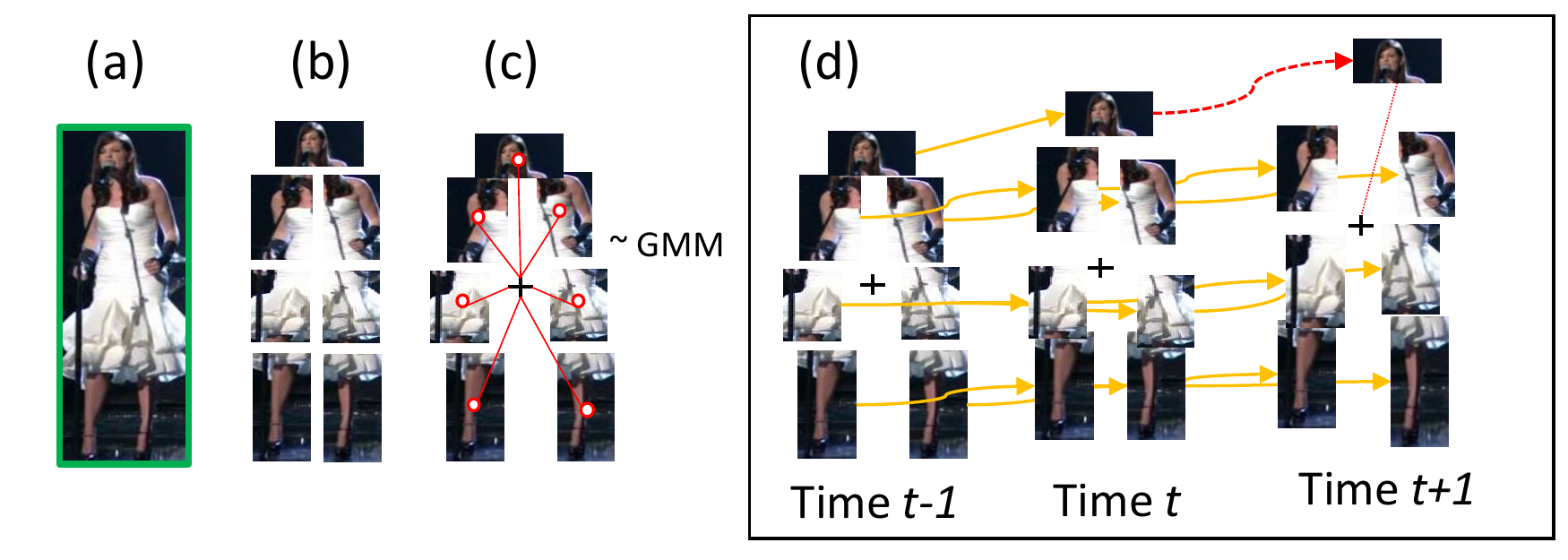}
    \vspace{-8mm}
    \caption{ALMM tracking approach  \cite{zhang2017adaptive}. (a) Target detection. (b)Target patches extracted from input image. (c) Target patches to follow GMM based on distance of  patches from center of gravity. (d) ALMM prunes out patches to make robust estimation.}
     \label{ALMM_approach}
  \end{minipage}
  \vspace{-7mm}
\end{figure}

\subsubsection{Graph Based Trackers}

A graph has  vertices (which may be pixels, superpixels or object parts) and edges (correspondence among vertices). Graphs are used to predict the labels of unlabeled vertices.
Graph based algorithms have been successfully used for object detection \cite{filali2016multi}, human activity recognition \cite{li2016multiview}, and face recognition \cite{meena2017improved}. Generally, graph-based trackers use superpixels as nodes to represent the object  appearance, while edges  represent the inner geometric structure. Another strategy  is to construct graphs among the parts of objects in different frames. Many trackers  have been developed using graphs \cite{wang2018gracker, du2016online, du2017geometric, nam2016modeling, yeo2017superpixel}.

The Tree structure CNN (TCNN) \cite{nam2016modeling} tracker employed CNN to model target appearance in tree structure. 
Multiple hierarchical CNN-based target appearance models are used to build a tree where vertices are CNNs and edges are relations among CNNs (Fig. \ref{tCNN_framework}). Each path maintains a separate history for target appearance in an interval. During tracking, candidate samples are cropped at target location estimated in the last frame.
Weighted average scores computed using multiple CNNs are used to calculate abjectness for each sample. Reliable patch along the CNN defines the weight of CNN in the tree structure. The maximum score from multiple CNNs is used to estimate target location.
Bounding box regression methodology is also utilized to enhance the estimated target position in the subsequent frames.

Du et al. \cite{du2016online} proposed Structure Aware Tracker (SAT) that constructs hypergraphs in temporal domain to exploit higher order dependencies. SAT gathers candidate parts in frame buffer from each frame by computing superpixels. A graph cut algorithm is employed to minimize the energy to produce the candidate parts. A Structure-aware hyper graph is constructed using candidate parts as nodes, while hyper edges denote relationship among parts. 
Object parts across multiple frames contribute to build a subgraph by grouping superpixels considering both appearance and motion consistency. Finally, the target location and boundary is estimated by combining all the target parts using coarse-to-fine strategy. 
Graph tracker (Gracker) \cite{wang2018gracker} uses undirected graphs to model planar objects and exploits the relationship between local parts. Search region is divided into grids, and a graph is constructed where vertices represent cells with  maximum  SIFT response  and edges are constructed using Delaunay triangulation. During  tracking, geometric graph-matching is performed to explore optimal correspondence between graph models and the candidate graph. 

A Geometric hyperGraph Tracker (GGT) \cite{du2017geometric} constructs geometric hpyergraphs by exploiting higher order geometric relationships among target parts. Target parts in previous frame are matched with candidate parts in new frame. The relationship between target and candidate parts is represented by correspondence hypothesis. A geometric hypergraph is constructed from the superpixels where vertices are correspondence hypothesis  while edges constitute the geometric relationship within the hypothesis.
Reliable parts are computed from correspondence hypotheses learned from the matched target and candidate part sets. During  tracking, reliable parts are extracted with high confidence to predict target location.  An Absorbing Markov Chain Tracker (AMCT) \cite{yeo2017superpixel} recursively propagates  the predicted segmented target in subsequent frames.  AMCT has two states: an absorbing and a transient state. In an AMCT, any state can be entered to absorbing state, and once entered, cannot leave, while other states are transient states. An graph is constructed between two consecutive frames based on superpixels, where vertices are background superpixels (represents absorbing states) and target superpixels (transient states). Edges weights are learned from SVM to distinguish foreground and background superpixels. Motion information is imposed by spatial proximity using inter-frame edges.
The target is estimated from the superpixel components after vertices have been evaluated against the absorption time threshold. 






\subsubsection{Part-Based NCFT Trackers}

Part-based modeling have been activity used in NCFTs to handle deformable parts. Various techniques are employed to perform object detection \cite{ouyang2017deepid}, action recognition \cite{du2016representation}, and face recognition \cite{zhang2016face} using parts. Object local parts are utilized to model a tracker \cite{zhang2017adaptive, yao2017part, wang2016part, li2017visual, gao2018p2t}.

Adaptive Local Movement Modeling (ALMM) \cite{zhang2017adaptive}  exploits  intrinsic local movements of object parts for tracking. Image parts are estimated using a base tracker such as Struck \cite{hare2016struck}, and  GMM is used to prune out drifted parts.
GMM is employed to model the parts movement based on displacement of centers from the global object center (Fig. \ref{ALMM_approach}). A weight is assigned to each part, based on motion and appearance for better estimation. The target position is estimated from a strong tracker by combining all parts trackers in a boosting framework. 

Yao et al. \cite{yao2017part} proposed a Part based Tracker (PT) where object is decomposed into parts and an adaptive weight is assigned to each part. A structural spatial constraint is  applied to each part using minimum spanning tree where vertices represents parts and edges define  consistent connections. A weight is assigned to each edge corresponding to Euclidean distance between two parts. Online structured learning using SVM is performed to distinguish target and its parts from background. 
During tracking, the maximum classification scores of target and parts is used to estimate the new target position.

Li et al. \cite{li2017visual} used local covariance descriptors as target appearance and exploited the relationship among parts. The target is divided into non overlapping parts. A pyramid is constructed having  multiple local covariance descriptors that are fused using max pooling depicting target appearance. Parts are modeled using star graph and central part of target representing central node.  During tracking,  target parts  are selected from candidate part pools and template parts by solving a linear programming problem. Target is estimated from selected parts using a weighted voting mechanism based on relationship between center part and surrounding parts.
Part-based Multi-Graph Ranking Tracker (PMGRT) \cite{wang2016part}  constructs graphs to rank target parts. During tracking, target is divided into parts and different features are extracted for each target part. Multiple graphs are constructed based on both target parts and feature types, where one graphs is from one feature type. An affinity weight matrix is formed where rows represent graphs for different features and columns denotes the graphs of various parts.  Augmented Lagrangian formulation is optimized to select parts associated with high confidence.


\subsubsection{Sparsity Based Trackers}

All algorithms studied so far are discriminative tracking methods. On the other hand, Generative methods  learn target representation and search target in each frame  with minimal reconstruction error \cite{kwon2010visual}.  Sparse representation is a good example for generative models.  Sparse representations are widely used in computer vision, signal processing, and image processing communities for numerous applications such as face recognition \cite{lou2016graph}, object detection \cite{peng2017salient}, and image classification \cite{romero2016unsupervised}.
The objective is to discover an optimal representation of the target which is sufficiently sparse and minimizes the reconstruction error. 
Mostly sparse coding is performed by first learning a dictionary. Assume $\textbf{X} =[x_1,...,x_N] \in \mathcal{R}^{m \times n}$ represents  gray scale images $x_i \in \mathcal{R}^m$.  A dictionary $\textbf{D}=[d_1,...,d_k] \in \mathcal{R}^{m \times k}$ is learned on $\textbf{X}$ such that each image in \textbf{X} can be sparsely represented by a linear combination of items in \textbf{D}: $x_i=\textbf{D}\alpha_i$, where $\alpha_i=[\alpha_1,...,\alpha_k] \in \mathcal{R}^{k}$ denotes the spares coefficients. 
When $k>r$, where $r$ is the rank of $\textbf{X}$, then dictionary $\textbf{D}$ is overcomplete. For a known $\textbf{D}$, a constrained minimization using $\ell_1-$norm is often applied to find $\alpha$ for sufficiently sparse solution: 

\vspace{-10pt}
\begin{equation}\label{SC:3}
\vspace{-2pt}
\alpha_i^* \equiv \underset{\alpha_i}{\text{arg min}}\frac{1}{2}\parallel x_i-\textbf{D}\alpha_i \parallel_2^2 + \lambda \parallel \alpha_i \parallel_1,
\end{equation}
where $\lambda$ gives relative weights to the sparsity and reconstruction error. Dictionary \textbf{D} is learned in such  a way that all images in $\textbf{X}$ can be sparsely represented with a small error.  Dictionary $\textbf{D}$ is learned to solve the following optimization problem:
\vspace{-10pt}
\begin{equation}\label{SC:4}
\vspace{-2pt}
\{\alpha^*, \textbf{D}^*\} \equiv \underset{\textbf{D},\alpha}{\text{minimize}}  \sum_{i=1}^N \parallel \textbf{X}-\textbf{D}\alpha \parallel_2^2 + \lambda \parallel \alpha \parallel_1,
\end{equation}
There are two alternative phases for dictionary learning. \textbf{D} is assumed to be fixed and the coefficients $\alpha$ are calculated in the initial phase. 
While in the second phase, dictionary $\textbf{D}$ is updated while $\alpha$ is assumed to be fixed.  In visual object tracking, the objective of dictionary learning is to perform discrimination between  target and background patches by sparsely encoding target and background coefficients. Various sparsity based trackers have been proposed including \cite{zhang2015structural, zhang2018robust, guo2017visual, yi2017visual, zhangCEST2016robust, Zhang_IJCV15}. 




\begin{figure}
     \vspace{-5pt}
  \begin{center}
    \includegraphics[width=0.7\textwidth]{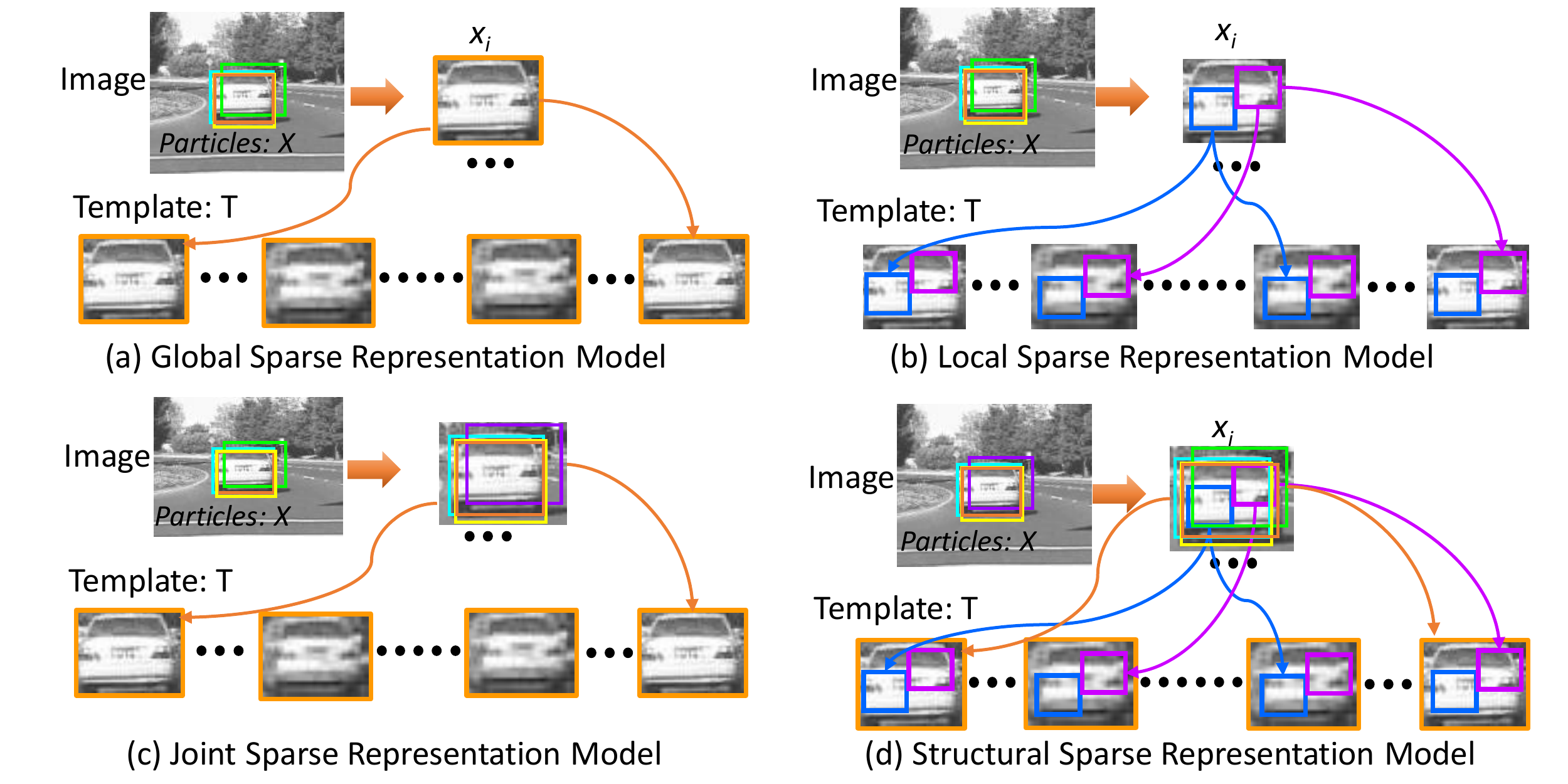}
    \end{center}
     \vspace{-16pt}
         \caption{(a) global sparse representation \cite{li2011real}, (b) local sparse representation model \cite{jia2012visual}, (c) joint sparse representation model \cite{zhang2012robust} and (d) structural sparse representation model \cite{zhang2015structural}.}
\label{sparse_representation}
\vspace{-15pt}
\end{figure}

Structural sparse tracking (SST) \cite{zhang2015structural}  is based on particle filter framework which exploits intrinsic relationship of local target patches and global target to jointly learn sparse representation.
Fig. \ref{sparse_representation} presents different types of sparse representations of target.
Fig \ref{sparse_representation} (a) shows global sparse representation which exploits holistic representation of the target and is optimized using $\ell_1$ minimization \cite{li2011real}. Local sparse representation models present the target patches  sparsely in  local patch dictionary as shown in Fig. \ref{sparse_representation} (b) \cite{jia2012visual}.  The joint sparse representation exploits the intrinsic correspondence among particles to jointly learn the dictionary \cite{zhang2012robust} (Fig. \ref{sparse_representation} (c)). Joint sparsity enforces particles to be jointly sparse and share the same dictionary template. 
SST exploits all the three sparse representations in a unified form as shown in Fig. \ref{sparse_representation} (d). SST estimates target from target dictionary templates and corresponding patches having  maximum similarity score from all the particles by preserving the spatial layout structure. The model is constructed on a number of particles representing target, and each target representation is decomposed into patches, and dictionary is learned. The patch coefficient is learned such that it minimizes the patch reconstruction error. SST considers that the same local patches from all the particles are similar. But this is not the case for tracking because usually outliers exist. Another problem in SST is that some local patches can select different target templates due to noise or occlusion. Zhang et al. \cite{zhang2018robust} improved SST and proposed Robust Structural Sparse Tracking (RSST) to exploit the shared correspondence by the local patches and also modeled the outliers because of noise and occlusion.

Context aware Exclusive Sparse Tracker (CEST) \cite{zhangCEST2016robust} exploits context information utilizing particle filter framework. The CEST performs linear combination of dictionary elements to represent particles. Dictionary is modeled as groups containing  target, occlusion and noise, and context templates.  In particle framework, new target is estimated as best particle from target template dictionary.
Guo et al. \cite{guo2017visual} computed  weight maps to represent target and background structure. 
A reliable structural constraint is imposed using the weight maps by penalizing the occluded target pixels.
Using  a Bayesian filtering framework, target is estimated using maximum likelihood from the estimated object state for all the particles.
Yi et al. \cite{yi2017visual} proposed Hierarchical Sparse Tracker (HST) to integrate the discriminative and generative models. The proposed appearance model is comprised of Local Histogram Model (LHM), Sparsity based Discriminant Model (SDM), and Weighted Aligment Pooling (WAP).
LHM encodes the spatial information among  target parts while the WAP assigns weights to local patches based on similarities between target and candidates. The target template sparse representation is computed in SDM. Finally, candidate with the maximum score from LHM, WAP, and SDM determines the new target position.

In this section we have studied CFTs and NCFTs  and elaborated different tracking frameworks and classified trackers into different subcategories. We summarized the characteristics of important well-cited trackers  from each subcategory in Table \ref{trackers_characteristics}.

\begin{table}[]
\caption{CFTs and NCFTs, and their characteristics.}
\vspace{-12pt}
\label{trackers_characteristics}
  \scalebox{0.4}{
  \begin{tabular}{|c|c|llcccccllr|}
\hline
\text{\textbf{Category}} & \text{\textbf{Subcategory}} & \multicolumn{1}{c|}{\textbf{Tracker}}   & \multicolumn{1}{c|}{\textbf{Features}}             & \multicolumn{1}{c|}{\textbf{Baseline tracker}} & \multicolumn{1}{c|}{\textbf{Scale estimation}} & \multicolumn{1}{c|}{\textbf{Offline training}} & \multicolumn{1}{c|}{\textbf{Online learning}} & \multicolumn{1}{c|}{\textbf{FPS}}  & \multicolumn{1}{c|}{\textbf{Benchmarks}}                         & \multicolumn{1}{c|}{\textbf{Publication}}    & \multicolumn{1}{c|}{\textbf{year}} \\ \hline
\multirow{23}{*}{CFTs}             & \multirow{6}{*}{Basic}                & KCF                            & HOG, Raw pixels                           &                                       & No                                    & No                                    & Yes                                  & 172                       & OTB2013                                                 & TPAMI                               & 2015                      \\
                                   &                                       & CF2                            & CNN                                       & KCF                                   & No                                    & No                                    & Yes                                  & 10.4                      & OTB2013,OTB2015                                         & ICCV                                & 2015                      \\
                                   &                                       & HDT                            & CNN                                       & KCF                                   & No                                    & No                                    & Yes                                  & 10                        & OTB2013,OTB2015                                         & CVPR                                & 2016                      \\
                                   &                                       & LCT                            & HOG                                       & KCF                                   & Yes                                   & No                                    & Yes                                  & 27.4                      & OTB2013                                                 & CVPR                                & 2015                      \\
                                   &                                       & fDSST                          & HOG                                       & DCF                                      & Yes                                   & No                                    & Yes                                  & 54.3                      & OTB2013, VOT2014                                        & TPAMI                               & 2015                      \\ 
                                   &                                       & \multicolumn{1}{l}{Staple}    & \multicolumn{1}{l}{HOG, Color Histogram} & \multicolumn{1}{c}{KCF}                 & \multicolumn{1}{c}{Yes}              & \multicolumn{1}{c}{No}               & \multicolumn{1}{c}{Yes}             & \multicolumn{1}{c}{80}   & \multicolumn{1}{l}{OTB2013, VOT14, VOT2015}            & \multicolumn{1}{l}{CVPR}           & 2016                      \\ \cline{2-12} 
                                   & \multirow{6}{*}{R-CFTs}               & SRDCF                          & HOG, CN                                   &                                       & No                                    & No                                    & Yes                                  & 5                         & OTB-2013, OTB-2015, ALOV++ and VOT2014                  & ICCV                                & 2015                      \\
                                   &                                       & CCOT                           & CNN                                       &                                       & Yes                                   & No                                    & Yes                                  & \multicolumn{1}{r}{0.3}   & OTB2015, Temple-Color, VOT2015                          & ECCV                                & 2016                      \\
                                   &                                       & ECO                            & CNN, HOG, CN                              & CCOT                                  & Yes                                   & No                                    & Yes                                  & 8                         & VOT2016, UAV123, OTB2015, TempleColor                   & CVPR                                & 2017                      \\
                                   &                                       & CSRDCF                         & HOG, Colornames,  color histogram         &                                       & Yes                                   & No                                    & Yes                                  & 13                        & OTB2015, VOT2015, VOT2016                               & CVPR                                & 2017                      \\
                                   &                                       & CACF                           & HOG                                       & SAMF                                  & Yes                                   & No                                    & Yes                                  & 13                        & OTB2013, OTB2015                                        & CVPR                                & 2017                      \\
                                   &                                       & \multicolumn{1}{l}{SCT}       & \multicolumn{1}{l}{HOG, RGB, colornames} & \multicolumn{1}{c}{KCF}              & \multicolumn{1}{c}{No}               & \multicolumn{1}{c}{No}               & \multicolumn{1}{c}{Yes}             & \multicolumn{1}{c}{40}   & \multicolumn{1}{l}{OTB2013}                            & \multicolumn{1}{l}{CVPR}           & 2016                      \\ \cline{2-12} 
                                   & \multirow{4}{*}{Siamese-based}        & SiameseFC                      & CNN                                       &                                       & Yes                                   & Yes                                   & No                                   & 58                        & OTB2013, VOT2014, VOT2015, VOT2016                      & ECCV                                & 2016                      \\
                                   &                                       & CFNet                          & CNN                                       & SiameseFC                             & Yes                                   & Yes                                   & Yes                                  & 43                        & OTB2013, OTB2015,  VOT2014, VOT2016, TempleColor        & CVPR                                & 2017                      \\
                                   &                                       & Dsiam                          & CNN                                       &                                       & Yes                                   & Yes                                   & Yes                                  & 45                        & OTB2013, VOT2015                                        & ICCV                                & 2017                      \\ 
                                   &                                       & \multicolumn{1}{l}{EAST}      & \multicolumn{1}{l}{HOG, raw pixels, CNN} & \multicolumn{1}{c}{}                 & \multicolumn{1}{c}{Yes}              & \multicolumn{1}{c}{Yes}              & \multicolumn{1}{c}{Yes}             & \multicolumn{1}{c}{23}   & \multicolumn{1}{l}{OTB2013. OTB2015, VOT2014, VOT2015} & \multicolumn{1}{l}{ICCV}           & 2017                      \\ \cline{2-12} 
                                   & \multirow{4}{*}{Part-based}           & RPAC                           & HOG                                       & KCF                                   & Yes                                   & No                                    & Yes                                  & 30                        & selected Sequences                                      & CVPR                                & 2015                      \\
                                   &                                       & RPT                            & HOG                                       & KCF                                   & No                                    & No                                    & Yes                                  & 4                         & OTB2013 and 10 selected sequences                       & CVPR                                & 2015                      \\
                                   &                                       & RTT                            & HOG, RNN                                  &                                       & Yes                                   & Yes                                   & Yes                                  & 3.5                       & OTB2013                                                 & CVPR                                & 2016                      \\ 
                                   &                                       & \multicolumn{1}{l}{PKCF}      & \multicolumn{1}{l}{HOG, CN}              & \multicolumn{1}{c}{KCF}              & \multicolumn{1}{c}{Yes}              & \multicolumn{1}{c}{No}               & \multicolumn{1}{c}{Yes}             & \multicolumn{1}{c}{0.5}  & \multicolumn{1}{l}{OTB2013}                            & \multicolumn{1}{l}{Neurocomputing} & 2016                      \\ \cline{2-12} 
                                   & \multirow{3}{*}{Fusion-based}         & Ma's                           & pixels, color histogram, Haar features    &                                       & Yes                                   & No                                    & Yes                                  & 0.4                       & OTB2013                                                 & ICCV                                & 2015                      \\
                                   &                                       & Wang's                         & CNN                                       &                                       & Yes                                   & Yes                                   & Yes                                  &                           & 12 Selected sequences                                   & ICASSP                              & 2017                      \\ 
                                   &                                       & \multicolumn{1}{l}{CTF}       & \multicolumn{1}{l}{HOG, binary features} & \multicolumn{1}{c}{KCF, TLD}         & \multicolumn{1}{c}{Yes}              & \multicolumn{1}{c}{No}               & \multicolumn{1}{c}{Yes}             & \multicolumn{1}{c}{25}   & \multicolumn{1}{l}{ALOV300++, OTB2013, VOT2015}        & \multicolumn{1}{l}{TIP}            & 2017                      \\ \hline
\multirow{23}{*}{Non-CFTs}         & \multirow{8}{*}{Patch learning}       & MDNet                          & CNN                                       &                                       & Yes                                   & Yes                                   & Yes                                  & 1                         & OTB2013, OTB2015, VOT2014                               & CVPR                                & 2016                      \\
                                   &                                       & SANET                          & CNN, RNN                                  & MDNet                                 & Yes                                   & Yes                                   & Yes                                  & 1                         & OTB205, TempleColor, VOT2015                            & CVPR                                & 2017                      \\
                                   &                                       & CNT                            & raw pixels                                &                                       & No                                    & No                                    & Yes                                  & 1.5                       & OTB2013                                                 & TIP                                 & 2016                      \\
                                   &                                       & ADNet                          & CNN                                       &                                       & Yes                                   & Yes                                   & Yes                                  & 3                         & OTB2013, OTB2015, VOT2013, VOT2014, VOT2015, ALOV300++  & CVPR                                & 2017                      \\
                                   &                                       & DRLT                           & CNN, RNN                                  &                                       & Yes                                   & Yes                                   & No                                   & 45                        & OTB2015                                                 & arXiv                               & 2017                      \\
                                   &                                       & Obli-Raf                       & CNN, HOG                                  &                                       & Yes                                   & No                                    & Yes                                  & 2                         & OTB2013, OTB2015                                        & CVPR                                & 2017                      \\
                                   &                                       & SDLSSVM                        & LRT (low Rant Transform)                  & Struck                                & Yes                                   & No                                    & Yes                                  & 5.4                       & OTB2013, OTB2015                                        & CVPR                                & 2016                      \\ 
                                   &                                       & \multicolumn{1}{l}{DeepTrack} & \multicolumn{1}{l}{CNN}                  & \multicolumn{1}{c}{}                 & \multicolumn{1}{c}{Yes}              & \multicolumn{1}{c}{Yes}              & \multicolumn{1}{c}{Yes}             & \multicolumn{1}{c}{2.5}  & \multicolumn{1}{l}{OTB2013, VOT2013}                   & \multicolumn{1}{l}{TIP}            & 2016                      \\ \cline{2-12} 
                                   & \multirow{3}{*}{MIL}                  &  MILTrack                              &    Haar                                       &   MILBoost                                    &   Yes                                    &     No                                  &     Yes                                 &    25                       &   Selected Sequences                                                      &     TPAMI                                &    2011                       \\
                                  &               &    FMIL                                   &     Haar                           &                                           &  Yes                                     &         No                              &         Yes                              &   50                                   &    12 Selected Sequences                       &       Pattern Recognition                                                                                  &    2015                       \\ 
                                   &                                       & \multicolumn{1}{l}{CMIL}          & \multicolumn{1}{l}{HOG, Distribution Field}                     & \multicolumn{1}{c}{}                 & \multicolumn{1}{c}{No}                 & \multicolumn{1}{c}{No}                 & \multicolumn{1}{c}{Yes}                & \multicolumn{1}{c}{22}     & \multicolumn{1}{l}{20 Selected Sequences}                                   & \multicolumn{1}{l}{Neurocomputing}               &         2017                  \\ \cline{2-12} 
                                   & \multirow{3}{*}{Sparsity-based}       & SST                            & gray scale                                &                                       & Yes                                   & No                                    & Yes                                  & 0.45                      & 20 Selected Sequences                                   & CVPR                                & 2015                      \\
                                   &                                       & RSST                           & gray scale, HOG and CNN                   & SST                                   & Yes                                   & No                                    & Yes                                  & 1.8                       & 40 selected sequences, OTB2013, OTB2015 and VOT2014     & TPAMI                               & 2018                      \\ 
                                   &                                       & \multicolumn{1}{l}{CEST}      & \multicolumn{1}{l}{Pixel color values}   & \multicolumn{1}{c}{}                 & \multicolumn{1}{c}{Yes}              & \multicolumn{1}{c}{No}               & \multicolumn{1}{c}{Yes}             & \multicolumn{1}{c}{4.77} & \multicolumn{1}{l}{15 Selected Sequences}              & \multicolumn{1}{l}{T-CYBERNETICS}  & 2016                      \\ \cline{2-12} 
                                   & \multirow{3}{*}{Siamese-based}        & GOTURN                         & CNN                                       &                                       & Yes                                   & Yes                                   & No                                   & 100                       & VOT2014                                                 & ECCV                                & 2016                      \\
                                   &                                       & SINT                           & CNN                                       &                                       & Yes                                   & Yes                                   & No                                   &                           & OTB2013                                                 & CVPR                                & 2016                      \\ 
                                   &                                       & \multicolumn{1}{l}{YCNN}      & \multicolumn{1}{l}{CNN}                  & \multicolumn{1}{c}{}                 & \multicolumn{1}{c}{Yes}              & \multicolumn{1}{c}{Yes}              & \multicolumn{1}{c}{No}              & \multicolumn{1}{c}{45}   & \multicolumn{1}{l}{OTB2015, VOT2014}                   & \multicolumn{1}{l}{T-CSVT}         & 2017                      \\ \cline{2-12} 
                                   & \multirow{2}{*}{Superpixel}           & BKG                            & HSI color histogram                       &                                       & Yes                                   & No                                    & Yes                                  & 0.77                      & 12 Selected sequences                                   & T-CSVT                              & 2014                      \\ 
                                   &                                       & \multicolumn{1}{l}{SPT}       & \multicolumn{1}{l}{HSI color histogram}  & \multicolumn{1}{c}{}                 & \multicolumn{1}{c}{Yes}              & \multicolumn{1}{c}{No}               & \multicolumn{1}{c}{Yes}             & \multicolumn{1}{c}{5}    & \multicolumn{1}{l}{12 Selected sequences}              & \multicolumn{1}{l}{TIP}            & 2014                      \\ \cline{2-12} 
                                   & \multirow{2}{*}{Graph}                & TCNN                           & CNN                                       &                                       & Yes                                   & Yes                                   & Yes                                  & 1.5                       & OTB2013, OTB2015, VOT2015                               & arXiv                               & 2016                      \\ 
                                   &                                       & \multicolumn{1}{l}{SAT}       & \multicolumn{1}{l}{HSV color histogram}  & \multicolumn{1}{c}{}                 & \multicolumn{1}{c}{Yes}              & \multicolumn{1}{c}{No}               & \multicolumn{1}{c}{Yes}             & \multicolumn{1}{c}{0.5}  & \multicolumn{1}{l}{Deform-SOT}                         & \multicolumn{1}{l}{TIP}            & 2016                      \\ \cline{2-12} 
                                   & \multirow{2}{*}{Part-based}           & ALMM                           & Haar, Raw pixels, Historgam features      & Stuck                                 & No                                    & No                                    & Yes                                  & 40                        & OTB2013, 57 Selected Sequences                          & T-CSVT                              & 2017                      \\
                                   &                                       & PT                             & Haar Features                             &                                       & No                                    & No                                    & Yes                                  & 2.5                       & 13 Selected Sequences                                   & TPAMI                               & 2017                      \\ \hline
\end{tabular}}
\vspace{-25pt}
\end{table}

\section{Experiments and Analysis}\label{experimental_study}

We have performed exhaustive experiments on three publicly available visual object tracking benchmarks including OTB2013 \cite{wu2013online}, OTB2015 \cite{wu2015object}, and TC-128 \cite{liang2015encoding}. 
We also evaluated these trackers on our newly introduced benchmark \textbf{Object Tracking and Temple Color} (OTTC) as explained in Section \ref{benchmarks}.
First, we give brief introduction of selected benchmarks, evaluation protocols and selected trackers for comparison. Then, we report a detailed analysis of experimental study performed over selected benchmarks and  provide our insights and findings. 
A project page is available containing benchmark videos and results on \url{http://bit.ly/2TV46Ka}.

\begin{table}
\vspace{-5pt}
\caption{Details of different benchmarks.}
\vspace{-12pt}
\scalebox{0.8}{
\begin{tabular}{lccccc}
\hline
Benchmarks   & OTB2013 & OTB2015 & TC-128 & OTTC & VOT2017 \\
 \hline  \hline
Sequences    & 50      & 100     & 128    & 186    & 60      \\
Min frames   & 71      & 71      & 71     & 71     & 41      \\
Max frames   & 3872    & 3872    & 3872   & 3872   & 1500    \\
Total frames & 29491   & 59040   & 55346  &  92023      & 21356  \\ \hline
\vspace{-15pt}
\end{tabular}}
\label{benchmarks_compare}
\end{table}

\subsection{Benchmarks}\label{benchmarks}
\textbf{OTB2013}  \cite{wu2013online} contains 50 sequences which are divided into 11 different challenges including Motion Blur (MB), Occlusion (Occ), Deformation (DEF), In-Plane Rotation (IPR),  Fast Motion (FM), Low Resolution (LR), Scale Variation (SV),  Background Clutter (BC), Out-of-View (OV), Illumination Variation (IV), and Out-of-Plane Rotation (OPR). 
\textbf{OTB2015} \cite{wu2015object} is an improved version of OTB2013 consisting of 100 sequences and covering the same 11  challenges. \textbf{TC-128}  \cite{liang2015encoding} is another  benchmark comprising of 128 sequences distributed over the same 11 challenges. We combined all the sequences from OTB2015 and TC-128 to form a more challenging benchmark, named as Object Tracking and Temple Color (\textbf{OTTC}). The new benchmark is a union of unique sequences from OTB2015 and TC-128 avoiding repetitions. OTTC contains 186 sequences distributed over 11  challenges. Tracking performances varies depending upon the number of sequences and the length of each sequence. Since OTB2015 and TC-128 contained 42 common sequences, a benchmark contained each sequence only once was needed to evaluate the tracking performance  in a more comprehensive way.
Table \ref{benchmarks_compare} shows some details of these benchmarks. We also report results over Visual Object Tracking (VOT2017) \cite{Kristan2017visual} benchmark which   covers only five challenges including size change, occlusion, motion change, illumination change, and camera motion. Note that these challenges are more elaborately covered by OTTC benchmark, though using slightly different challenge names. Nevertheless, VOT is an important benchmark due to inclusion of very small target tracking and IR sequences. 


\begin{table}[]
\centering
\caption{Detailed information of selected trackers including category, features, FPS computed over OTTC benchmark, implementation details and resource link. Abbreviations are as follows: Basic (B), Regularized (R), Siamese (S), Part Based (PB), Patch Learning (PL),  Intensity Adjusted Histogram (IAH), Pixel Intensity Histogram (PIH), Color Names (CN),  Color Histogram (CH), Matlab (M), and MatConvNet (m) \cite{vedaldi2015matconvnet}.}
\label{trackers_list}
\vspace{-10pt}
\scalebox{0.54}{
\begin{tabular}{|c|c|c|c|c|c|c|}
\hline
\textbf{Trackers} & \textbf{Category} & \textbf{Feature} & \textbf{FPS} & \textbf{Implementation} & \textbf{GPU} & \textbf{Resource Link}                                                                                           \\ \hline
CSRDCF            & R-CFT              & HOG, CN, Gray    & 8.17         & M                       & No           & https://github.com/alanlukezic/csr-dcf                                                                  \\ \hline
ECO               & R-CFT              & CNN, HOG, CN     & 6.72         & M+m                     & Yes          & https://github.com/martin-danelljan/ECO                                                                 \\ \hline
CCOT              & R-CFT              & CNN              & 0.41         & M+m                     & Yes          & https://github.com/martin-danelljan/Continuous-ConvOp                                                   \\ \hline
STRCF             & R-CFT              & HOG, CN, Gray    & 19.03        & M                       & No           & https://github.com/lifeng9472/STRCF                                                                     \\ \hline
MCPF              & B-CFT              & CNN              & 0.15         & M+m                     & Yes          & http://nlpr-web.ia.ac.cn/mmc/homepage/tzzhang/mcpf.html                                                 \\ \hline
SRDCF             & R-CFT              & HOG, CN          & 3.78         & M                       & No           & https://www.cvl.isy.liu.se/research/objrec/visualtracking/regvistrack/                                  \\ \hline
DCFNet            & S-CFT        & CNN              & 1.72         & M+m                     & Yes          & https://github.com/foolwood/DCFNet\#dcfnet-discriminant-correlation-filters-network-for-visual-tracking \\ \hline
deepSRDCF         & R-CFT              & CNN              & 1.62         & M+m                     & Yes          & https://www.cvl.isy.liu.se/research/objrec/visualtracking/regvistrack/                                  \\ \hline
BACF              & R-CFT              & HOG              & 18.49        & M                       & No           & http://www.hamedkiani.com/bacf.html                                                                     \\ \hline
SRDCFdecon        & R-CFT              & HOG              & 1.48         & M                       & No           & https://www.cvl.isy.liu.se/research/objrec/visualtracking/decontrack/index.html                         \\ \hline
CF2               & B-CFT              & CNN              & 7.01         & M+m                     & Yes          & https://sites.google.com/site/jbhuang0604/publications/cf2                                              \\ \hline
SDLSSVM           & PL     & IAH, RGB Image   & 5.92         & M                       & No           & http://www4.comp.polyu.edu.hk/$\sim$cslzhang/DLSSVM/DLSSVM.htm                                          \\ \hline
HDT               & B-CFT              & CNN              & 5.68         & M+m                     & Yes          & https://sites.google.com/site/yuankiqi/hdt/                                                             \\ \hline
RPT               & PB-CFT     & HOG              & 6.27         & M                       & No           & https://github.com/ihpdep/rpt                                                                           \\ \hline
ECT               & PL     & CNN              & 0.4          & M+m                     & Yes          & https://sites.google.com/site/changxingao/ecnn                                                          \\ \hline
ILCT              & B-CFT              & HOG, PIH, IAH    & 19.29        & M                       & No           & https://sites.google.com/site/chaoma99/cf-lstm                                                          \\ \hline
SiameseFC         & S-CFT         & CNN              & 24.8         & M+m                       & Yes          & http://www.robots.ox.ac.uk/$\sim$luca/siamese-fc.html                                                   \\ \hline
CFNet         & S-CFT         & CNN              & 13.64         & M+m                       & Yes          & http://www.robots.ox.ac.uk/~luca/cfnet.html                                                   \\ \hline
STAPLE            & B-CFT              & HOG, CH   & 6.35         & M                       & No           & https://github.com/bertinetto/staple                                                                    \\ \hline
fDSST             & B-CFT              & HOG              & 65.8         & M                       & No           & http://www.cvl.isy.liu.se/en/research/objrec/visualtracking/scalvistrack/index.html                     \\ \hline
Obli-Raf          & PL     & CNN              & 1.73         & M+m                     & Yes          & https://sites.google.com/site/zhangleuestc/incremental-oblique-random-forest                            \\ \hline
KCF               & B-CFT              & HOG              & 80.85        & M                       & No           & http://www.robots.ox.ac.uk/$\sim$joao/circulant/                                                        \\ \hline
CNT               & PL     & Image Pixels     & 0.46         & M                       & No           & http://faculty.ucmerced.edu/mhyang/project/cnt/                                                         \\ \hline
BIT               & PL     & Gabor, CN        & 37.02        & M                       & No           & http://caibolun.github.io/BIT/index.html                                                                \\ \hline
\end{tabular}}
\vspace{-10pt}
\end{table}

\subsection{Evaluation Protocols}
We employed three metrics including precision, success and speed for comparison.
We have evaluated the robustness of the tracking algorithms using traditional One Pass Evaluation (OPE) technique.
Precision and success plots are drawn to examine the performance of trackers.
For precision, the Euclidean distance is computed between the estimated centers and ground-truth centers as: $\delta_{gp}$ = $\sqrt{(x_g-x_p)^2 + (y_g-y_p)^2}$,
where $(x_g,y_g)$ represents ground truth center location, and $(x_p, y_p)$ is the predicted target center position  in a frame. 
If  $\delta_{gp}$ is less than a threshold than that frame is considered as a successful.
In the precision plot, the threshold $\delta_{gp}$ is fixed 20 pixels. Precision does not give a clear picture of estimated target size and shape because center position error quantifies the pixel difference.
Therefore, a more robust measure known as success has often been used. 
For success, an Overlap Score (OS) between ground truth and the estimated bounding boxes is calculated.
Let $r_g$ be the ground-truth bounding box and $r_t$ be the target bounding box.
An overlap score is defined as: $o_s = (|r_t \cap r_g|)/(|r_t \cup r_g|)$,
where $\cap$ and $\cup$ denotes the intersection and union of two regions respectively, while $| \cdot |$ represents the number of pixels.  OS is used to determine whether a tracking algorithm has successfully tracked a target in the frame. Those frames having $o_s$ scores greater than a threshold,  are referred as successful frames. In the success plot, the threshold value $t_0$ varies between 1 and 0, hence producing varying resultant curves. OS threshold $t_0$ value is set at 0.5 for evaluation. In this work we report average success by computing average OS over all the frames in the benchmark. Similarly, we report average precision over all the frames in a benchmark. We reported precision and success curves as Area under the curve (AUC).
We also reported the speed of the trackers in Frames Per Second (FPS). FPS is the average speed of the trackers over all the sequences in a benchmark.

\subsection{Tracking Algorithms}
We have selected 24 trackers proposed over the last 4 years having error-free easily executable codes (Table \ref{trackers_list}). Selected trackers include CSRDCF \cite{Lukezic_CVPR_2017}, STRCF \cite{Li2018STRCF}, deepSRDCF \cite{danelljan2015convolutional}, CF2 \cite{ma2015hierarchical}, DCFNet \cite{wang17dcfnet}, BACF \cite{kiani2017learning}, ECO \cite{DanelljanCVPR2017}, CCOT \cite{DanelljanECCV2016}, CFNet \cite{valmadre2017end}, CNT \cite{zhang2016robust}, KCF \cite{henriques2015high}, HDT \cite{qi2016hedged}, ECT \cite{gao2016enhancement}, Obli-Raf \cite{Zhang2017CVPR}, MCPF \cite{Zhang_2017_CVPR}, SiameseFC \cite{bertinetto2016fully}, SRDCF \cite{danelljan2015learning}, SRDCFdecon \cite{danelljan2016adaptive}, STAPLE \cite{bertinetto2016staple}, fDSST \cite{danelljan2017discriminative}, DLSSVM \cite{ning2016object}, ILCT \cite{ma2018adaptive}, RPT \cite{li2015reliable}, and BIT \cite{cai2016bit}.
Moreover, the selected trackers are popular among the research community and are often compared on publicly available benchmarks. Most of the trackers do not require any pre-training and easy to execute without any technical difficulty. However, SiameseFC, CFNet, and DCFNet require pre-training but tested  without offline training.
All codes are executed by using default parameters as suggested by the original authors. We have presented both quantitative and qualitative analysis of all the compared trackers. Most of the compared trackers are from CFTs category. In future, if we find any executable codes especially for Non-CFTs, we will upload results on our project page.

\begin{figure}
\centering
\includegraphics[width=\columnwidth]{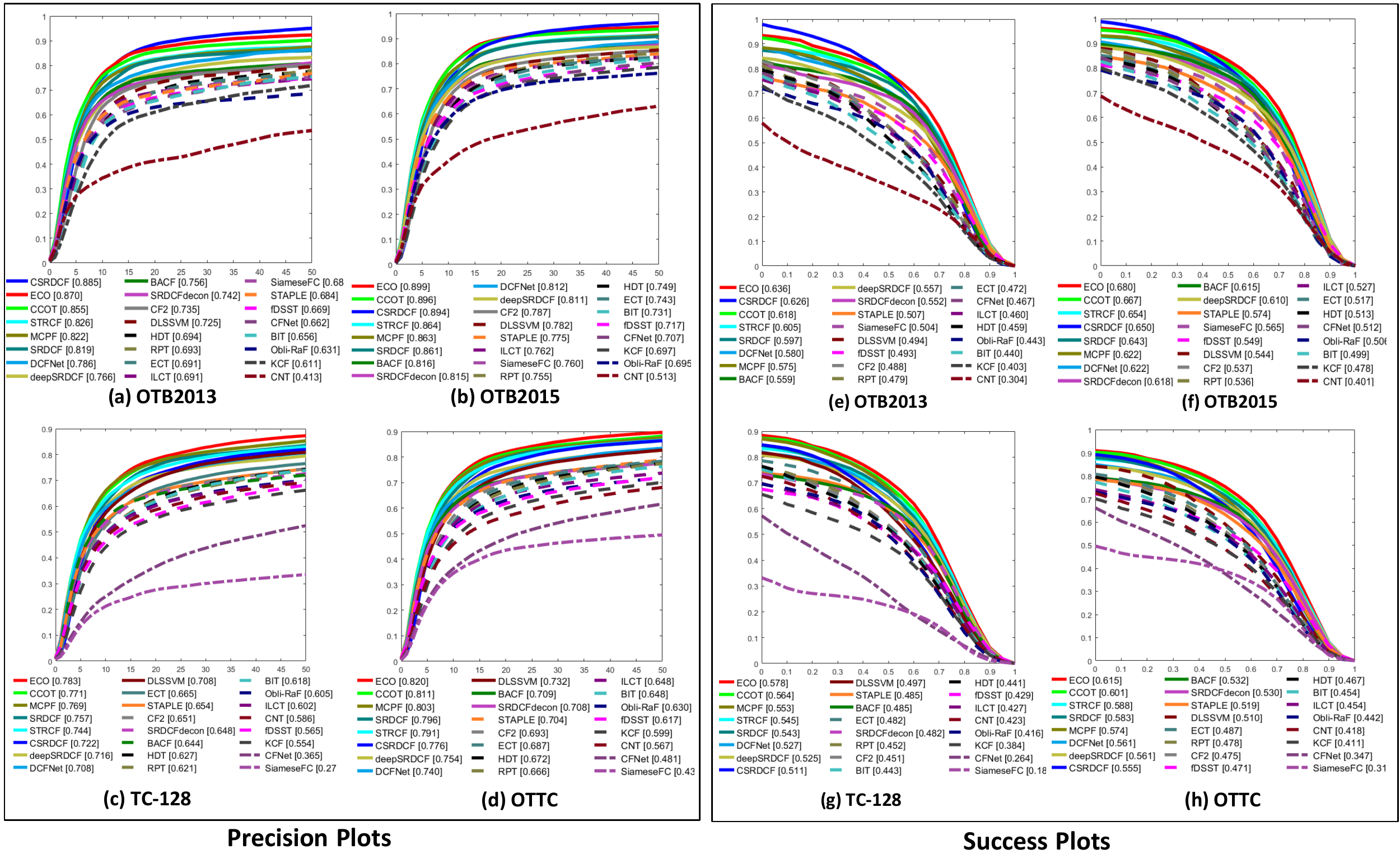}
\vspace{-8mm}
\caption{Precision plots for the 24 selected trackers over (a) OTB2013, (b) OTB2015, (c) TC-128, and (d) OTTC benchmarks. Success plots over (e) OTB2013, (f) OTB2015, (g) TC-128, and (h) OTTC benchmarks are also shown.}
\label{P_S_overall}
\vspace{-12pt}
\end{figure}

\subsection{Quantitative Evaluation}

\subsubsection{Precision and Success}
The performance of trackers is reported in terms of average precision and success (Fig. \ref{P_S_overall}) on four benchmarks including OTB2013, OTB2015, TC-128 and OTTC. The performance of  all the compared trackers in terms of average precision is shown in Fig. \ref{P_S_overall} (a)-(d). On OTB2013, where six trackers including CSRDCF, ECO, CCOT, STRCF, MCPF, and SRDCF obtained the average precision more than 80\% while the performance of the remaining  trackers degraded (Fig. \ref{P_S_overall} (a)).   Similarly,  over OTB2015, the same six trackers  obtained a  precision score of 85\% or above. These 6 trackers, except MCPF, are regularized CFTs, while MCPF is a basic CFT. Four trackers including BACF, SRDCFdcon, DCFNet, deepSRDCF obtained a  precision score of 80\% or above whereas the remaining 14 trackers achieved less than 80\% average precision.
On TC-128 benchmark, majority of the trackers showed degraded performance (average precision less than 75$\%$) while only four trackers ECO, CCOT, MCPF, and SRDCF achieved relatively better performance, \{78.3$\%$, 77.1$\%$,  76.9$\%$, and 75.7$\%$\} respectively (Fig. \ref{P_S_overall} (c)). The OTTC benchmark has been proved to be the most challenging benchmark, with only three trackers including ECO, CCOT, and MCPF achieved more than 80$\%$ average precision (Fig. \ref{P_S_overall} (d)).

The success plots for four benchmarks are shown in Fig. \ref{P_S_overall} (e)-(h). Only four trackers including ECO, CSRDCF, CCOT, and STRCF obtained the  average OS \{63.6$\%$, 62.6$\%$, 61.8$\%$, and 60.5$\%$\} on OTB2013, while the remaining 20 trackers showed a degraded average OS of less than 60.0$\%$ (Fig. \ref{P_S_overall} (e)). The best performing trackers are regulerized CFTs. On OTB2015  (Fig. \ref{P_S_overall} (f)), 10 trackers attained significant improvement in terms of average OS (more than 60.0$\%$) while the remaining trackers showed degraded performance. In contrast, none of the compared trackers achieved an average OS of more than 60.0$\%$ on TC-128 benchmark (Fig. \ref{P_S_overall} (g)).  In case of OTTC benchmark, only ECO and CCOT  performed better (OS 61.5$\%$ and 60.1$\%$) while the other compared trackers achieved less than 60\% OS (Fig. \ref{P_S_overall} (h)). Overall,  ECO obtained the first rank in all of the four benchmarks using both precision and success.  CNT obtained the lowest rank in both OTB2013 and  OTB2015 while  SiameseFC attained the lowest rank in TC-128 and OTTC benchmarks (Figs. \ref{P_S_overall} (e)-(h)).  ECO and  CCOT showed best performance because of the contextual information preserved by using multi-resolution deep features. Most of the best performing trackers are regulerized CCOTs.

We observe that a better performing tracker on one benchmark may not maintain its ranking on the  other benchmarks. For example, CSRDCF and CCOT change their ranking in different benchmarks. CSRDCF performed better than CCOT on OTB2013 while CCOT performed  better  on the other three benchmarks (Fig. \ref{P_S_overall}). The number of sequences and distribution of challenges in a benchmark may change the ranking of a tracker. As OTB2015, TC-128 and OTTC has more number of sequences therefore   ranking may change significantly. Tracking performance is also affected by the number of frames in a sequence. Trackers with poor target update can only perform well over shorter sequences. If a sequence has more  frames then performance on that particular sequence may degrade due to noisy update of the tracker.

\begin{figure*}
\centering
\includegraphics[width=\textwidth]{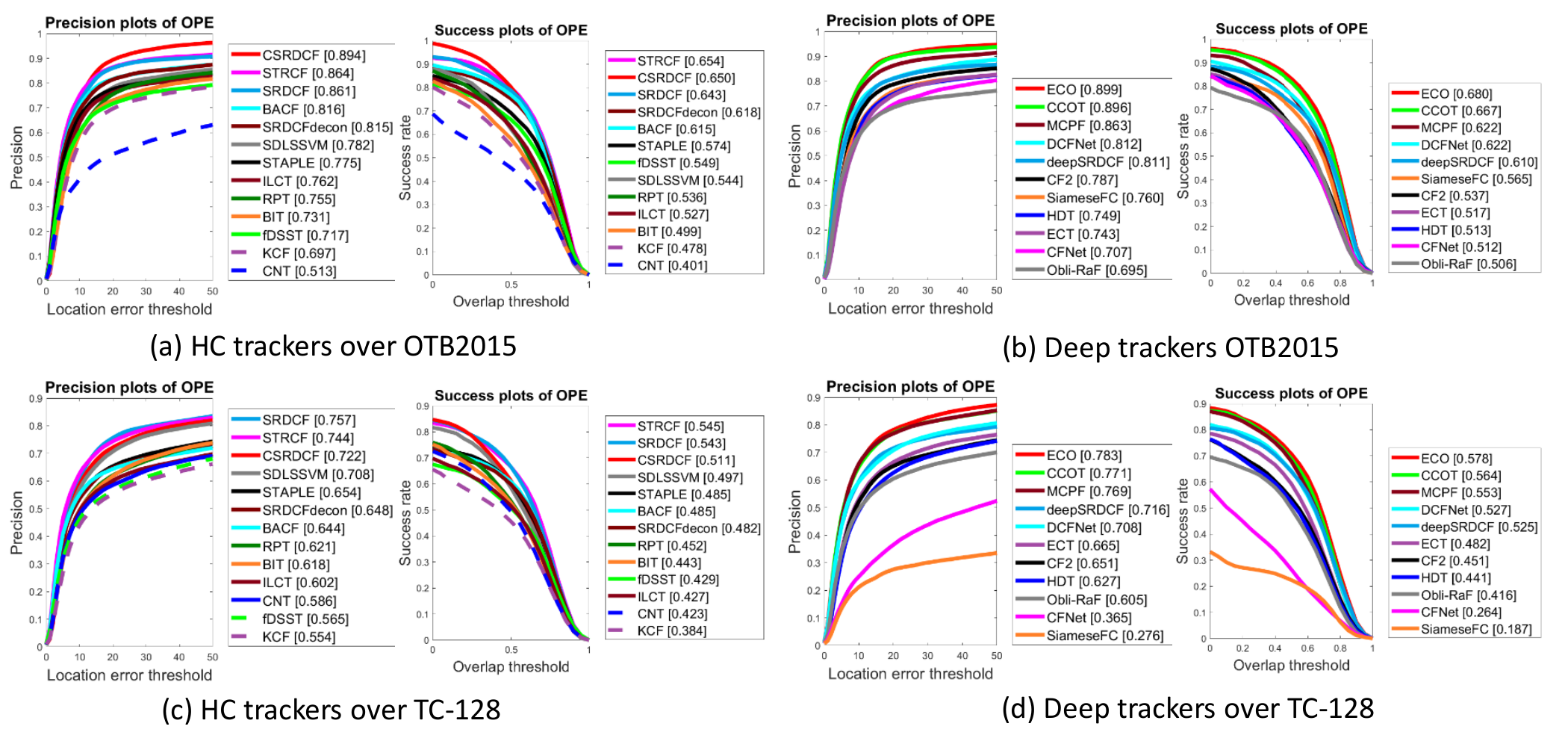}
\vspace{-10mm}
\caption{Precision and success plots for handcrafted (HC) and deep features over OTB2015 and TC-128 benchmarks.}
\label{HC_Deep_Comp}
\vspace{-12pt}
\end{figure*}

\subsubsection{Features based Comparison} 
VOT requires a rich representation of the target appearance based on different types of features. We classify the aforementioned 24 trackers into two categories. The first category  comprises of those trackers which are based on HandCrafted (HC) features such as CSRDCF \cite{Lukezic_CVPR_2017}, STRCF \cite{Li2018STRCF}, SRDCF \cite{danelljan2015learning}, SRDCFdecon \cite{danelljan2016adaptive}, BACF \cite{kiani2017learning},  DLSSVM \cite{ning2016object}, STAPLE \cite{bertinetto2016staple}, ILCT \cite{ma2018adaptive}, RPT \cite{li2015reliable}, BIT \cite{cai2016bit}, fDSST \cite{danelljan2017discriminative}, KCF \cite{henriques2015high}, and CNT\cite{zhang2016robust}. 
While the second category consists of deep feature based trackers including deepSRDCF \cite{danelljan2015convolutional}, CF2\cite{ma2015hierarchical}, DCFNet \cite{wang17dcfnet},  ECO\cite{DanelljanCVPR2017}, CCOT \cite{DanelljanECCV2016}, CFNet \cite{valmadre2017end},  HDT\cite{qi2016hedged},  ECT \cite{gao2016enhancement}, Obli-Raf \cite{Zhang2017CVPR}, MCPF \cite{Zhang_2017_CVPR}, and SiameseFC \cite{bertinetto2016fully}.  We used the default features setting  as proposed  by the original authors.

Figs. \ref{HC_Deep_Comp} (a-d) present the performance of HC and deep feature based trackers using an average precision and success over OTB2015 and TC-128 benchmarks. On OTB2015 benchmark, CSRDCF and STRCF outperformed other HC trackers in terms of precision \{89.4$\%$ and 86.4$\%$\} and success \{65.4$\%$ and 65.0$\%$\} shown in Fig. \ref{HC_Deep_Comp} (a). While in case of deep trackers, ECO and CCOT attained the best performance with precision \{89.9$\%$ and 89.6$\%$\} and success \{68.0$\%$ and 66.7$\%$\} exhibited in (Fig. \ref{HC_Deep_Comp} (b)).
Likewise, over TC-128 benchmark, SRDCF and STRCF performed well using precision and success (Fig. \ref{HC_Deep_Comp} (c)), HC trackers, while ECO and CCOT trackers maintained the highest performance (Fig. \ref{HC_Deep_Comp} (d))  among deep trackers.
\begin{figure*}
\centering
\includegraphics[width=0.88\textwidth]{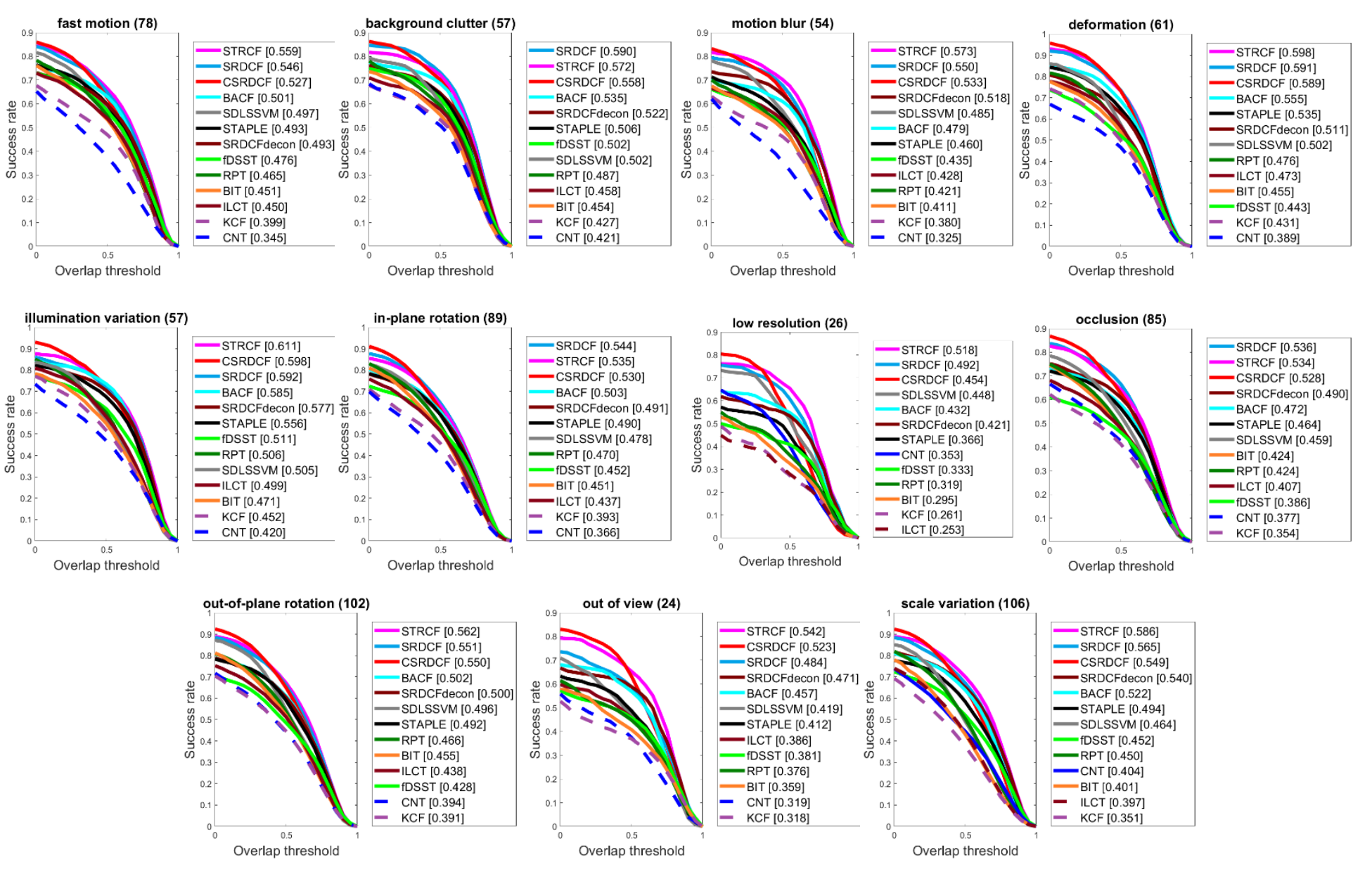}
\vspace{-5mm}
\caption{Success plots of HC trackers for eleven object tracking challenges over OTTC benchmark.}
\vspace{-3mm}
\label{Succ_allseq_HC}
\end{figure*}

\begin{figure}
\centering
\includegraphics[width=0.88\textwidth]{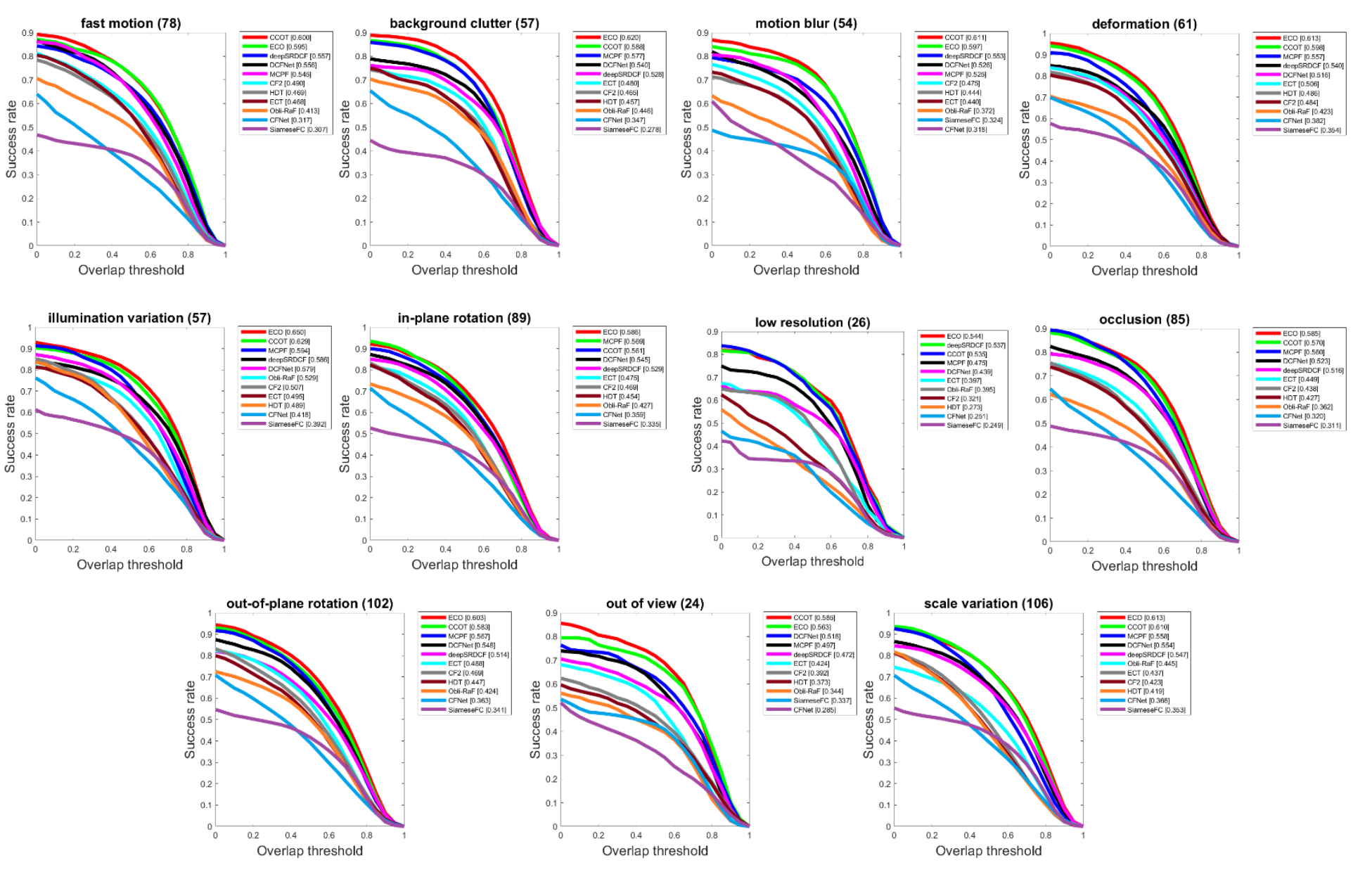}
\vspace{-4mm}
\caption{Success plots of deep trackers for eleven object tracking challenges over OTTC benchmark.}
\label{Succ_allseq_deep}
\vspace{-6mm}
\end{figure}

\begin{table}[]
\caption{Precision  for HC and deep trackers on different challenges over OTTC benchmark.}
\vspace{-10pt}
\label{T_HC_deep_allseq_P_challenges}
\scalebox{0.46}{
\begin{tabular}{c|ccccccccccccc|ccccccccccc|}
\cline{2-25}
\multicolumn{1}{l|}{}            & \multicolumn{13}{c|}{Handcrafted Trackers}                                                                                                                                                                                                                                                                                                                                            & \multicolumn{11}{c|}{Deep Trackers}                                                                                                                                                                                                                                                                                            \\ \hline
\multicolumn{1}{|c|}{Challenges} & \multicolumn{1}{c|}{STRCF} & \multicolumn{1}{c|}{SRDCF} & \multicolumn{1}{c|}{CSRDCF} & \multicolumn{1}{c|}{SDLSSVM} & \multicolumn{1}{c|}{SRDCFdecon} & \multicolumn{1}{c|}{BACF} & \multicolumn{1}{c|}{STAPLE} & \multicolumn{1}{c|}{fDSST} & \multicolumn{1}{c|}{BIT} & \multicolumn{1}{c|}{ILCT} & \multicolumn{1}{c|}{RPT} & \multicolumn{1}{c|}{KCF} & \multicolumn{1}{c|}{CNT} & \multicolumn{1}{c|}{ECO} & \multicolumn{1}{c|}{CCOT} & \multicolumn{1}{c|}{MCPF} & \multicolumn{1}{c|}{deepSRDCF} & \multicolumn{1}{c|}{DCFNet} & \multicolumn{1}{c|}{CF2} & \multicolumn{1}{c|}{HDT} & \multicolumn{1}{c|}{ECT} & \multicolumn{1}{c|}{Obli-Raf} & \multicolumn{1}{c|}{SiameseFC} & \multicolumn{1}{c|}{CFNet} \\ \hline
\multicolumn{1}{|c|}{FM}         &\textbf{71.3}                       & 69.9                       & 68.4                        & 65.9                         & 62.1                            & 64.2                      & 62.6                        & 58.3                       & 58.0                     & 58.2                      & 58.2                     & 53.3                     & 41.0                     & 74.6                     & \textbf{75.2}                      & 72.5                      & 70.0                           & 69.0                        & 66.0                     & 61.8                     & 60.7                     & 55.0                          & 40.1                           & 38.9                       \\ \cline{1-1}
\multicolumn{1}{|c|}{BC}         & 79.0                       & \textbf{83.0}                       & 80.3                        & 73.3                         & 70.9                            & 73.2                      & 70.9                        & 70.5                       & 69.0                     & 66.9                      & 70.2                     & 63.3                     & 58.7                     & \textbf{85.7}                     & 82.6                      & 81.5                      & 74.4                           & 74.4                        & 69.7                     & 68.4                     & 69.5                     & 65.0                          & 40.4                           & 52.0                       \\ \cline{1-1}
\multicolumn{1}{|c|}{MB}         & \textbf{74.2}                       & 73.9                       & 72.4                        & 67.4                         & 67.4                            & 62.4                      & 62.7                        & 56.2                       & 56.7                     & 58.9                      & 58.7                     & 52.8                     & 39.7                     & 78.0                     & \textbf{80.6}                      & 72.5                      & 73.7                           & 68.5                        & 66.9                     & 61.0                     & 60.4                     & 50.2                          & 42.2                           & 42.8                       \\ \cline{1-1}
\multicolumn{1}{|c|}{DEF}        & \textbf{83.7}                       & 82.2                       & 83.0                        & 72.2                         & 70.0                            & 74.9                      & 73.6                        & 57.5                       & 63.5                     & 67.4                      & 67.9                     & 60.5                     & 53.7                     & \textbf{84.2}                     & 83.6                      & 80.1                      & 75.1                           & 75.1                        & 70.7                     & 69.7                     & 71.6                     & 58.4                          & 49.1                           & 53.2                       \\ \cline{1-1}
\multicolumn{1}{|c|}{IV}         & 77.3                       & 75.3                       & \textbf{79.0}                        & 68.3                         & 72.7                            & 74.0                      & 72.4                        & 65.3                       & 65.3                     & 69.4                      & 71.4                     & 64.4                     & 51.1                     & \textbf{82.0}                     & 81.4                      & 80.4                      & 74.7                           & 73.5                        & 72.2                     & 69.6                     & 68.9                     & 69.8                          & 52.3                           & 55.6                       \\ \cline{1-1}
\multicolumn{1}{|c|}{IPR}        & 70.8                       & \textbf{73.1}                       & 71.8                        & 66.5                         & 65.3                            & 66.0                      & 65.3                        & 57.5                       & 61.5                     & 61.0                      & 62.7                     & 57.5                     & 47.3                     & 76.8                     & 75.1                      & \textbf{79.3}                      & 69.9                           & 70.0                        & 67.1                     & 62.4                     & 65.7                     & 59.4                          & 44.8                           & 47.2                       \\ \cline{1-1}
\multicolumn{1}{|c|}{LR}         & 75.9                       & 77.3                       & \textbf{77.7}                        & 72.4                         & 60.1                            & 63..4                     & 57.0                        & 50.8                       & 51.9                     & 46.1                      & 54.7                     & 48.0                     & 62.2                     & 82.1                     & \textbf{83.1}                      & 75.0                      & \textbf{83.1}                           & 66.4                        & 58.6                     & 51.3                     & 66.3                     & 65.5                          & 42.9                           & 45.1                       \\ \cline{1-1}
\multicolumn{1}{|c|}{OCC}        & 70.3                       & \textbf{71.9}                       & 71.0                        & 65.1                         & 63.6                            & 60.5                      & 60.9                        & 49.1                       & 57.0                     & 56.4                      & 56.1                     & 49.3                     & 49.7                     & \textbf{77.3}                     & 74.5                      & 77.1                      & 68.1                           & 67.5                        & 61.5                     & 59.0                     & 61.5                     & 51.1                          & 41.3                           & 42.6                       \\ \cline{1-1}
\multicolumn{1}{|c|}{OPR}        & \textbf{75.9}                       & 74.7                       & 74.5                        & 71.0                         & 66.6                            & 65.7                      & 66.2                        & 54.5                       & 63.4                     & 61.9                      & 62.7                     & 56.4                     & 51.4                     & \textbf{80.0}                     & 78.0                      & 77.8                      & 68.5                           & 70.7                        & 68.0                     & 62.2                     & 66.8                     & 59.3                          & 47.1                           & 48.6                       \\ \cline{1-1}
\multicolumn{1}{|c|}{OV}         & 71.6                       & 67.6                       & \textbf{75.2}                        & 58.9                         & 60.7                            & 61.6                      & 58.7                        & 47.2                       & 46.9                     & 51.3                      & 49.9                     & 41.3                     & 42.7                     & 77.9                     & \textbf{81.8}                      & 68.2                      & 66.3                           & 69.6                        & 51.9                     & 49.5                     & 57.0                     & 47.6                          & 46.1                           & 39.0                       \\ \cline{1-1}
\multicolumn{1}{|c|}{SC}         & \textbf{79.0}                       & 76.9                       & 77.1                        & 69.8                         & 72.2                            & 70.0                      & 68.3                        & 58.8                       & 60.4                     & 60.8                      & 64.4                     & 55.8                     & 54.7                     & 81.0                     & 81.4                      & \textbf{81.7}                      & 73.1                           & 72.7                        & 66.2                     & 64.7                     & 67.1                     & 63.5                          & 47.6                           & 51.1                       \\ \hline \cline{1-1}
\multicolumn{1}{|c|}{Overall}    & 79.1                       & \textbf{79.6}                       & 77.6                        & 73.2                         & 70.8                            & 70.9                      & 70.4                        & 61.7                       & 64.8                     & 64.8                      & 66.6                     & 59.9                     & 56.7                     & \textbf{82.0}                     & 81.1                      & 80.3                      & 75.4                           & 74.0                        & 69.3                     & 67.2                     & 68.7                     & 63.0                          & 43.5                           & 48.1                       \\  \hline \cline{1-1}
\end{tabular}}
\vspace{-15pt}
\end{table}

\subsection{Challenges-based Analysis}
Most of the  trackers cannot exhibit excellent performance on all the  tracking challenges. The challenges included in this study are Fast Motion (FM), Motion Blur (MB), Occlusion (OCC), Deformation (DEF), Illumination Variation (IV),  Low Resolution (LR),  In-Plane Rotation (IPR),  Out-of-Plane Rotations (OPR), Out-of-View (OV), Background Clutter (BC), and Scale Variations (SV).  A challenge based evaluation of tracker performance has been performed in terms of precision (Table \ref{T_HC_deep_allseq_P_challenges}) and  success (Figs. \ref{Succ_allseq_HC}, \ref{Succ_allseq_deep}) over OTTC benchmark. 

In the fast motion and motion blur challenges, the target appearance is blurred by target or camera motion. In these challenges, STRCF and SRDCF has performed  the best among HC trackers. While CCOT and ECO handled these challenge successfully among  the deep  trackers. Both CCOT and ECO achieved the best performance for these challenges compared to all the trackers  (Table \ref{T_HC_deep_allseq_P_challenges} and Figs. \ref{Succ_allseq_HC}, \ref{Succ_allseq_deep}). In fast motion sequences, target position changes rapidly and target is tracked by exploring a large search space. Whereas for motion blur where target suffers significant appearance variations due to target or camera motion, tracking is performed by exploiting useful features to assist in localizing the target. An overview of all the trackers included in this study reveals that multi-resolution feature maps are only  computed by ECO and CCOT which may be considered as significantly contributing to the performance of these trackers in these challenges. Other trackers were not able to  efficiently handle these challenges because of the fast movement of target or camera in the video.

In deformation challenge, target exhibits different variations in shape and orientation while in occlusion challenge target hides itself partially or fully behind other objects in the scene. Tracking performance is highly influenced by these challenges. Usually, trackers add the background information as noise when target observes appearance variation which leads to the drift problem that is the tracked bounding box gradually moves away from the actual target bounding box and starts tracking un-required objects. In these challenges, performance of the tackers may be improved by handling drift problem efficiently. SRDCF, CSRDCF, and STRCF have best handled these challenges among HC trackers while ECO, CCOT, and MCPF showed better performance among the deep trackers. The best performing ECO tracker has handled these challenges efficiently by  computing Gaussian Mixture Models for each target appearance. Also ECO has proposed reduced number of features compared to CCOT thus dropping some of the weak features which may correspond to the undesired target regions under occlusion thus obtaining better performance.


In some sequences in OTTB benchmark, illumination variations and low resolution  challenges simultaneously appear posing challenge for most of the  trackers. To address illumination challenge, a tracker needs to maintain a target model as well as a local background model. In low  resolution videos, target appearance is also low resolution.  This problem can be handled by utilizing more stronger features of target. CSRDCF, STRCF and SRDCF handled these challenges in a more appropriate manner  compared to the other HC trackers. In deep trackers, ECO, deepSRDCF and CCOT  showed improved performed  over both challenges.   It is noted that ECO is best in terms of success while CCOT and deepSRDCF performed best in terms of precision compared to both the HC and the deep trackers. For low resolution targets, efficient feature extraction is difficult thus deep features help in improved performance for low resolution.  Both ECO and CCOT also take the advantage of multi-resolution deep features for improved performance.


The influence of  Out of View (OV)  challenge presented one of the most severe challenges to majority of the trackers which can handle OV challenge by maintaining useful target samples to re-detect the tracker after failure.  HC trackers such as CSRDCF, STRCF, SRDCF showed better performance over OV sequences using precision while STRCF showed significant improvement in terms of success.  On the other hand, deep CCOT and ECO handled OV challenge with respect to precision and success.  ECO and CCOT handled OV challenge efficiently with the assistance of multiple convolutional operators to learn multi-resolution features. Other challenges such as In-Plane Rotation (IPR) and Out-of-Plane Rotations (OPR) are handled by rotation invariant features and keeping multiple target representations.  Over the group of IPR sequences,  trackers including SRDCF, CSRDCF, and STRCF  exhibited the best performance among HC trackers and among deep trackers ECO, MCPF, CCOT obtained better performance. Similarly, STRCF and SRDCF tackled OPR challenge efficiently in HC category while ECO, CCOT and MCPF performed well among deep  trackers.  We observe that overall ECO and CCOT achieved significant efficiency over  IPR and OPR challenges.

Background clutter and scale variation challenges also difficult to handle in visual object tracking. For background clutter videos, target matches with the background texture while in scale variations target exhibits significant changes in size. Trackers including STRCF, SRDCF,  and CSRDCF in HC category handled well both the background clutter and the scale variations and hence obtained the best precision and success.  The trackers ECO, CCOT, and MCPF best handled these challenges in deep trackers.  Overall, ECO is the best performing deep  tracker in these challenges and showed improved performance by taking advantage   of factorized convolutional operators and updating  GMM components. It is worth noting that the best performing trackers in each challenge are in the Regularized-CFTs category in the proposed hierarchy.

\subsection{Qualitative Evaluation}
For the qualitative comparison, we selected six sequences including \textit{Bolt, CarScale,  FaceOcc1, Jogging-1, Matrix, and Skiing} which cover all the tracking challenges.

\begin{figure}
  \begin{minipage}[b]{.48\textwidth}
    \centering
    \includegraphics[width=\linewidth]{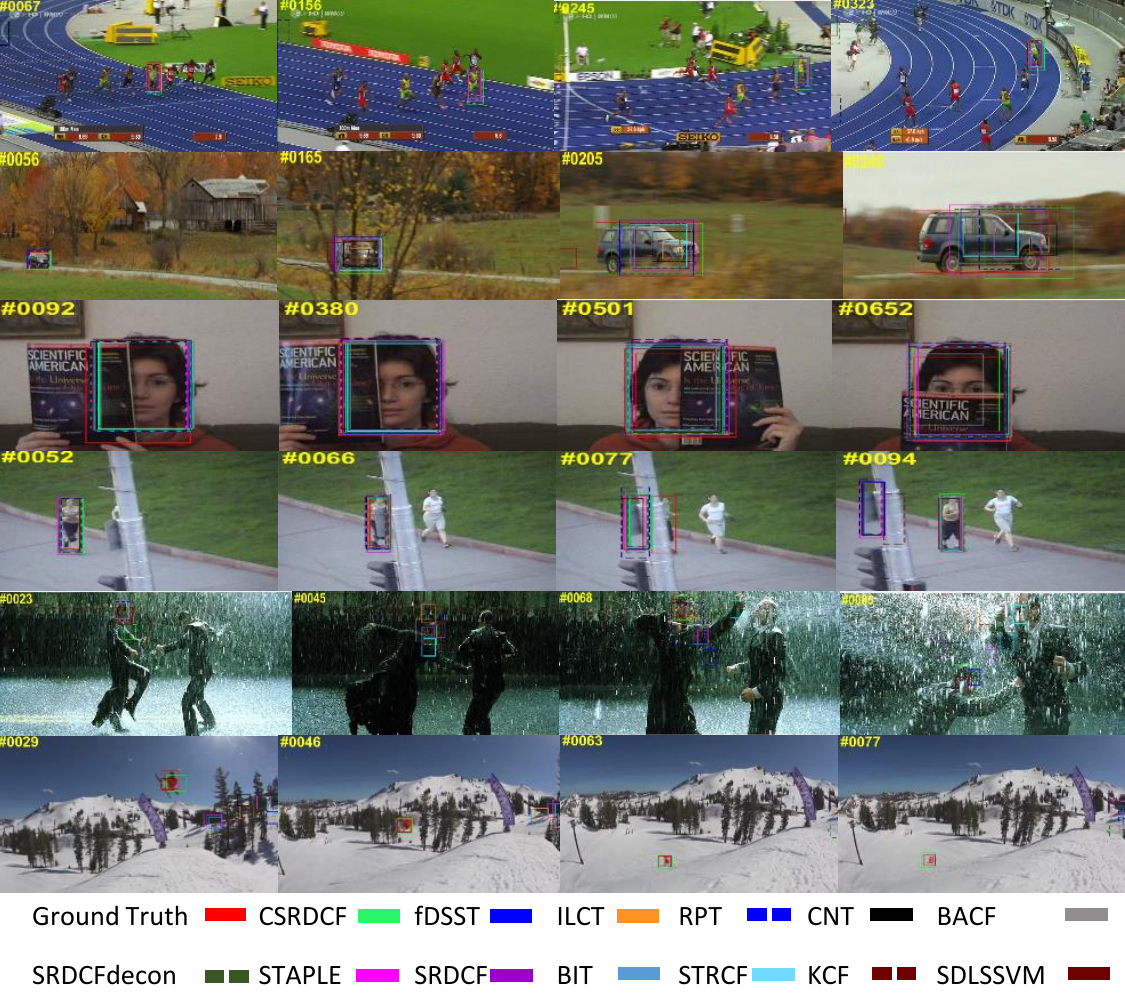}
    \vspace{-8mm}
    \caption{Qualitative analysis of HC trackers on  challenging sequences (from top to bottom \textit{Bolt, CarScale,  FaceOcc1, Jogging-1, Matrix, and Skiing} respectively).}
     \label{hc_qualitative}
  \end{minipage}\hfill
  \begin{minipage}[b]{.48\textwidth}
    \centering
    \includegraphics[width=\linewidth]{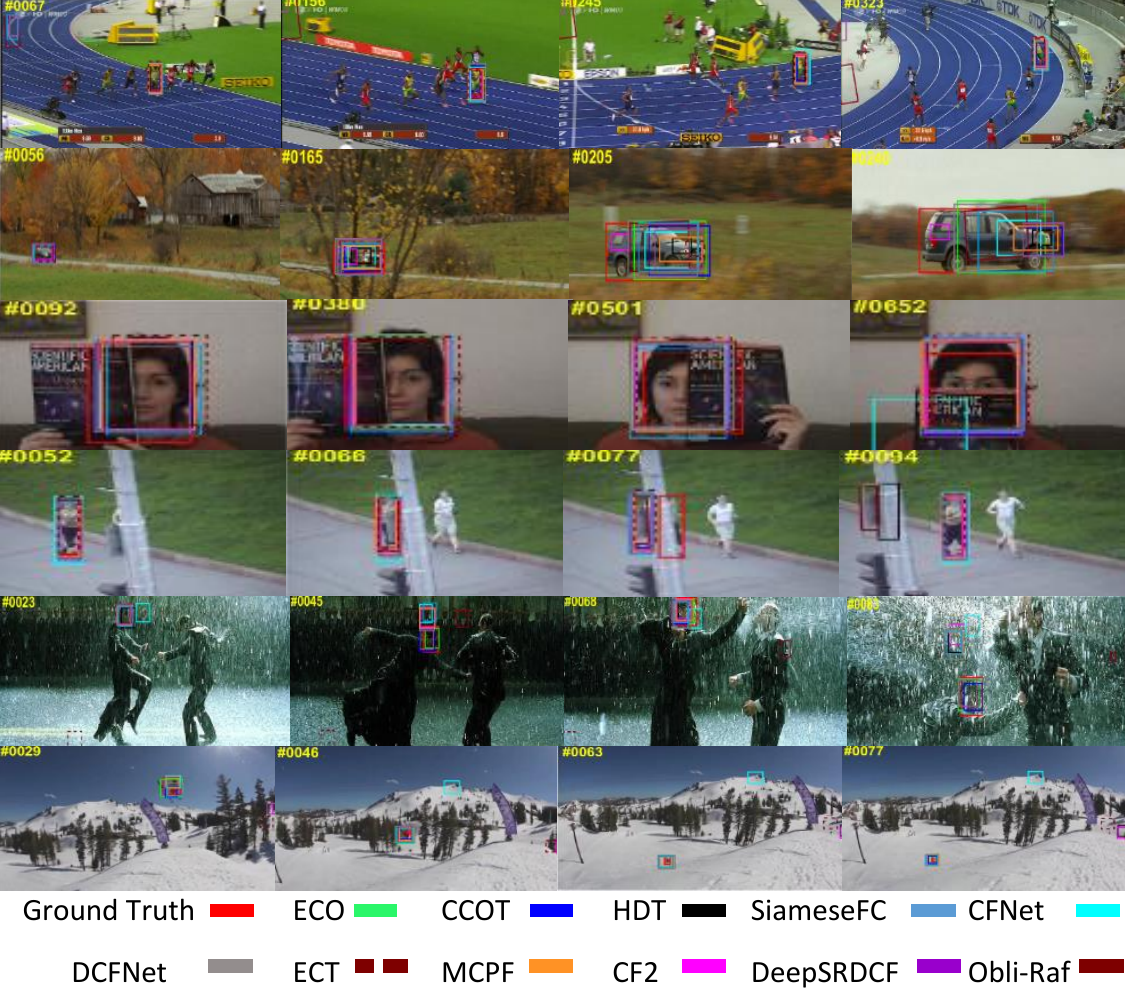}
    \vspace{-8mm}
    \caption{Qualitative analysis of deep trackers on  challenging sequences (from left to right \textit{Bolt, CarScale,  FaceOcc1, Jogging-1, Matrix, and Skiing} respectively).}
     \label{deep_qualitative}
  \end{minipage}
  \vspace{-4mm}
\end{figure}

Figure \ref{hc_qualitative} demonstrates the qualitative comparison for HC trackers. For \textit{Bolt} sequence, most of the trackers such as CSRDCF, STRCF, and SRDCF etc tracked the player successfully while only four trackers SRDCFdecon, RPT, fDSST, and CNT were not able to track the activity of the player through out the sequence.
In the \textit{CarScale}  sequence, majority of the trackers were not able to completely track the car on a road because of the scale variation. 
\textit{FaceOcc1} presents the challenge of partial occlusion, where a woman rotates a book around her face. 
In case of \textit{Jogging-1} sequence,  occlusion because of the pole presented a major challenge for STAPLE, fDSST, KCF, RPT, ILCT and CNT trackers while the remaining trackers successfully tracked the person occluded by the pole.
The lighting variations and fast motion challenges in the sequences \textit{Matrix} and  \textit{Skiing} degraded the performance for most of the trackers excluding  CSRDCF and SDLSSVM in  \textit{Skiing} and CSRDCF, fDSST and SDLSSVM in  \textit{Matrix}.

A qualitative study of deep trackers has also been performed to analyze the visual tracking performance. 
Obli-Raf, SiameseFC, and deepSRDCF missed the player and made their bounding box far beyond the runner for \textit{Bolt} sequence while ECO, CCOT, HDT, CF2, CFNet, DCFNet, MCPF, and ECT turned out to be the successful trackers.  For scale variation sequence \textit{CarScale},  none of the tracker  handled scaling efficiently, however, ECO, DCFNet, Obli-Raf, and SiameseFC tracked the major parts of the vehicle. Over partial occlusion \textit{FaceOcc1} sequence,  all tracker succeeded to track the face except for  CFNet, although it achieved some success but eventually its performance falls off. While for complete occlusion \textit{Jogging-1} sequence, HDT, DCFNet, and  Obli-Raf presented degraded performance.
Another challenging \textit{Matrix} sequence where it is raining, HDT, DCFNet, CF2, MCPF, ECT Obli-Raf, SiameseFC, and CFNet exhibited degraded performance and only ECO, CCOT, and deepSRDCF succeeded. For challenging sequence like \textit{Skiing}, ECO, CCOT, MCPF, SiameseFC, and DCFNet accomplished the tracking task successfully as compared to other deep trackers. 

In short, qualitative study revealed CSRDCF and SRDCF as best trackers among HC tracker while ECO and CCOT showed efficient performance among deep trackers. Additionally, It is noted that these tracking algorithms employed spatial regularization using discriminative correlation filters.

\subsection{Comparison over VOT2017}

\begin{table}

\centering
\caption{Comparison of HC and deep trackers using Robustness (R), Accuracy (A), and Expected average overlap (EAO) measures for baseline and realtime experiments over VOT2017.}
\vspace{-8pt}
\label{VOT_compare}
\scalebox{0.8}{
\begin{tabular}{c|c|c|c|c|c|c|c|}
\cline{2-8}
\multirow{2}{*}{}                                                         & \multirow{2}{*}{\textbf{Tracker}} & \multicolumn{3}{c|}{Baseline}                    & \multicolumn{3}{c|}{Realtime}                    \\ \cline{3-8} 
                                                                          &                                   & \textbf{EAO}   & \textbf{A}     & \textbf{R}     & \textbf{EAO}   & \textbf{A}     & \textbf{R}     \\ \hline
\multicolumn{1}{|c|}{\multirow{5}{*}{HC trackers}} & CSRDCF                            & \textbf{0.256} & 0.491          & \textbf{0.356} & 0.099          & 0.477          & 1.054          \\ \cline{2-8} 
\multicolumn{1}{|c|}{}                                                    & STAPLE                            & 0.169          & \textbf{0.530} & 0.688          & \textbf{0.170} & \textbf{0.530} & \textbf{0.688} \\ \cline{2-8} 
\multicolumn{1}{|c|}{}                                                    & KCF                               & 0.135          & 0.447          & 0.773          & 0.134          & 0.445          & 0.782          \\ \cline{2-8} 
\multicolumn{1}{|c|}{}                                                    & SRDCF                             & 0.119          & 0.490          & 0.974          & 0.058          & 0.377          & 1.999          \\ \cline{2-8} 
\multicolumn{1}{|c|}{}                                                    & DSST                              & 0.079          & 0.395          & 1.452          & 0.077          & 0.396          & 1.480          \\ \hline
\multicolumn{1}{|c|}{\multirow{5}{*}{Deep trackers}}        & CF2                               & \textbf{0.286} & 0.509          & 0.281          & 0.059          & 0.339          & 1.723          \\ \cline{2-8} 
\multicolumn{1}{|c|}{}                                                    & ECO                               & 0.280          & 0.483          & \textbf{0.276} & 0.078          & 0.449          & 1.466          \\ \cline{2-8} 
\multicolumn{1}{|c|}{}                                                    & CCOT                              & 0.267          & 0.494          & 0.318          & 0.058          & 0.326          & 1.461          \\ \cline{2-8} 
\multicolumn{1}{|c|}{}                                                    & SiameseFC                         & 0.188          & 0.502          & 0.585          & \textbf{0.182} & \textbf{0.502} & \textbf{0.504} \\ \cline{2-8} 
\multicolumn{1}{|c|}{}                                                    & MCPF                              & 0.248          & \textbf{0.510} & 0.427          & 0.060          & 0.325          & 1.489          \\ \hline
\end{tabular}}
\vspace{-12pt}
\end{table}

\begin{figure}
  \begin{center}
    \includegraphics[width=\textwidth]{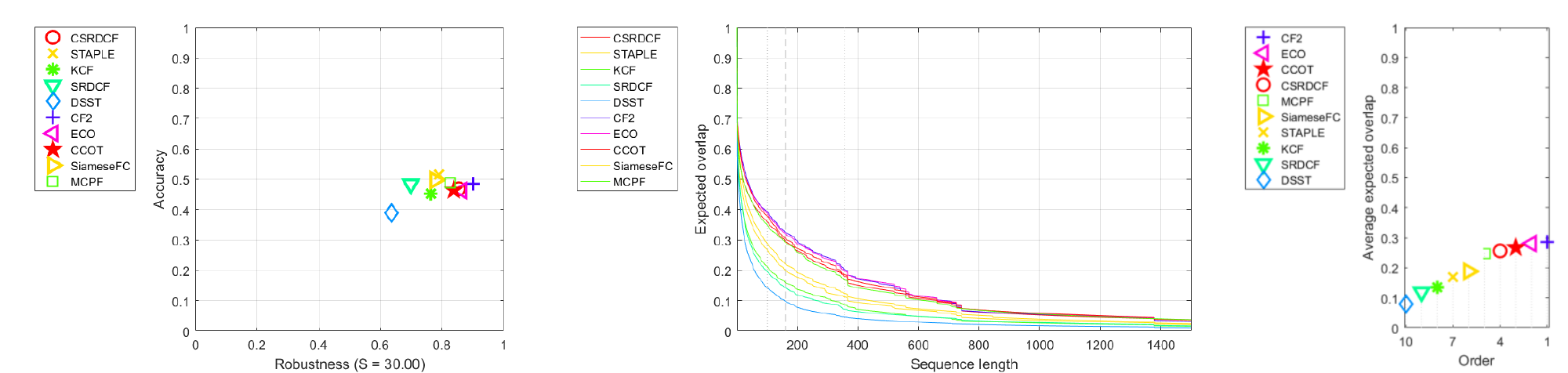}
    \end{center}
         \caption{The AR plot (left), the EAO curves (middle) and EAO plot (right) for the
VOT2017 baseline experiment.}
\label{vot_2017_baseline}
\vspace{-12pt}
\end{figure}

We have also reported results over VOT2017 \cite{Kristan2017visual} benchmark  comparing the performance of various trackers. In VOT toolkit, if a tracker looses the target, it is re-initialized using the ground-truth bounding box. This procedure is named as reset property of the VOT toolbox. It should be noted that no reset is performed in OTB toolkit. In addition to the reset property, VOT toolkit requires multiple runs of the same tracker on the same video to evaluate performance of that tracker, while OTB toolkit performs one pass evaluation.  For performance comparison, VOT has three primary measures such as Accuracy (A), Robustness (R) and Expected Average Overlap (EAO). The average overlap score between ground truth and predicted bounding boxes for successful tracking intervals is known as accuracy. 
Robustness is reported as the how many times a tracker failed and required reset. Stochastic tracking algorithms are executed multiple times for each sequence. VOT toolkit reduces the potential bias added due to reset property in the accuracy measure by discarding 10 frames after re-initialization. Averaging robustness and accuracy over multiple runs for a single sequence gives the per sequence robustness and accuracy.
EAO is  used to measure the expected  overlap score computed for typical short-term sequence lengths over an interval by averaging the scores for the expected average overlap curve (see more details \cite{Kristan2015visual}). 
The results shown in Table  \ref{VOT_compare} are drawn from \cite{Kristan2017visual}. Tracking results are shown for baseline and real-time experiments separately for handcrafted and deep feature based trackers. Fig. \ref{vot_2017_baseline} shows the results from baseline experiment. In VOT toolkit, in real-time experiments a tracker at the first frame, is initialized  and waits for the output bounding box to be computed and the next frame to be read from memory. If the next frame becomes available before completion of computation for the previous bounding box, than that  bounding box  is   output for the next frame as well. In baseline experiments, a bounding box is computed for each frame.
In our analysis, we have included trackers with results reported in \cite{Kristan2017visual} and overlapping with the trackers shown in Table \ref{trackers_list}. Among the compared trackers, CSRDCF performed best among HC trackers using EAO and R measures, while STAPLE performed best using A measure for the baseline experiments. On the other hand, for real-time experiments, STPALE showed  significant improvement in  performance among HC trackers  for all measures.  For deep trackers, in baseline experiments, CF2 exhibited best results with EAO=0.286, MCPF is the best with A=0.510,  and ECO is the best with R=0.276. For real-time experiments, SiameseFC showed best performance  in terms of all measures.

\begin{figure}

     \vspace{-5pt}
  \begin{center}
    \includegraphics[width=0.8\textwidth]{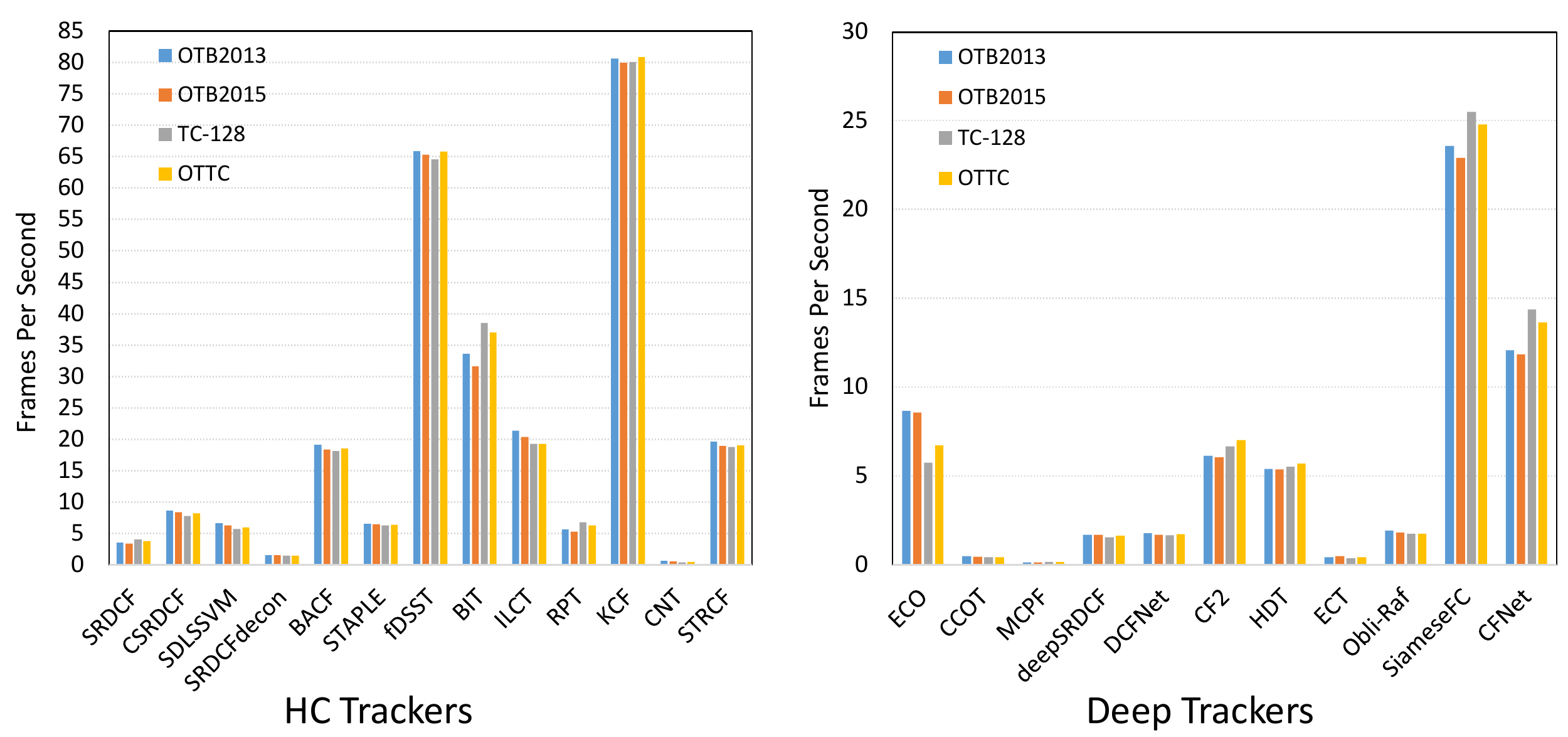}
    \end{center}
     \vspace{-16pt}
         \caption{Speed analysis of HC and deep trackers in frames per second.}
\label{hc_deep_fps}
\vspace{-15pt}
\end{figure}


\subsection{Speed Comparison}
An efficient tracker must track target accurately and robustly with low computational cost. 
For real world applications, computational complexity takes a vital role in the selection of a specific tracker. Tracking model complexity and target update frequency are the fundamental reasons effecting the speed of a tracker.  We have reported the average speed of the HC and deep feature based trackers over OTB2013, OTB2015, TC-128, and OTTC benchmarks (Fig. \ref{hc_deep_fps}).  For all the compared trackers, we have used default  parameters as given by the original authors. For a fair speed comparison, all experiments are performed on the same computer with Intel Core i5 CPU 3.40 GHz and 8 GB RAM. GeForce GTX 650 GPU is being used for deep trackers. Overall HC trackers require less computational cost as compared to deep trackers which often take a lot of time for feature extraction at each frame. The online model update for each frame in deep trackers incur high computational cost. The KCF tracker is ranked as the fastest tracker compared to all the trackers. Speed comparison demonstrates that KCF, fDSST and BIT HC trackers process more frames in one second than the fastest deep tracker which is SiameseFC. Some HC trackers including SRDCF, SRDCFdecon, and CNT require high computational cost therefore  FPS is less than 5. Similarly, deep trackers including CCOT, MCPF, deepSRDCF, DCFNet, ECT, Obli-Raf, and ECT are also computational complex with speed less than 5 FPS as shown in Fig. \ref{hc_deep_fps}. 


\subsection{Comparison of Tracking Performance over Different Features}

We evaluated the performance of ECO over OTB2015 to get better understanding of HC and deep features as shown in Table \ref{comparison_different_features}. We compared HOG and Color Names (CN) as  HC features with deep features obtained from VGGNet and also made different feature combinations. HC features are obtained as the predefined image properties using different algorithms. For example, HOG is a special feature extractor designed to count the occurrence of gradient orientations in a specific region of image. 
However, deep features are able to learn the global as well as local information from the images that is most suitable for a required task. The information is automatically learned by the network while optimizing a given objective function. Multiple convolution operations are performed by sliding customized kernels which are automatically learned over the training data. Each filter computes a different image attribute keeping in view a required tracking task. Each convolution layer computes features at a different level providing multiple abstraction levels. Earlier layers preserve low level spatial information while later layers retain high level semantic information \cite{ma2015hierarchical}. Deep features computed by the earlier layers may be considered in a way similar to HC features given that a HC feature was most suitable for the challenge to be handled. In our experiment, deep features are extracted from each VGG layer represented as Conv\textit{i}, where \textit{i} is the layer index. Deep features are also extracted from more than one VGG layers, for example Conv123 are extracted from layers 1, 2 and 3. We observe that Conv12 obtained best performance compared to only HC and single layer deep features. However,  the combination of both HOG and deep features i.e., HOG$\_$Conv15 exhibited overall best performance. The increase in accuracy shows that there is some information in HC features which was not encoded by the deep features. Therefore, fusion of both types of features was able to obtain improved accuracy.  For real-time applications, computation time is very important. In general, deep features take more execution time compared to the HC features. Thus, there is a trade-off between accuracy and computational complexity, that is more accurate feature combinations may take more computational time. 

\begin{table}[]
\caption{Tracking performance comparison of ECO algorithm using different types of features}
\vspace{-12pt}
\scalebox{.7}{
\label{comparison_different_features}
\begin{tabular}{lccccccccccccc}
\hline
                   & \textbf{HOG} & \textbf{CN} & \textbf{HOG\_CN} & \textbf{Conv1} & \textbf{Conv2} & \textbf{Conv3} & \textbf{Conv4} & \textbf{Conv5} & \textbf{Conv12} & \textbf{Conv123} & \textbf{Conv1234} & \textbf{Conv12345} & \textbf{HOG\_Conv15} \\ \hline \hline
\textbf{Precision} & 78.5         & 58.3        & 80.6             & 69.8           & 77.6           & 62.1           & 69.5           & 73.0           & 81.5            & 74.4             & 73.6              & 73.4               & \textbf{90.0}                 \\
\textbf{Success}   & 59.9         & 43.4        & 61.1             & 51.6           & 54.6           & 42.5           & 46.5           & 47.1           & 61.5            & 55.7             & 55.4              & 55.0               & \textbf{68.2}                 \\
\textbf{FPS}       & \textbf{51.50}        & 14.11       & 23.77            & 21.33          & 20.76          & 17.54          & 17.13          & 16.80          & 11.88           & 6.84             & 5.76              & 4.25               & 6.58            \\ \hline    
\end{tabular}}
\vspace{-12pt}
\end{table}

\subsection{Findings and Recommendations}
 
In this work we have  explored  tracking problem using Correlation Filter based Trackers (CFTs) and Non-CFTs proposed over the past few years and we believe that it is still an open problem. We also performed an extensive  analysis of state-of-the-art trackers with respect to handcrafted and deep features. 
Qualitative and quantitative evaluations reveal  that on a broader level, CFTs  have shown better performance over Non-CFTs. The CFTs are using both handcrafted and deep features.  At a finer level, we observe that regularized CFTs  exhibit excellent performance within this category.  Therefore, we recommend the use of Regularized-CFTs because of their potential for further improvement. Spatial and temporal regularization has been used in Disrcriminative Correlation Filters (DCF) to improve the tracking performance.

Our study concludes that it is required to learn efficient discriminative features preserving geometric, structural, and spatial target information. Structural information encodes  appearance variations while geometric information captures the shape and spatial information encodes the location of different parts. Deep convolutional features encode low-level spatial and high-level semantic information which is vital for precise target positioning while HC features encode less semantic information. Thus efficient fusion of HC and deep features including low-level and high-level information captures invariant complex features from geometry and structure of targets and enhances the tracking performance.

The performance of the trackers can be improved by including temporal information along with spatial information. Recurrent Neural Network (RNN) models capture the temporal relationship among sequential images. Although trackers using RNN models are proposed to integrate temporal information including RTT \cite{cui2016recurrently} and SANet \cite{fan2016sanet}, however the improvement in results is not significant. Currently, RNN have not been much explored by the tracking community, therefore it remains unclear how much performance improvement can be obtained by employing architectures similar to RNN. Thus, it may be a possible future research direction.

Learning based tracking algorithms suffer due to lack of training data availability. Usually object bounding box is available only in the first frame. Recently, zero-shot and one-shot learning are studied to alleviate data limitation problem and it is also a new direction yet to be explored in tracking domain.
In tracking-by-detection framework, online updating current frame due to limited positive samples can lead to over-fitting problem. For example, at detection stage, target is partially occluded and thus a tracker can learn noisy target shape. For long term tracking, positive target patches capturing all target shape variations are required to learn complete target appearance.
Generative Adversarial Networks (GAN) \cite{goodfellow2014generative} has ability to produce realistic images. Recently, Song et al. \cite{song-cvpr18-VITAL} employed GAN and proposed VIsual Tracking via Adversarial Learning (VITAL) algorithm. Inclusion of GAN and reinforcement learning in tracking frameworks can also effectively improve the tracking performance and is a promising future direction. In short, the tracker's ability to efficiently learn target's shape, appearance, and geometry, as well as temporal variations is vital for efficient tracking. Experimental study reveals that more accurate trackers often have high computational cost. Therefore, it is necessary to obtain high tracking speeds without losing accuracy.


\section{Conclusions}

In this study,  a survey of  recent visual object tracking algorithms is performed. Most of these algorithms were published during the last four years. These trackers are classified into two main groups, CFTs and Non-CFTs. Each category is further organized into different classes based on the methodology and framework.  This paper provides interested readers an organized overview of the diverse tracking area as well as trends and proposed frameworks. It will help the readers to find research gaps and provides insights for developing better tracking algorithms. This study also enables the readers to select appropriate trackers for specific real world applications keeping in view the performance of the trackers over different challenges. 
Four different benchmarks including OTB50, OTB100, TC-128 and a new proposed benchmark OTTC are used for performance comparison of 24 algorithms using precision and success measures and execution time for each tracker. This study concludes that regularized CFTs have yielded  better performance compared to the others. Spatial and temporal regularizations emphasizing the object information and suppressing the background in DCFs further enhance the tracking performance. Deep  features have ability to encode low-level and high-level information compared to handcrafted features. Efficient transfer learning while improving accuracy, robustness, and solving the limitation of training data will be new progression.

\section{Acknowledgments}
This research was supported by Basic Science Research Program through the National Research Foundation of Korea (NRF) funded by the Ministry of Education, Science and Technology (NRF-2016R1A2B1015101).


\bibliographystyle{ACM-Reference-Format}
\bibliography{file}



\end{document}